\PassOptionsToPackage{svgnames}{xcolor}
\documentclass[journal]{vgtc}                     % final (journal style)
\onlineid{1111}

\vgtccategory{Representations \& Interaction}

\title{ECoNGS: Efficient Compressive Neural Gaussian Splats for\\Volume Visualization}

\author{%
  \authororcid{Kaiyuan Tang}{0009-0001-3512-0112} and
  \authororcid{Chaoli Wang}{0000-0002-0859-3619}
}

\authorfooter{
  \item The authors are with
the Department of Computer Science and Engineering, University of Notre Dame, Notre Dame, IN 46556, USA.
  \item E-mail: \{ktang2, chaoli.wang\}@nd.edu.
}

\abstract{%
Recent advances in differentiable Gaussian splatting have highlighted the potential of primitive-based approaches as alternative scene representations for interactive, high-quality, volume visualization (VolVis) of large datasets. However, the explicit nature of current primitive-based methods, combined with isolated optimization for each VolVis scene, results in redundant, non-compact representations. We present ECoNGS, an efficient compressive neural Gaussian splatting framework for VolVis scene representation. ECoNGS employs lightweight neural networks to dynamically predict implicit, editable Gaussian splats from explicit anchor points, effectively combining model compactness and parameter efficiency of implicit representations with high-performance rendering of explicit primitives. We explore a joint learning strategy that clusters geometrically similar scenes and shares parameters across them, significantly reducing overall training time and model size while maintaining reconstruction fidelity. To achieve a more compact scene representation, we further compress the explicit anchor attributes using a neural entropy model that estimates their probability distributions, enabling compact storage via entropy coding. We systematically investigate Gaussian initialization strategies and propose a simple yet effective scheme tailored for VolVis scenes, improving reconstruction accuracy and accelerating convergence. We evaluate ECoNGS qualitatively and quantitatively across various univariate and multivariate VolVis scenes, highlighting its superior performance over prior methods in training time, reconstruction quality, and model size.
\hot{In particular, compared with the prior method iVR-GS, ECoNGS improves reconstruction quality by up to 2.2~dB in PSNR while reducing the model size by up to $6.1\times$ and the training time by up to $5.9\times$.} \hot{The code is available at \url{https://github.com/TouKaienn/ECoNGS}.}
}


\graphicspath{{figs/}{figures/}{pictures/}{images/}{./}} % where to search for the images

\usepackage{mathptmx}
\usepackage{graphicx}
\usepackage{times}
\usepackage{algorithm}
\usepackage{algorithmic}
\usepackage{amsmath,amsfonts}
\usepackage{amssymb}
\usepackage{bm}
\usepackage{booktabs}
\usepackage{gensymb}
\usepackage{comment}
\usepackage{caption}
\usepackage{multirow}
\usepackage{array}
\usepackage{textcomp}
\usepackage{stfloats}
\usepackage{url}
\usepackage{verbatim}

\usepackage{colortbl}
\usepackage{xcolor}

\definecolor{myred}{RGB}{255,200,200}
\definecolor{myyellow}{RGB}{255,255,200}

\usepackage{soul}

\newcommand{\hot}[1]{{\color{black} #1}}

\DeclareMathAlphabet{\altmathcal}{OMS}{cmsy}{m}{n}

\begin{document}

%%%%%%%%%%%%%%%%%%%%%%%%%%%%%%%%%%%%%%%%%%%%%%%%%%%%%%%%%%%%%%%%
%%%%%%%%%%%%%%%%%%%%%% START OF THE PAPER %%%%%%%%%%%%%%%%%%%%%%
%%%%%%%%%%%%%%%%%%%%%%%%%%%%%%%%%%%%%%%%%%%%%%%%%%%%%%%%%%%%%%%%

%% The ``\maketitle'' command must be the first command after the
%% ``\begin{document}'' command. It prepares and prints the title block.
%% the only exception to this rule is the \firstsection command
\firstsection{Introduction}
\maketitle
Direct volume rendering (DVR) has long been a key technique in volume visualization (VolVis).
It generates high-quality rendering images by casting rays through the volume, accumulating voxel samples, and mapping voxel values to color and opacity via user-defined transfer functions (TFs).
Users can interactively adjust the TFs and lighting parameters to reveal and analyze the internal structures of complex volume datasets.
However, despite recent advances in GPU architectures, real-time DVR for large-scale volumes remains challenging due to high computational costs and memory bandwidth pressure.

To reduce the cost of DVR for large volumes, recent work has proposed representing VolVis scenes with \textit{novel view synthesis} (NVS) models~\cite{Niedermayr-VMV24, Tang-TVCG25-StyleRF, Tang-TVCG25, Tang-VIS25, Yao-PVIS25, Yao-CG25, Yao-VIS25, Ai-VIS25}.
In particular, primitive-based methods such as \textit{3D Gaussian splatting} (3DGS)~\cite{Kerbl-TOG23} have shown promising results for their superior rendering speed and reconstruction quality compared to \textit{neural radiance field} (NeRF)~\cite{Mildenhall-ECCV20} methods, and further extensions have tailored them to VolVis scene representation. 
For example, 
Niedermayr et al.~\cite{Niedermayr-VMV24} proposed leveraging 3DGS to represent the VolVis scene in large-scale medical datasets, enabling interactive rendering on lightweight mobile devices.
However, vanilla 3DGS inherently embeds the preset lighting and TF information into its learned representation, thereby making them fixed and non-editable during rendering.
To overcome this limitation, iVR-GS~\cite{Tang-TVCG25} introduces editable Gaussian primitives that enable color, opacity, and lighting adjustments at inference time. 
Moreover, it establishes a paradigm for constructing a complete VolVis scene by combining multiple basic TF-specific models, each corresponding to a disjoint visible region.
This paradigm supports real-time visualization and interactive editing of the entire VolVis scene.
Based on iVR-GS, NLI4VolVis~\cite{Ai-VIS25} and TexGS-VolVis~\cite{Tang-VIS25} further extend the framework by integrating multimodal large language models or texture attributes, enabling intuitive interactions and expressive editing of the VolVis scene.

% mention why init is important, cite paper to prove that the init is a non-trivial task for VolVis since alternative methods in CV, such as COLMAP cannot work on texture-less scenes
Despite these advances, existing primitive-based scene representations still suffer from several limitations.
First, although these methods achieve faster optimization and rendering than purely implicit NeRF-based models, they often yield significantly larger model sizes due to their explicit nature. 
Second, existing approaches treat each VolVis scene representation as an independent optimization task, ignoring the inherent correlations among VolVis scenes, resulting in relatively slow convergence and redundant parameter structures. 
%Third, many VolVis images contain limited or homogeneous textures, making it challenging for conventional structure-from-motion (SfM) tools, such as COLMAP, to extract reliable 3D point clouds for initializing the Gaussian primitives.
%Due to the lack of effective initialization strategies, most primitive-based VolVis scene representation methods start optimization from randomly sampled point clouds, often leading to inefficient and unstable training.
Third, many VolVis images contain limited or homogeneous textures, making it difficult for conventional \textit{structure-from-motion} (SfM) tools, such as COLMAP~\cite{Schonberger-CVPR16},
%~\footnote{\url{https://github.com/colmap/colmap/issues/2514}}, 
to extract reliable 3D point clouds for initializing Gaussian primitives, forcing most primitive-based methods to start from random point clouds and resulting in inefficient, unstable training. % at the early stage.

To respond, we propose ECoNGS, an \underline{E}fficient \underline{Co}mpressive \underline{N}eural \underline{G}aussian \underline{S}platting framework for VolVis scene representation. 
To reduce redundancy, ECoNGS adopts a hybrid structure consisting of \textit{anchor points} and lightweight \textit{multilayer perceptrons} (MLPs) that predict editable neural Gaussian splats on-the-fly during rendering, combining the efficiency of \textit{explicit} methods with the compactness of \textit{implicit} models. 
We further integrate a \textit{neural entropy model} to compress anchor points, substantially reducing storage costs.
To exploit correlations across different VolVis scenes, we address \textit{multi-scene modeling} through \textit{joint learning}. 
ECoNGS captures inter-scene relationships and yields more compact, accurate representations by clustering similar TF-based scenes and partially sharing parameters among them. 
We study initialization strategies for editable Gaussian primitives and propose a simple yet effective scheme tailored for VolVis scenes that greatly improves convergence speed and reconstruction accuracy.
The contribution of ECoNGS can be summarized as follows. 
%\vspace{-0.05in}
%\begin{myitemize}
	%\item 
    (1) We present ECoNGS, a hybrid neural-explicit Gaussian splatting framework that reduces redundancy while preserving training and rendering efficiency.
	%\item 
    (2) We introduce a multi-scene joint optimization paradigm that leverages correlations among TF-based VolVis scenes to improve model compactness and convergence speed.
	%\item 
    (3) We study editable Gaussian initialization for VolVis and propose a lightweight yet effective initialization strategy.
%\vspace{-0.05in}
%\end{myitemize}

\vspace{-0.075in}
\section{Related Work}

{\bf Deep learning in VolVis.}
Deep learning has emerged as a powerful tool in VolVis~\cite{Wang-TVCG23}, with recent studies tackling diverse tasks, including data generation~\cite{Han-VIS19, Han-VIS21, Han-TVCG22, Tang-CG24}, compression~\cite{Lu-CGF21, Tang-PVIS24, Han-VIS25, Tang-VIS26}, and neural representation~\cite{Han-TVCG23, Chen-TVCG25, Yang-PVISVN25, Tang-VISSP26}.
For example,
\hot{Engel and Ropinski~\cite{Engel-TVCG20} used a CNN to predict per-voxel occlusion, enabling more realistic volumetric rendering.}
\hot{Han et al.~\cite{Han-VIS21} introduced STNet, an end-to-end framework for spatiotemporal super-resolution of volumetric data.}
\hot{Lu et al.~\cite{Lu-CGF21} proposed compressive implicit neural representations for volumetric scalar fields.}
\hot{Wu et al.~\cite{Wu-TVCG25} developed a distributed neural representation for interactive in situ visualization at scale.}
\hot{Chen et al.~\cite{Chen-TVCG25} designed an explorable INR to explore ensemble simulations across spatial and parameter domains efficiently.}
\hot{Sun et al.~\cite{Sun-VIS25} proposed F-Hash to speed up implicit neural representation for encoding time-varying volumes.}
\hot{Bi et al.~\cite{Bi-VIS25} introduced CD-TVD for accurate 3D super-resolution from scarce samples.}

Beyond that, recent advances have explored VolVis synthesis.
For instance,
\hot{Berger et al.~\cite{Berger-TVCG19} introduced a GAN-based framework that renders volume images conditioned on viewpoints and TFs.}
\hot{He et al.~\cite{He-InsituNet} presented InSituNet, a surrogate that maps simulation and visualization parameters to DVR images for in situ ensemble exploration.}
\hot{Shi et al.~\cite{Shi-VDLSurrogate} proposed VDL-Surrogate, which uses view-dependent latent encodings for high-quality previews of ensemble results.}
\hot{Li et al.~\cite{Li-VIS24} introduced ParamsDrag, allowing parameter-space exploration by directly dragging visual features in image space.}

\hot{Unlike these methods that directly predict visualization images from parameters, our approach uses lightweight neural networks to build 3D neural Gaussian splats capturing the scene geometry, and renders them for real-time visualization.}

%--------------------------------------
\begin{figure*}[!ht]
\centering
\includegraphics[width=\linewidth]{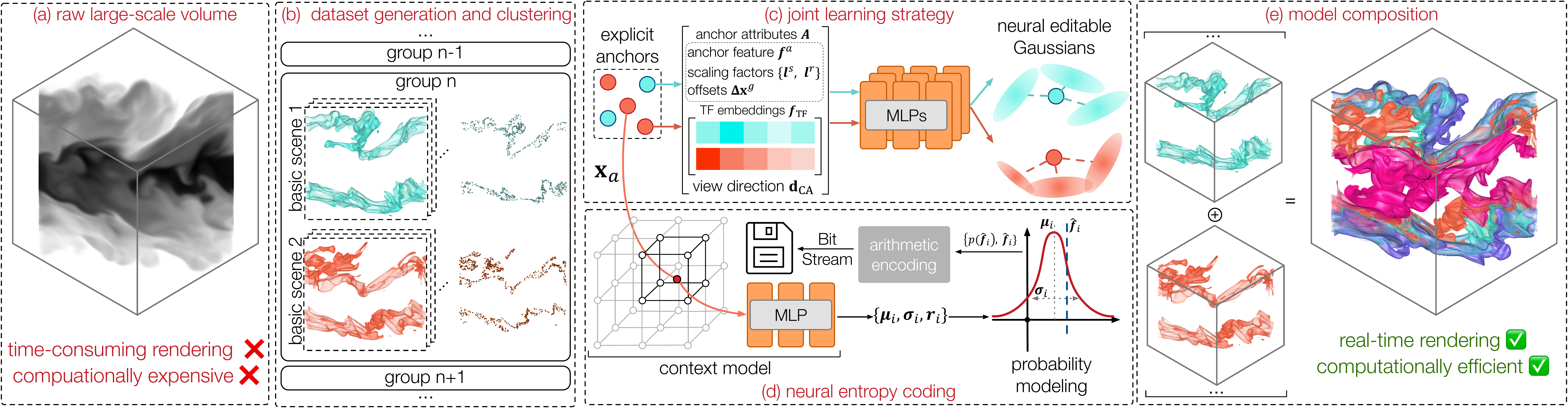}
\vspace{-.25in}
\caption{Overview of the ECoNGS framework.
(a) Rendering large-scale volumetric data through DVR is computationally expensive.
(b) The dataset is decomposed into multiple TF-specific basic scenes, which are further clustered based on geometric similarity.
(c) During joint learning, neural editable Gaussians are decoded from shared MLPs using anchor attributes, TF embeddings, and view directions.
(d) A neural entropy model learns the probability distributions of explicit anchor attributes and compresses them into compact bitstreams through arithmetic encoding.
(e) Finally, all trained basic scenes are composed into a complete VolVis scene that supports real-time rendering and interactive visualization.}
\label{fig:workflow}
\vspace{-.1in}
\end{figure*}
%--------------------------------------

{\bf Primitive-based differentiable rendering.}
Early efforts such as EWA splatting~\cite{Zwicker-VIS01} and point-based rendering~\cite{Sainz-CG04} laid the foundation for primitive-based approaches with efficient splat-based rasterization.
More recently, Kerbl et al.~\cite{Kerbl-TOG23} proposed 3DGS, a differentiable primitive representation that enables real-time rendering of radiance fields.
\hot{Many follow-up works adapt Gaussian splatting to diverse scenarios.}
For instance,
\hot{Gao et al.~\cite{Gao-ECCV24} extended 3DGS with BRDF parameters to decompose scene lighting.}
\hot{Kocabas et al.~\cite{Kocabas-CVPR24} developed HUGS to represent animatable humans by combining 3DGS with parametric body models.}
\hot{Wu et al.~\cite{Wu-CVPR25} proposed 3DGUT, replacing the EWA splatting in 3DGS with the unscented transform for distorted cameras and rolling-shutter effects.}

3DGS has also inspired a series of follow-up works in VolVis.
Tang et al.~\cite{Tang-TVCG25} introduced iVR-GS, establishing a VolVis scene representation paradigm that optimizes editable Gaussian primitives for disjoint basic scenes and composes them into a complete representation for real-time exploration.
Building on this paradigm,
\hot{Ai et al.~\cite{Ai-VIS25} developed NLI4VolVis, integrating multi-agent large language models to interact with language-augmented editable Gaussian splats.}
\hot{Tang et al.~\cite{Tang-VIS25} introduced TexGS-VolVis, a textured 2DGS framework for image- and text-driven non-photorealistic VolVis scene editing.}
\hot{Yao and Wang~\cite{Yao-VIS25} presented VolSegGS, which employs deformable Gaussian splats for volumetric segmentation of dynamic scenes.}
\hot{Bauer et al.~\cite{David-VIS25} proposed GSCache, a radiance caching scheme for interactive photorealistic volume path tracing.}
\hot{Dyken et al.~\cite{Dyken-VEG25} proposed VEG, a TF-agnostic Gaussian model that encodes the scalar field underlying the VolVis scene.}

\hot{Similar to iVR-GS~\cite{Tang-TVCG25}, we represent a complete VolVis scene by optimizing multiple basic scenes, while exploiting neural Gaussian representations~\cite{Lu-CVPR24, Mihajlovic-ECCV24, Liu-ICASSP25} to remove potential primitive redundancy and improve accuracy.}
Beyond this, we investigate the underexplored issues and the relations among various VolVis scenes during training.
\hot{Compared with iVR-GS, ECoNGS improves all four metrics simultaneously, including training time, rendering framerate, reconstruction accuracy, and model size, rather than trading one off against another.}

{\bf Joint learning.}
\hot{Joint learning trains multiple related tasks or components together in a unified model, sharing information to improve performance.}
For example,
\hot{Zhang et al.~\cite{Zhang-ECCV14} showed that jointly learning facial landmark detection with auxiliary tasks, such as head pose and attribute estimation, improves robustness to occlusion and pose.}
\hot{Zhao et al.~\cite{Zhao-RecSys19} built a large-scale multi-objective ranking system with a multi-gate mixture-of-experts architecture for multiple, potentially conflicting objectives.}
\hot{Wang et al.~\cite{Wang-CVPR25} pre-trained a transformer on multiple 3D prediction tasks, including camera, depth, and track estimation, improving downstream 3D applications.}
In line with these works, ECoNGS shows that jointly optimizing over multiple related VolVis scenes reduces parameter redundancy and enables faster training.

{\bf Neural entropy model.}
Existing VolVis representations often rely on vector quantization~\cite{Niedermayr-VMV24, Tang-TVCG25} to reduce the storage size of Gaussian splats, which is efficient but often introduces noticeable distortion.
In contrast, the image~\cite{Cheng-CVPR20, He-CVPR21, He-CVPR22, Pan-CVPR26} and video~\cite{Li-NeurIPS21, Sheng-TMM23, Jia-CVPR25} compression communities have widely adopted neural entropy models that learn probability distributions over latent codes for more compact and adaptive compression.
Recently, such neural entropy models have also shown their effectiveness for compressing NVS representations~\cite{Wang-NeurIPS24, Liu-Arxiv24, Chen-CVPR24, Chen-ECCV24}.
\hot{Unlike vector quantization used by existing VolVis methods~\cite{Niedermayr-VMV24, Tang-TVCG25}, this work introduces a neural entropy model that compresses explicit anchor attributes with minimal distortion and shares them across jointly learned scenes to exploit inter-scene redundancy for a more compact representation.}

\vspace{-0.075in}
\section{ECoNGS}

Figure~\ref{fig:workflow} illustrates the overview of ECoNGS.
Our goal is to transform the time-consuming rendering of large-scale volumes into a computationally efficient process using the proposed scene representation.
To achieve this, we extract multi-view images and corresponding point clouds (Section~\ref{subsec:initialization}) from different basic scenes for optimizing neural editable \hot{Gaussians} (Section~\ref{subsec:neuralEditableGaussian}) and group geometrically similar scenes for joint learning (Section~\ref{subsec:jointTrain}).
In addition, to further reduce the size of the explicit model components, we apply neural entropy coding (Section~\ref{subsec:InContextCompression}) to compress the anchor attributes.
Finally, all trained basic scenes are composited to form a complete VolVis scene representation that supports real-time, efficient rendering.

\vspace{-0.075in}
\subsection{Preliminaries}

{\bf 3D Gaussian splatting.}  
3DGS~\cite{Kerbl-TOG23} represents a scene using a collection of explicit 3D Gaussian primitives, each associated with a set of differentiable attributes that can be rendered efficiently through tile-based rasterization.
Given the position attribute $\mathbf{x}^g$ of one Gaussian point, it can be defined as
\begin{equation}
\vspace{-0.025in}
	G(\mathbf{x})=e^{-\tfrac{1}{2}(\mathbf{x}-\mathbf{x}^g)^\top \Sigma^{-1}(\mathbf{x}-\mathbf{x}^g)},
\vspace{-0.025in}
\end{equation}
where $\mathbf{x}$ is an arbitrary 3D position and $\Sigma$ denotes the covariance matrix of the Gaussian primitive.
To ensure positive semi-definiteness during optimization, $\Sigma$ is formulated by two scaling and rotation matrices that are parameterized by a scaling factor $\mathbf{s}$ and a quaternion rotation vector $\mathbf{q}$, respectively.

Unlike the original volumetric representation, 3DGS can efficiently render the scene via tile-based rasterization rather than computation-intensive ray marching.
Specifically, 3D Gaussian primitives $G(\mathbf{x})$ are first transformed to 2D Gaussians $G'(\mathbf{x'})$ on the image plane following the projection process described in EWA volume splatting~\cite{Zwicker-VIS01}.
Then the pixel color $\mathbf{C}(\mathbf{x'})$ at 2D position $\mathbf{x'}$ is determined by employing $\alpha$-blending to the color $\mathbf{c}$ and opacity $o$ attributes of $N$ depth-sorted layered 2D Gaussians. i.e.,
\begin{equation}
\vspace{-0.025in}
    \mathbf{C}(\mathbf{x'}) = \sum_{i \in N} c_i \alpha_i \prod_{j=1}^{i-1} (1 - \alpha_j), \ \alpha_i = o_i G_i'(\mathbf{x'}),
\vspace{-0.025in}    
\end{equation}
where $\mathbf{c}$ is modeled by a set of \textit{spherical harmonics} (SH) parameters.
By using the differentiable rasterizer, all attribute values $\{\mathbf{x}^g, \mathbf{q}, \mathbf{s}, o, \mathbf{c}\}$ are learnable and optimized via training-view reconstruction.

{\bf Editable Gaussians for explorable VolVis.} 
\label{subsubsec:editableGaussian}
Directly applying 3DGS to fit VolVis scenes bakes the TFs (color and opacity) and lighting conditions into the Gaussian primitive representation, rendering them uneditable during visualization and severely limiting interactive volume exploration.
To overcome this limitation, several studies~\cite{Tang-TVCG25, Tang-VIS25, Ai-VIS25} represent VolVis scenes using editable Gaussian primitives that decouple color and lighting information during training, thereby enabling color, opacity, and lighting editing at inference time.
Specifically, each editable Gaussian, in addition to the existing geometry-related attributes $\{\mathbf{x}^g, \mathbf{q}, \mathbf{s}, o\}$ for vanilla 3DGS, is equipped with a set of differentiable shading-related attributes $\{\Delta\mathbf{c}, \bm{n}, k^a, k^d, k^s, \beta \}$. 
Among them, $\Delta\mathbf{c}$ denotes the offset color, which, together with a learnable palette color $\mathbf{c}_p$ shared by all Gaussians within a basic scene, represents the diffuse color of each Gaussian. 
The attribute $\bm{n}$ encodes the normal direction, while $k^a$, $k^d$, $k^s$, and $\beta$ correspond to the ambient, diffuse, specular, and shininess coefficients in the Blinn-Phong shading model~\cite{Blinn-Phong}. 
By optimizing these shading attributes during training, various editing operations can be achieved at inference time: adjusting the shared palette color $\mathbf{c}_p$ changes the color of the basic scene, scaling $o$ modifies its opacity, and altering the light source position or scaling $k^a$, $k^d$, $k^s$, and $\beta$ enables intuitive control over the lighting direction and intensity through the Blinn-Phong model.

\vspace{-0.075in}
\subsection{Sparse Point Cloud Extraction}
\label{subsec:initialization}

Traditional SfM methods (e.g., COLMAP~\cite{Schonberger-CVPR16}) provide sparse point clouds to initialize 3D Gaussians with reliable geometric priors. However, they often fail in VolVis scenes due to textureless regions and a lack of feature correspondences, while their bundle adjustment is computationally intensive.
Since the original volumetric data are available, we directly extract a sparse point cloud from the volume for initialization.
Specifically, we select voxels with nonzero opacity values and retain their colors according to the TF mappings. 
For efficiency, we subsample the volume to a volume of $ < 256^3$ and randomly select no more than 10k opaque voxels, \hot{ensuring the entire extraction completes in under one second with negligible overhead}.

\vspace{-0.075in}
\subsection{Neural Editable Gaussian Representation}
\label{subsec:neuralEditableGaussian}

% why hybrid-structure
% Following the hybrid neural Gaussian design~\cite {Lu-CVPR24, Mihajlovic-ECCV24, Liu-ICASSP25}, we develop ECoNGS with neural editable Gaussians for scene representation, reducing potential parameter redundancy while preserving fast training and inference.
\hot{Building on the hybrid neural Gaussian design~\cite{Lu-CVPR24, Mihajlovic-ECCV24, Liu-ICASSP25}, we develop ECoNGS with neural editable Gaussians for scene representation, reducing potential parameter redundancy while preserving fast training and inference.}
The core idea of the neural-editable Gaussian representation is to model the scene using \textit{anchor points} that distribute local, editable Gaussians.
In particular, we parameterize each anchor point with a unique spatial position $\mathbf{x}_a\in\mathbb{R}^3$ and anchor attributes $\bm{A}=\{\bm{f}^a\in\mathbb{R}^{50}, \bm{l}^s\in\mathbb{R}^3, \bm{l}^r\in\mathbb{R}^3, \Delta\mathbf{x}^{g}\in\mathbb{R}^{3K}\}$, where each component represents anchor feature, anchor scaling factors for scale and rotation, as well as position offsets for $K$ neighboring editable Gaussians of the anchor, respectively.
Before training, we initialize the anchor points according to the sparse point cloud described in Section~\ref{subsec:initialization}.
The point positions within the sparse point cloud \hot{are} utilized to initialize $\mathbf{x}_a$, and we assign the average color of points to palette color $\mathbf{c}_p$.

During training and inference, the attribute values described in Section~\ref{subsubsec:editableGaussian} of the $K$ local editable Gaussians can be derived from their anchor point.
Given an anchor point at $\mathbf{x}_a$, the positions for each of its $K$ surrounding neural editable Gaussians can be determined as 
\begin{equation}
\vspace{-0.025in}
	\{\mathbf{x}^g_0, \ldots, \mathbf{x}^g_{K-1}\} = \mathbf{x}_a + \{\Delta \mathbf{x}^g_0, \ldots, \Delta \mathbf{x}^g_{K-1}\}\cdot\bm{l}^s.
\vspace{-0.025in}
\end{equation}
%\begin{equation}
%\mu_i = \mathbf{x}_a + \Delta \mathbf{x}^g_i \cdot l_s, \quad i = 0, \ldots, K.
%\end{equation}
For other attributes, we decode them with the anchor feature $\bm{f}^a$ and four lightweight MLPs \hot{(two layers with 50 hidden dimensions)}, denoted as $F_o$, $F_{\text{cov}}$, $F_{s}$, and $F_n$, responsible for decoding opacity $o$, covariance matrix ($\mathbf{s}$ and $\mathbf{q}$), shading attribute $\{\Delta\mathbf{c}, k^a, k^d, k^s, \beta \}$, and normal $\bm{n}$, respectively.
For example, given one camera position $\mathbf{x}_c$ and anchor position $\mathbf{x}_a$, the opacity values for $K$ local editable Gaussians can be calculated as 
\begin{equation}
\vspace{-0.025in}
    \{o_0, \ldots, o_{K-1}\} = F_{o}\left(\bm{f}^a, \mathbf{v}_{\text{CA}}, \bm{f}_{\text{TF}}\right), \ 
    \mathbf{v}_{\text{CA}} = 
    \frac{\mathbf{x}_a - \mathbf{x}_c}{\left\lVert \mathbf{x}_a - \mathbf{x}_c \right\rVert_2},
\label{eq:MLPinput}
\vspace{-0.025in}
\end{equation}
where $\mathbf{v}_{\text{CA}}$ denotes the normalized viewing direction from the camera to the anchor and $\bm{f}_{\text{TF}}$ is a TF embedding vector (refer to Section~\ref{subsec:jointTrain}).
The covariance matrix and shading attribute are similarly derived with $F_{\text{cov}}$, and $F_{s}$, except that the decoding results from $F_{\text{cov}}$ are further scaled by anchor scaling factors $\bm{l}^s$ and $\bm{l}^r$ to obtain the final scaling $\mathbf{s}$ and quaternion rotation $\mathbf{q}$.
After decoding all attributes, the final color of each neural editable Gaussian can be computed through the Blinn-Phong shading model, which is the sum of ambient ($\mathbf{c}_a$), diffuse ($\mathbf{c}_d$), and specular ($\mathbf{c}_s$) colors.
We assume the specular lighting color is white, given the light direction $\mathbf{d}$, offset color $\Delta\mathbf{c}$, and the viewing direction $\mathbf{v}_{\text{CG}}$ from the camera to each Gaussian primitive. These three color terms are decided as
\begin{subequations}
\label{eqn:Blinn-Phong}
\vspace{-0.025in}
\begin{align}
	\mathbf{c}_a&=k^a\left(\mathbf{c}_p+\Delta\mathbf{c}\right), \\
	\mathbf{c}_d&=k^d\left(\mathbf{c}_p+\Delta\mathbf{c}\right) |\mathbf{n} \cdot \mathbf{d}|,\\
	\mathbf{c}_s&=
	\begin{cases}
		k^s |\mathbf{n} \cdot \frac{\mathbf{v}_{\text{CG}}+\mathbf{d}}{|\mathbf{v}_{\text{CG}}+\mathbf{d}|}|^{\beta},	& \text{if } |\mathbf{n} \cdot \mathbf{d}| > 0 \\
		0,		& \text{otherwise} 
	\end{cases}
\end{align}
\vspace{-0.025in}
\end{subequations}
All neural-editable Gaussians are then rasterized in the same way as in 3DGS~\cite{Kerbl-TOG23}.
It is worth noting that, although attributes such as opacity are typically view-independent in other methods (e.g.,~\cite{Kerbl-TOG23, Mildenhall-ECCV20}), we include the viewing direction as an input to the decoding MLPs, as we find that the view-dependent neural editable Gaussians are more expressive and can achieve higher reconstruction accuracy (refer to Section~\ref{subsec:ablation}). 
During optimization, we prune and grow anchor points based on their gradients and the corresponding opacity values of editable Gaussians, \hot{following the densification strategy of Scaffold-GS~\cite{Lu-CVPR24}}.
\hot{Because the attribute values of the $K$ local editable Gaussians of each anchor are decoded on the fly from one compact anchor feature and shared lightweight MLPs, rather than each Gaussian storing its full attribute set explicitly, this design helps reduce parameter redundancy while preserving fast rendering.}

\vspace{-0.075in}
\subsection{Joint Learning}
\label{subsec:jointTrain}

\begin{figure}[!ht]
 \begin{center}
%\resizebox{0.9\linewidth}{!}{
$\begin{array}{c@{\hspace{0.05in}}c}
\includegraphics[height=1.85in]{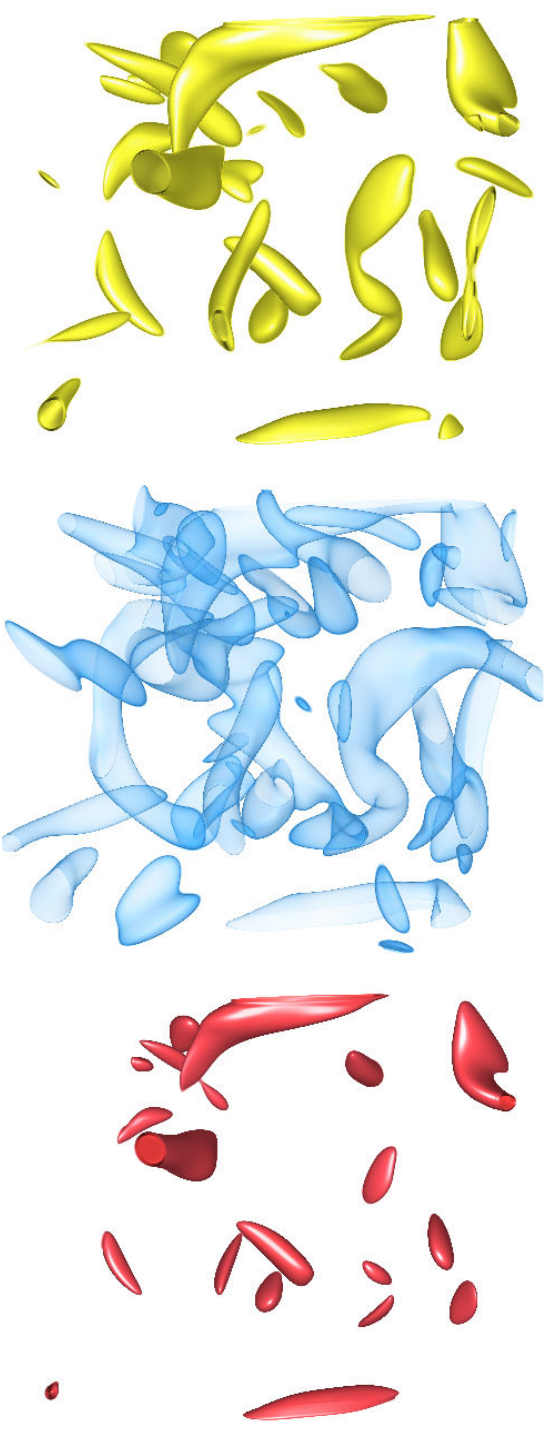}&
\includegraphics[height=1.85in]{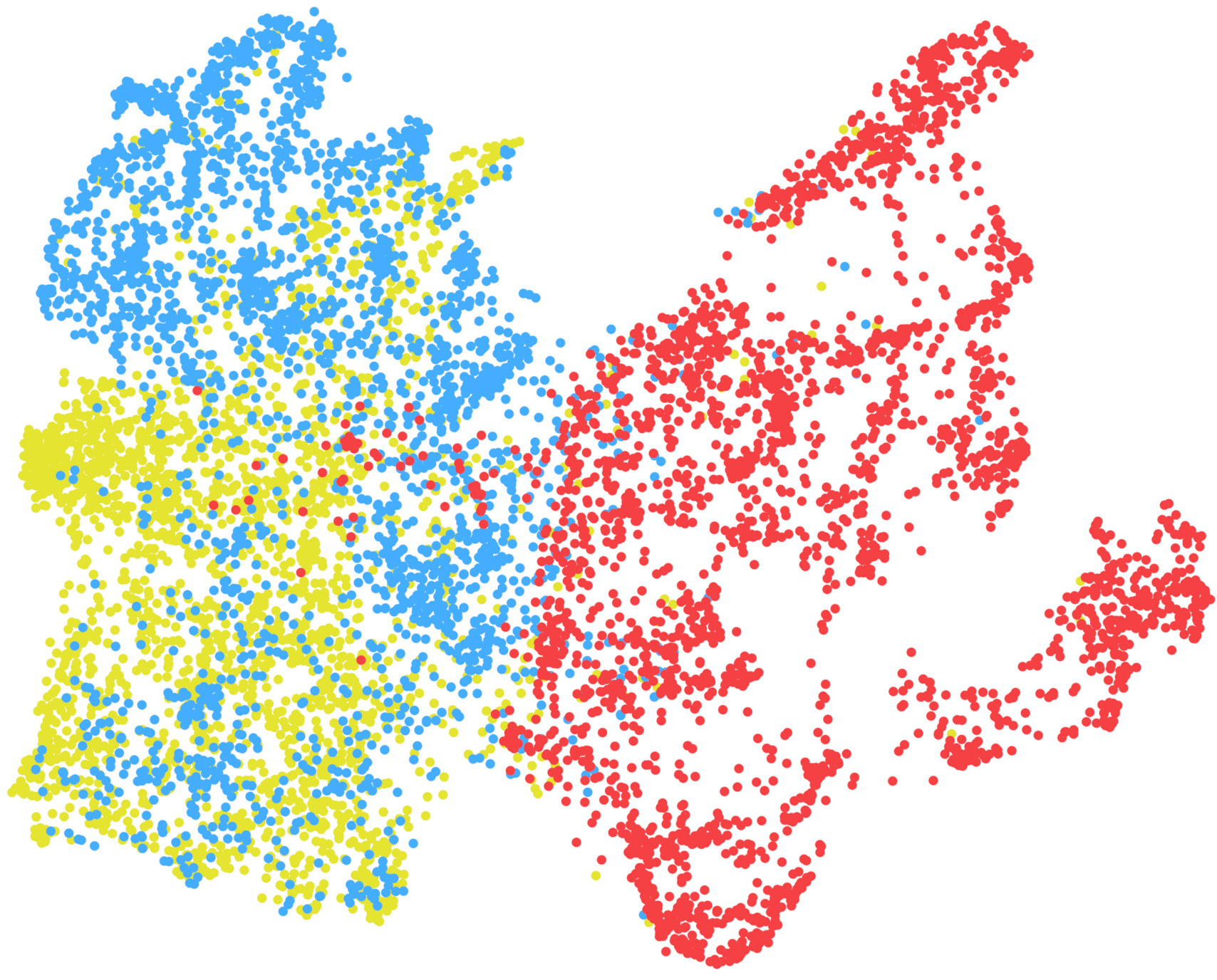}\\
\mbox{\small (a) basic scenes} &
\mbox{\small (b) anchor feature projection}\\
\end{array}$
%}
\end{center}
\vspace{-.25in} 
\caption{Visualization of inter-scene anchor feature similarity. (a) Three basic scenes from the vortex dataset. (b) UMAP projection of the corresponding anchor features.}
\label{fig:umap}
\vspace{-.1in}
\end{figure}

Given the formulation of neural editable Gaussians, when we train separate models to fit different VolVis scenes from the same volume data, we observe that the anchor features corresponding to scenes with \hot{similar spatial geometric distributions} tend to exhibit similar values.
As illustrated in Figure~\ref{fig:umap}, the anchor features of the yellow and blue basic scenes, which occupy similar spatial regions, show a significant degree of overlap in the 2D projection of corresponding anchor features.
This observation motivates us to treat the modeling of multiple VolVis scenes as a joint learning problem.
That is, we jointly optimize multiple related VolVis scene models with partial parameter sharing to exploit inter-scene correlations.
Our neural, editable Gaussian formulation keeps the anchor points independent within each basic scene while sharing MLPs across all jointly learned scenes.
In addition, when modeling multiple VolVis scenes with shared MLPs, to enable the networks to distinguish anchor features originating from different basic scenes, we create an additional learnable TF embedding feature $\bm{f}_{\text{TF}}\in\mathbb{R}^{50}$ for each jointly learned basic scene. 
This embedding is concatenated with the original inputs defined in Equation~\ref{eq:MLPinput} and jointly optimized during training.

%%--------------------------------------
%\begin{figure}[!htb]
% \begin{center}
%\includegraphics[width=1.0\linewidth]{figures/anchor_feat_project.pdf}\\
%\end{center}
%\vspace{-.15in} 
%\caption{\hot{caption goes here}}
%\label{fig:anchor-distribute}
%\end{figure}
%%--------------------------------------

%--------------------------------------
%--------------------------------------

% slightly discuss how we compose multiple ECoNGS, what's the benefit of modeling multiple basic scenes with one MLP
Given multiple basic scenes, we need a criterion to determine which scenes can be jointly learned for neural-editable Gaussians. Since color and lighting are decoupled via the shading attributes, the dominant factor affecting the parameters is geometry. We therefore use the sparse point clouds from Section~\ref{subsec:initialization} for clustering similar scenes.
Before optimization, we compute pairwise Chamfer distances between point clouds for all basic scenes and apply hierarchical clustering to group similar scenes for joint learning.
\hot{Although the number of pairwise comparisons grows quadratically with the number of basic scenes, downsampling each point cloud before computing the Chamfer distance makes this clustering step negligible compared to the overall training cost.}
By leveraging image data from multiple VolVis scenes during training, ECoNGS can achieve faster convergence and more accurate reconstruction results.
Meanwhile, parameter sharing in joint learning further reduces the overall model size and improves inference-time parallel efficiency.
% \hot{The benefit of joint learning grows with the number of basic scenes, reducing training time, the number of anchors, and model size by up to $2.2\times$, $2.5\times$, and $2.9\times$, respectively, with negligible loss in reconstruction quality, as analyzed across datasets of different scales in Appendix~\ref{sec:scalability} (Table~\ref{tab:joint-gain}).}

\vspace{-0.075in}
\subsection{Neural Entropy Coding with Context Model}
\label{subsec:InContextCompression}

\begin{figure}[!htb]
\begin{center}
\includegraphics[width=0.9\linewidth]{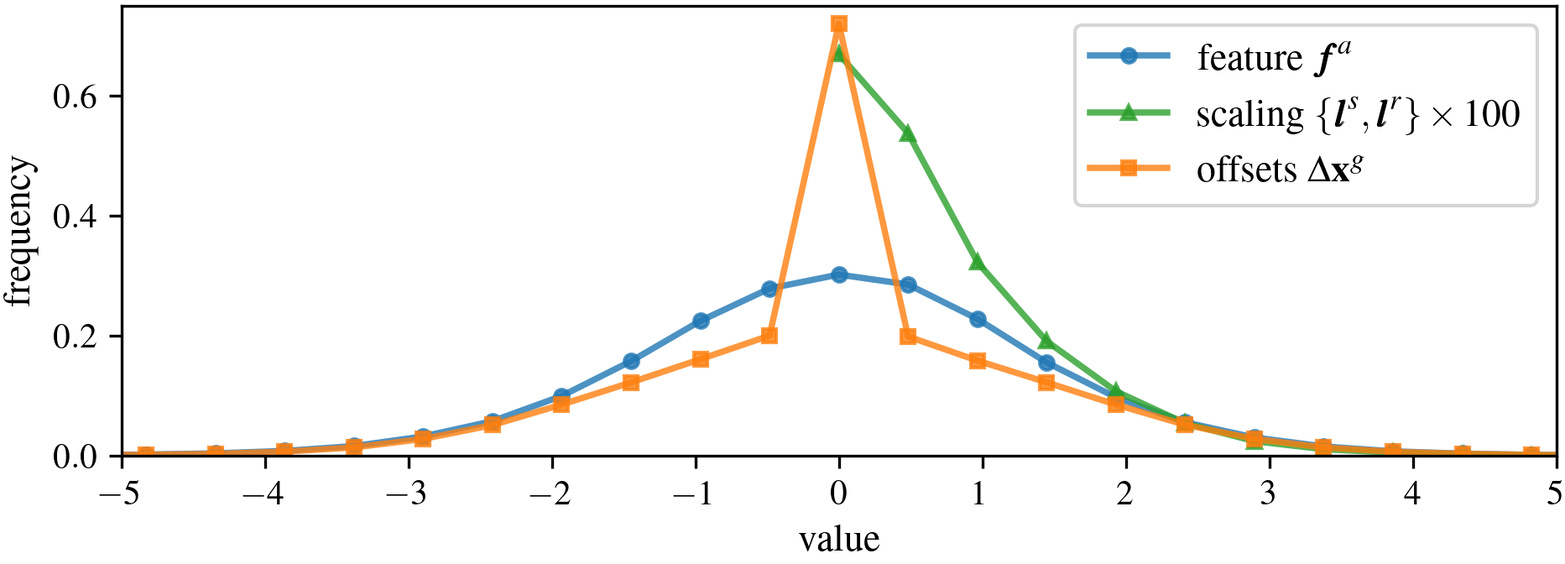}
\end{center}
\vspace{-.25in} 
\caption{Statistical analysis for the distribution of the anchor attribute values on the vortex dataset. All attributes demonstrate statistical Gaussian distributions. Note that we define the scaling factors $\bm{l}^s$ and $\bm{l}^r$ after the sigmoid activation, so only one side is shown, and we scale them by 100 for better visualization.}
\label{fig:anchor-distribute}
\vspace{-.1in}
\end{figure}

Although the hybrid structure and joint learning strategy of ECoNGS can effectively reduce redundant primitives, the remaining explicit anchor points can still consume substantial storage space.
Entropy coding is a promising technique for compressing these attributes by exploiting their statistical redundancy, thereby further improving compactness.
%To further improve the compactness of ECoNGS, we apply entropy coding to compress explicit anchor attributes $\bm{A}$.
%, where entropy coding aims to assign shorter codes to more frequently occurring symbols based on the statistical distribution, thereby achieving near-optimal lossless compression.
Figure~\ref{fig:anchor-distribute} shows the statistical distribution of $\bm{A}$ after fitting VolVis scenes from the vortex dataset, where all components exhibit approximately Gaussian-like distributions.
Based on the probability distributions, entropy coding employs an entropy encoder (e.g., arithmetic encoding~\cite{ArithmeticEncoding}) to represent different values with variable-length bitstreams in a lossless manner.

Despite its effectiveness, conventional entropy coding relies on fixed probability distributions, which limit its flexibility and often lead to suboptimal compression ratios. 
Therefore, we opt for neural entropy coding~\cite{Cheng-CVPR20, Li-NeurIPS21, Chen-ECCV24}. 
In particular, we leverage an auxiliary neural network, the \textit{context model}, to simultaneously learn the distribution of anchor parameters during training and to provide the entropy coder with probability estimates at both encoding and decoding times. 
In practice, we adopt BiRF~\cite{Shin-NeurIPS23} as our context model architecture, which incorporates multiple binary hash grids with parameter values either +1 or -1 and a shallow MLP for decoding the aggregated hash feature.
The context model takes the anchor position $\mathbf{x}_a$ as input and outputs a set of values that model the probability distribution for the anchor attributes. %, which will be detailed in the following paragraph. Note that 
The context model is shared across all VolVis scenes during joint learning to exploit inter-scene correlations, as with lightweight MLPs.
%At encoding time, the context model provides the entropy coder with probability estimations, thereby enabling end-to-end optimization and improved compression rate.

%--------------------------------------
%--------------------------------------

% introduce quantization
The entropy coder can only encode discrete values. 
Therefore, we must quantize continuous attribute values to finite sets before processing. %feeding them into the entropy coder.
To enable backpropagation, we implement the quantization process by injecting noise during training and rounding during inference, as described in~\cite{Balle-ICLR18, Li-NeurIPS21, Chen-ECCV24}.
Given $i$-th anchor $\bm{A}_i$ located at $\mathbf{x}_a$, for any component $\bm{f}_i\in\{\bm{f}^a_i, \bm{l}^s_i, \bm{l}^r_i, \Delta\mathbf{x}^{g}_i\}$ within the anchor, the quantization process can be defined as 
\begin{align}
\vspace{-0.025in}
    \hat{\bm{f}}_i &= \bm{f}_i + \mathcal{U}\!\left(-\tfrac{1}{2}, \tfrac{1}{2}\right) \times \mathbf{q}_i,
    && \text{for training} \label{eq:quantization_train} \\
    &= \operatorname{round}(\bm{f}_i / \mathbf{q}_i) \times \mathbf{q}_i, 
    && \text{for testing} \label{eq:quantization_test}
\vspace{-0.025in}
\end{align}
where $\mathcal{U}$ represents uniform sampling, and we compute the actual quantization step size $\mathbf{q}_i$ as
\begin{equation}
\vspace{-0.025in}
	\mathbf{q}_i = Q \times \left(1 + \tanh(\mathbf{r}_i)\right),
\vspace{-0.025in}
\end{equation}
where $\mathbf{r}_i$ is a refinement factor output from the context model for adaptively adjusting the predefined quantization step $Q$.
Empirically, $Q$ varies across components due to their different numerical scales. We set $Q$ as 1, 0.001, 0.001, and 0.2 for attributes $\bm{f}^a$, $\bm{l}^s$, $\bm{l}^r$, and $\Delta\mathbf{x}^{g}$, respectively.
After quantization during training, the quantized anchor attribute $\hat{\bm{f}}_i$ substitutes the original ones to achieve quantization-aware training.
% introduce probability modeling
When modeling the probability $p(\hat{\bm{f}}_i)$, as all attributes in $\bm{A}$ exhibit a Gaussian-like distribution, the context model outputs individual mean $\bm{\mu}_i$ and variance $\bm{\sigma}_i$ to formulate respective Gaussian distributions $\mathcal{N}(\bm{\mu}_i, \bm{\sigma}_i)$.
After modeling $p(\hat{\bm{f}}_i)$, the bit consumption for all anchor attributes can be computed and optimized according to their entropy.
We define the entropy loss as
\begin{equation}
\vspace{-0.025in}
L_{\text{entropy}} = \sum_{\bm{f} \in \{\bm{A}\}} \sum_{i=1}^{N^a} \left(-\log_{2} p(\hat{\bm{f}}_{i})\right),
\label{eq:entropy_loss}
\vspace{-0.025in}
\end{equation}
where $N^a$ denotes the number of anchors.
The hash grid component in the context model can also be optimized for compression.
Similarly, we can minimize $L_{\text{hash}}$ to reduce the bit consumption of the binary hash grids
\begin{equation}
\vspace{-0.025in}
L_{\text{hash}} =
M_{+} \, (-\log_{2}(h_{f})) +
M_{-} \, (-\log_{2}(1 - h_{f})),
\vspace{-0.025in}
\end{equation}
where $h_f$ is the occurrence frequency of +1 value, $M_{+}$ and $M_{-}$ correspond to the total numbers of +1 and -1 in the hash grids.
Reducing entropy and hash loss encourages distribution of values within the anchor attribute and a more concentrated hash grid, resulting in less bit consumption after encoding at the cost of reconstruction fidelity.

%--------------------------------------
\begin{table}[htb]
\centering
\caption{Volume datasets for univariate scenes.}
\vspace{-0.1in}
\label{tab:univariate-datasets}
\resizebox{\linewidth}{!}{%
%\resizebox{\columnwidth}{!}{
%\setlength{\tabcolsep}{5pt}
\begin{tabular}{lccccc}
%\toprule
             &    resolution              & volume size & \# basic  & CPU/GPU  \\
dataset & $x \times y \times z$ & (GB)            & scenes &  memory (GB)  \\
\hline %\midrule
argon bubble & 2,560$\times$1,024$\times$1,024 & 10.0  & 2   &  26.1/14.3    \\
combustion (MF) & 1,920$\times$2,880$\times$480 & 9.9  & 9  &  24.3/12.4    \\
ionization (T)  & 2,400$\times$992$\times$992  & 8.8  & 4  &  22.6/12.1   \\
supernova       & 864$\times$864$\times$864   & 2.4  & 4  &  7.5/3.9    \\
beetle          & 864$\times$864$\times$494   & 1.3  & 1  &  3.4/2.4    \\
vortex          & 512$\times$512$\times$512   & 0.5  & 4  &  3.0/1.4    \\
%\bottomrule
\end{tabular}
}
\vspace{-.1in}
\end{table}
%--------------------------------------

%--------------------------------------
\begin{table}[htb]
\centering
\caption{Volume datasets for multivariate scenes.}
\vspace{-0.1in}
\label{tab:multivariate-datasets}
\resizebox{\linewidth}{!}{%
%\resizebox{\columnwidth}{!}{
%\setlength{\tabcolsep}{5pt}
\begin{tabular}{lcccc}
%\toprule
             & resolution                 & variable    & \# basic  & CPU/GPU   \\
dataset & $x \times y \times z$ &  names  & scenes & memory (GB)  \\
\hline %\midrule
combustion &1,920$\times$2,880$\times$480 & CHI, HR, MF    & 41   & 24.3/12.4\\
 & & VORT, YOH    &   & \\ \hline
ionization &2,400$\times$992$\times$992  & H, H+, He, He+   & 24 &22.6/12.1 \\
&   & H2, PD, T   &  & \\ \hline
Tangaroa   & 600$\times$360$\times$240  & ACC, DIV          & 8 & 2.6/2.1  \\
 &  & VLM, VTM          & &   \\
%\bottomrule
\end{tabular}
}
\vspace{-.1in}
\end{table}
%--------------------------------------

% %-------------------------------
\begin{table*}[htb]
\caption{Comparison of average PSNR (dB), LPIPS, and rendering framerate (FPS), as well as total training time (TT, in minutes) and model size (MS, in MB) on four datasets. The best and second-best results are highlighted in bold and underline, respectively.}
\vspace{-0.1in}
\centering
\label{tab:univariate-baseline}
\resizebox{0.9\textwidth}{!}{%
\begin{tabular}{l l c c c c c | l l c c c c c}
%\toprule
dataset & method & PSNR $\uparrow$ & LPIPS $\downarrow$ & FPS $\uparrow$ & TT $\downarrow$ & MS $\downarrow$ &
dataset & method & PSNR $\uparrow$ & LPIPS $\downarrow$ & FPS $\uparrow$ & TT $\downarrow$ & MS $\downarrow$ \\

\hline

\multirow{7}{*}{\shortstack{combustion\\(MF)}}
& Plenoxels     & 30.28 & 0.055 & 25.83 & 40.88 & 2,596.10 &
\multirow{7}{*}{\shortstack{ionization\\(T)}}
& Plenoxels     &29.24  &0.069 & 25.81 &17.43  &1,231.72  \\

& 3DGS         & \underline{33.86} & \textbf{0.021} & \textbf{741.68} & \underline{23.15} & 328.50 &
& 3DGS         &32.43  &0.024 & \textbf{701.05} &\underline{10.83}  &140.34  \\

& Scaffold-GS  & 32.56 & 0.024 & 536.68 & 31.68 & 91.78 &
& Scaffold-GS  & 30.99 &0.028 & 464.17 &17.35  &55.51  \\

& CCNeRF       & 29.91 & 0.046 & 0.77 & 170.75 & 24.55 &
& CCNeRF       &30.80  &0.038 & 0.75 &74.28  &10.29  \\

& HAC          & 32.93 & 0.024 & 339.28 & 80.83 & \underline{11.43} &
& HAC          & 31.73 & 0.027 & 386.95 &38.53  &\underline{6.25}  \\

& iVR-GS       & 33.35 & \underline{0.023} & 133.96 & 103.32 & 28.02 &
& iVR-GS       & \underline{33.43} & \underline{0.021} & 124.89 & 46.33  &17.41  \\

& ECoNGS  & \textbf{34.70} & \textbf{0.021} & \underline{591.46} & \textbf{18.04} & \textbf{4.94} &
& ECoNGS  &\textbf{34.34}  &\textbf{0.019} & \underline{548.81} &\textbf{10.36}  &\textbf{3.32}  \\

\hline

\multirow{7}{*}{supernova}
& Plenoxels     &28.65&0.096& 25.92 &20.39&1,260.84 &
\multirow{7}{*}{vortex}
& Plenoxels     & 31.37 & 0.054 & 27.13 & 18.10 & 1,144.54 \\

& 3DGS         &\underline{30.77}&0.045& \textbf{681.48} &\textbf{11.62}&166.59 &
& 3DGS         & 33.98 & \underline{0.020} & \textbf{704.67} & \underline{10.33} & 126.92 \\

& Scaffold-GS  &29.54&0.050& 475.27 &16.33&56.61 &
& Scaffold-GS  & 27.14 & 0.065 & 515.62 & 14.82 & 46.38 \\

& CCNeRF       &27.91&0.093& 0.62 &96.37&10.81 &
& CCNeRF       & 31.21 & 0.055 & 0.60 & 75.03 & 10.28 \\

& HAC          &30.01&0.048& 404.61 &39.93&\underline{7.44} &
& HAC          & 26.74 & 0.071 & 352.81 & 36.05 & \underline{4.87} \\

& iVR-GS       &30.52&\underline{0.044}& 128.96 &47.23&17.07&
& iVR-GS       & \underline{34.54} & 0.021 & 171.45 & 45.83 & 11.18 \\

& ECoNGS  &\textbf{31.55}&\textbf{0.043}& \underline{514.47} &\underline{13.83}&\textbf{4.40} &
& ECoNGS  & \textbf{36.74} & \textbf{0.018} & \underline{590.21} & \textbf{9.32} & \textbf{1.83} \\

%\bottomrule
\end{tabular}}
\vspace{-.1in}
\end{table*}
% %-------------------------------------- 

\vspace{-0.075in}
\subsection{Optimization and Coding}
\label{subsec:optimization}

ECoNGS optimization is conducted in two stages, with each stage trained for 10,000 iterations. 
For the first stage, we only optimize the anchor point attributes and lightweight MLPs, which is formulated as
\begin{equation}
\vspace{-0.025in}
L_{\text{stage1}} = L_{\text{rec}} + \lambda_s \sum^{N^g}_j \text{prod}\left(\mathbf{s}_j\right),
\label{eq:stage1-loss}
\vspace{-0.025in}
\end{equation}
where $N^g$ denotes the number of predicted editable Guassians and $\text{prod}(\cdot)$ is the product operation for values within a vector. The first term $L_{\text{rec}}$ is an image reconstruction loss that combines L1 and SSIM; the second term is a scaling regularization term that encourages the primitives to be as small as possible to avoid model capacity loss caused by too many primitives overlapping.
The regularization term is controlled with $\lambda_s$, and we set it to 0.001 in our experiments.
Once the distribution of anchor attribute values stabilizes, we update all parameters, including the context model, for the remaining iterations. The overall loss for the second stage can be defined as
\begin{equation}
\vspace{-0.025in}
L_{\text{stage2}} = L_{\text{stage1}} + \lambda_e (L_{\text{entropy}} + L_{\text{hash}})/(N^a \times d_{\bm{A}}),
\label{eq:stage2-loss}
\vspace{-0.025in}
\end{equation}
where $d_{\bm{A}}$ is the sum of all anchor attribute dimensions and $\lambda_e$ are trade-off hyperparameters to balance reconstruction fidelity and model compactness.
\hot{To stabilize the early optimization, we turn off the uniform noise of Equation~\ref{eq:quantization_train} during the first 3,000 iterations and enable it afterward.}

At the encoding/decoding phase, we first encode/decode the context model. 
Then the anchor attributes are encoded into a binary bitstream using an arithmetic encoder based on probabilities estimated by the context model.
For all MLPs, we store their parameters as 32-bit floating-point values.

%\subsection{Interactive Interface}

\vspace{-0.075in}
%--------------------------------------
\begin{figure*}[!t]
 \begin{center}
%\resizebox{\textwidth}{!}{
$\begin{array}{c@{\hspace{0.05in}}c@{\hspace{0.05in}}c@{\hspace{0.05in}}c@{\hspace{0.05in}}c}
\includegraphics[height=1.675in]{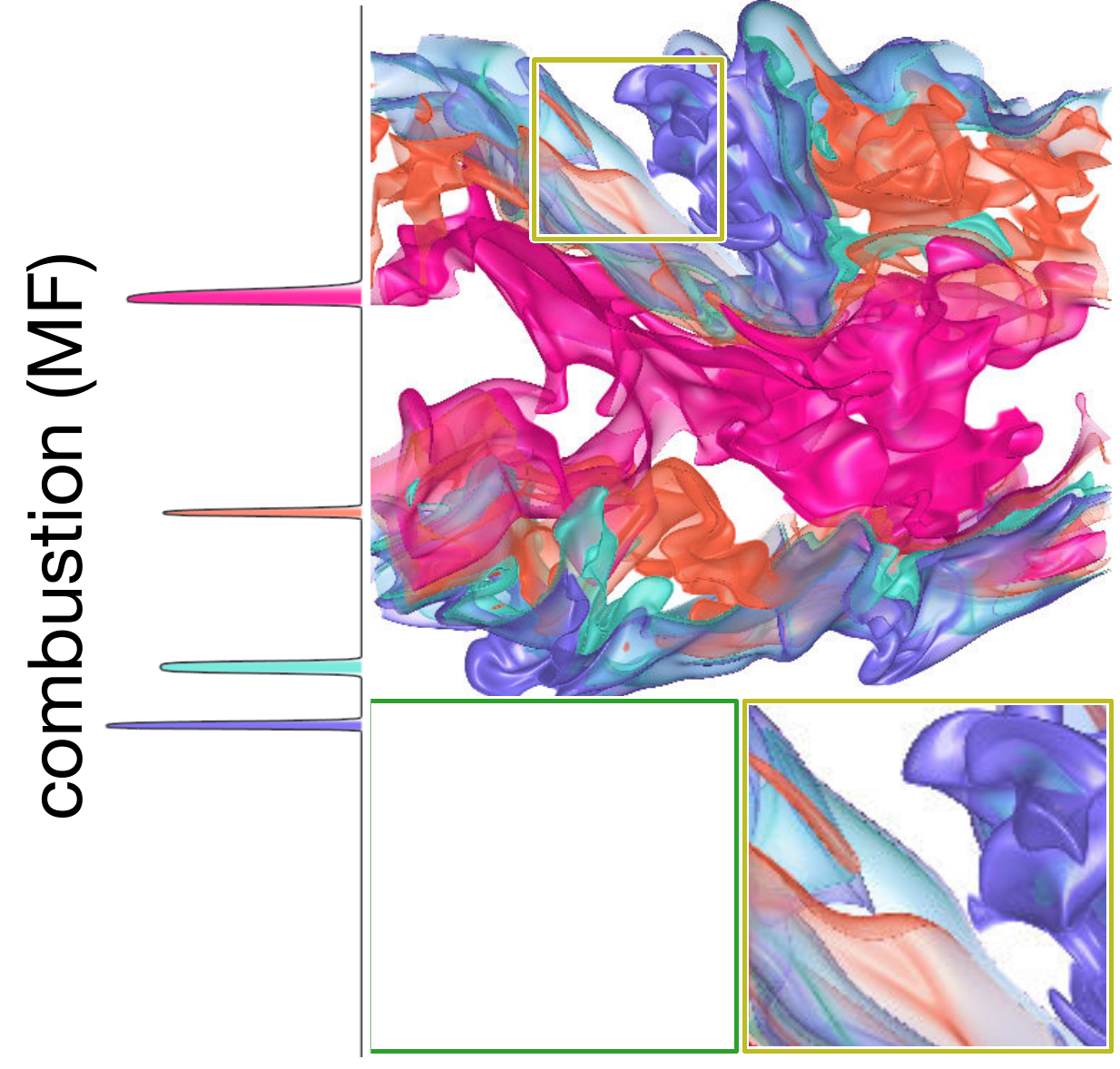}&
\includegraphics[height=1.675in]{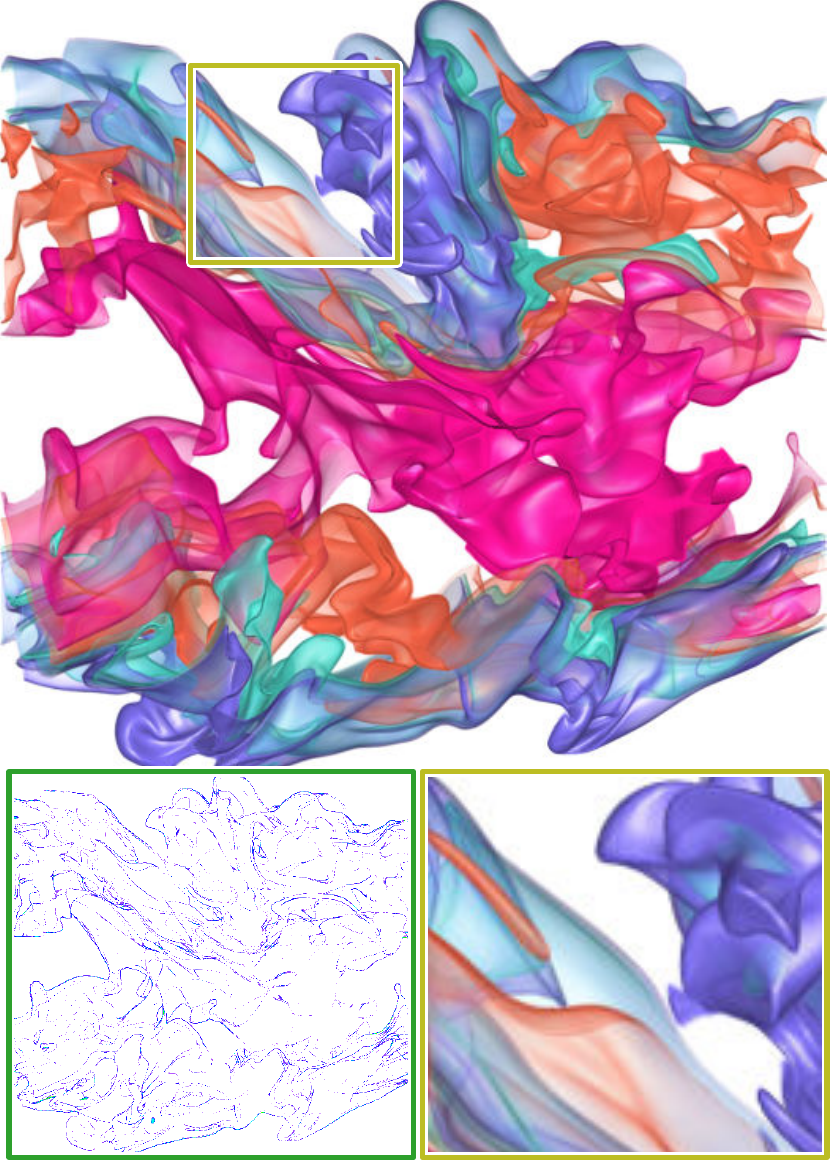}&
\includegraphics[height=1.675in]{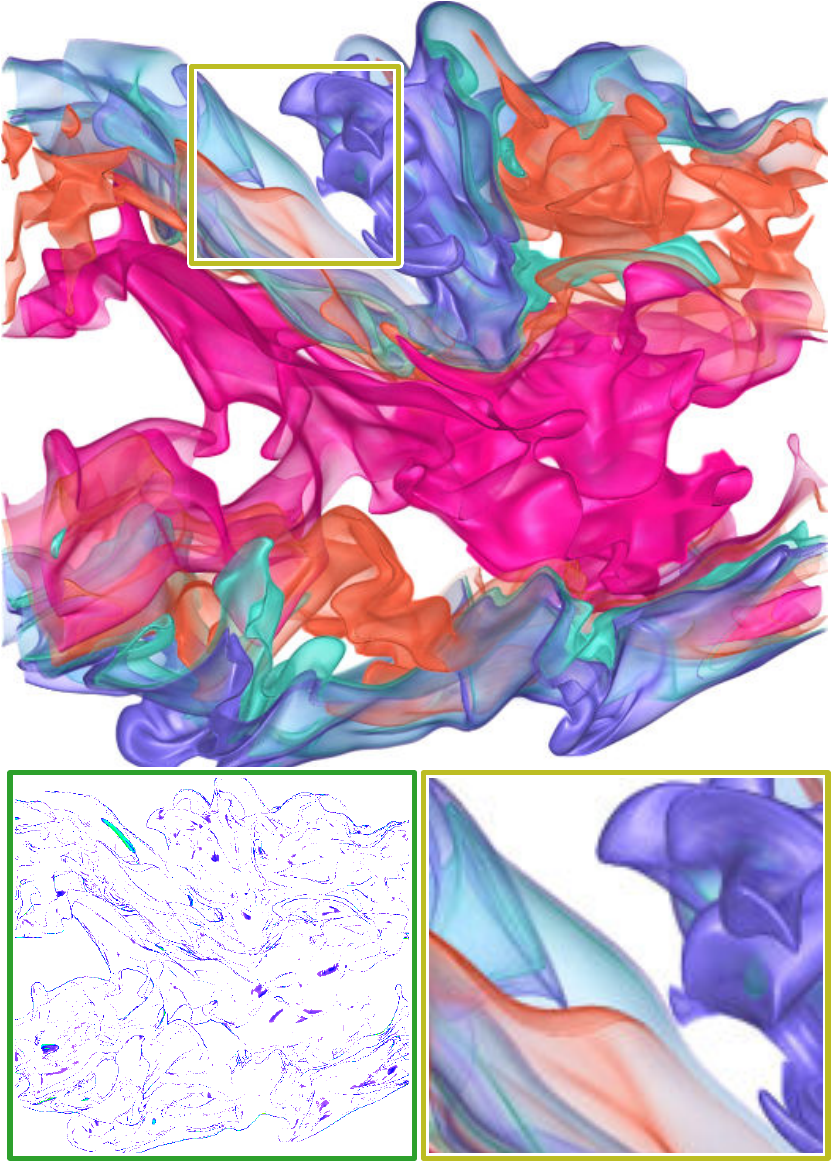}&
\includegraphics[height=1.675in]{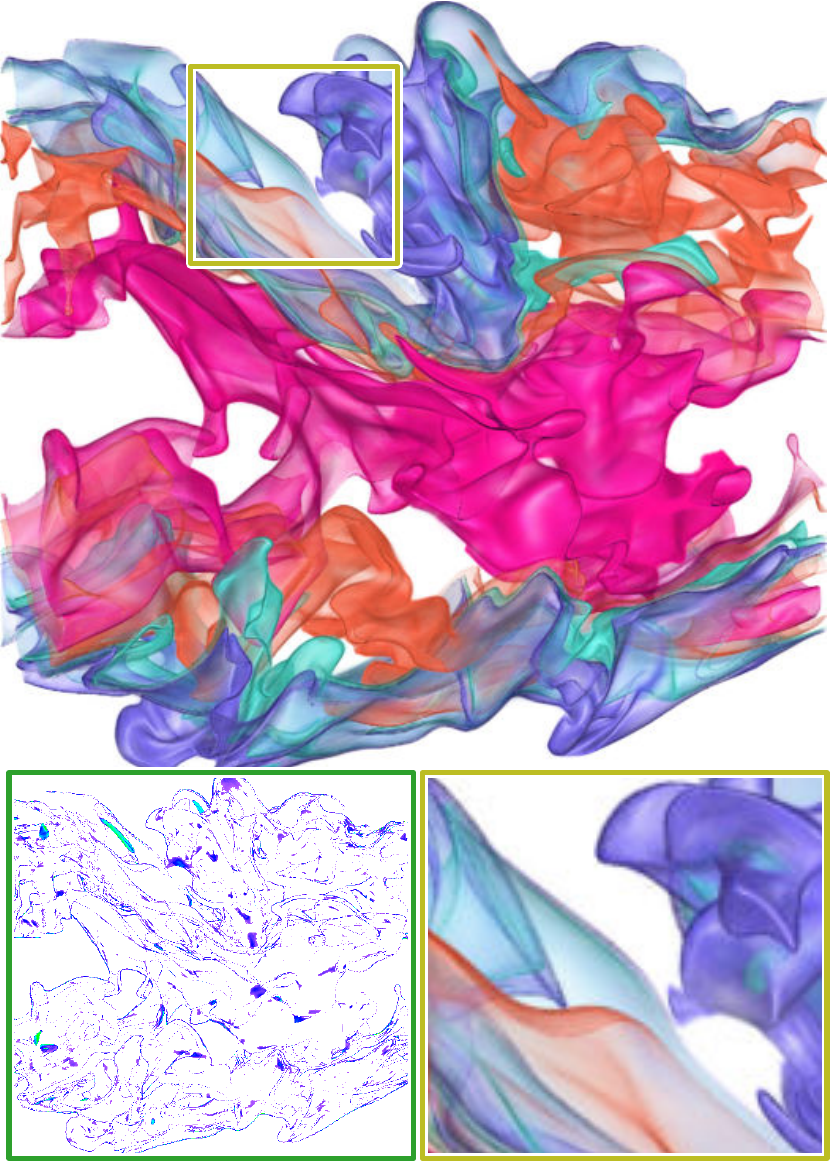}&
\includegraphics[height=1.675in]{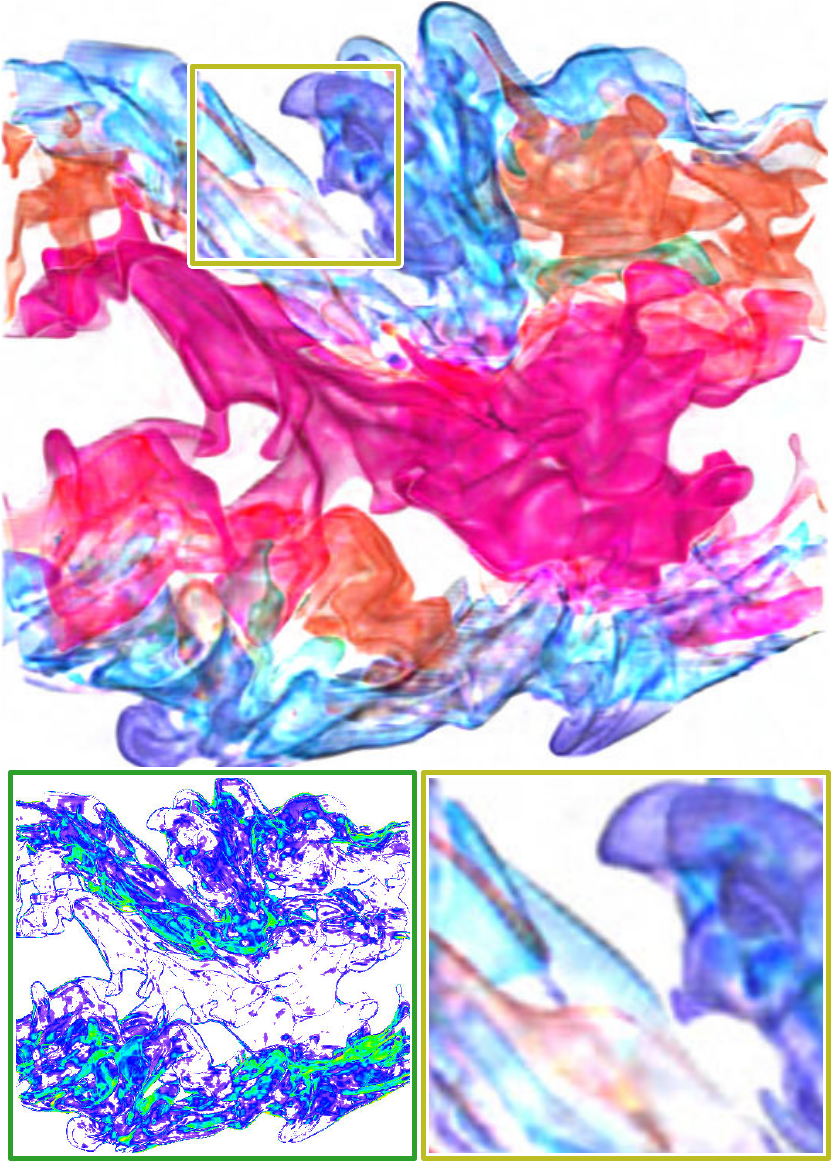}\\

\includegraphics[height=1.6in]{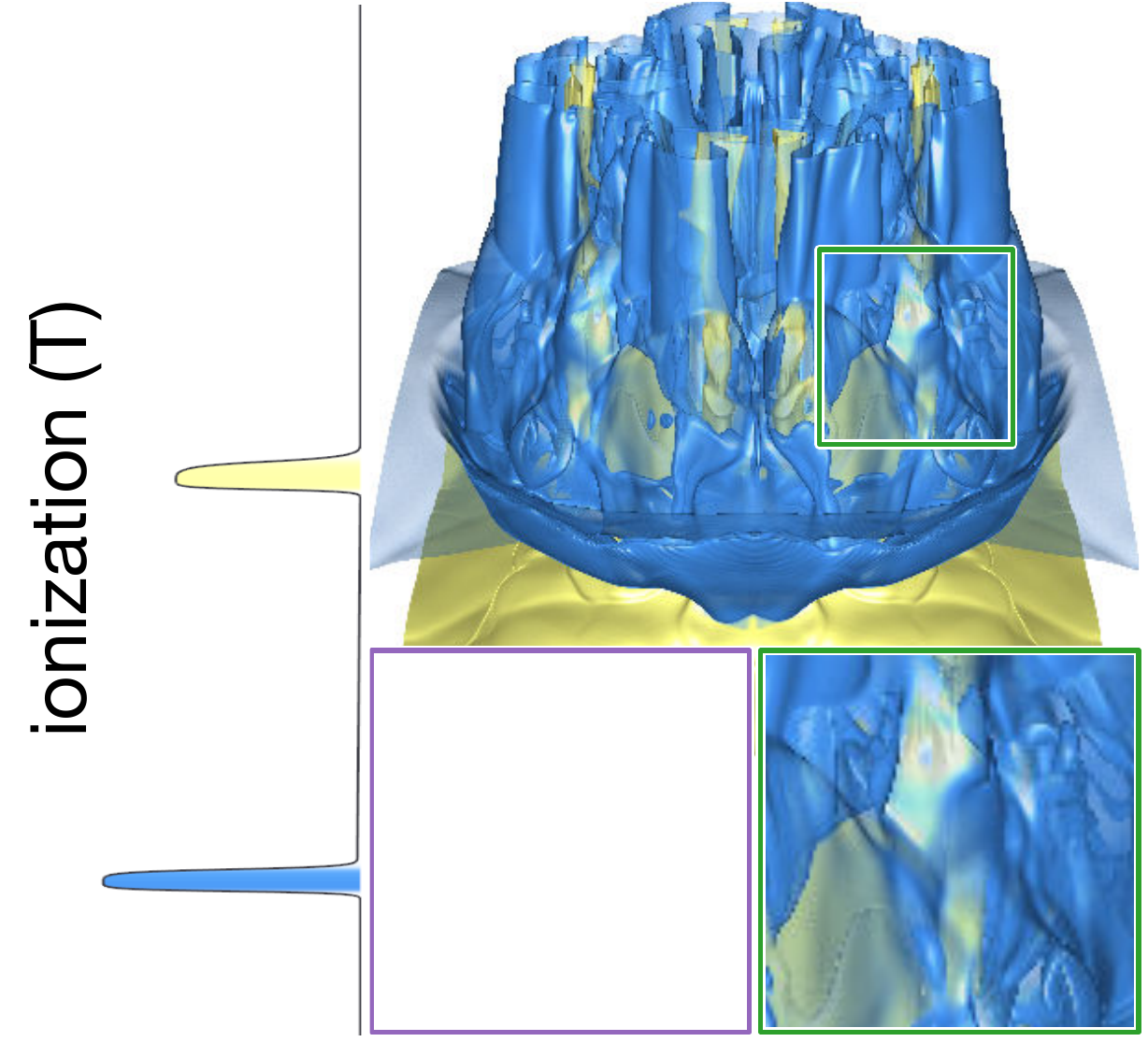}&
\includegraphics[height=1.6in]{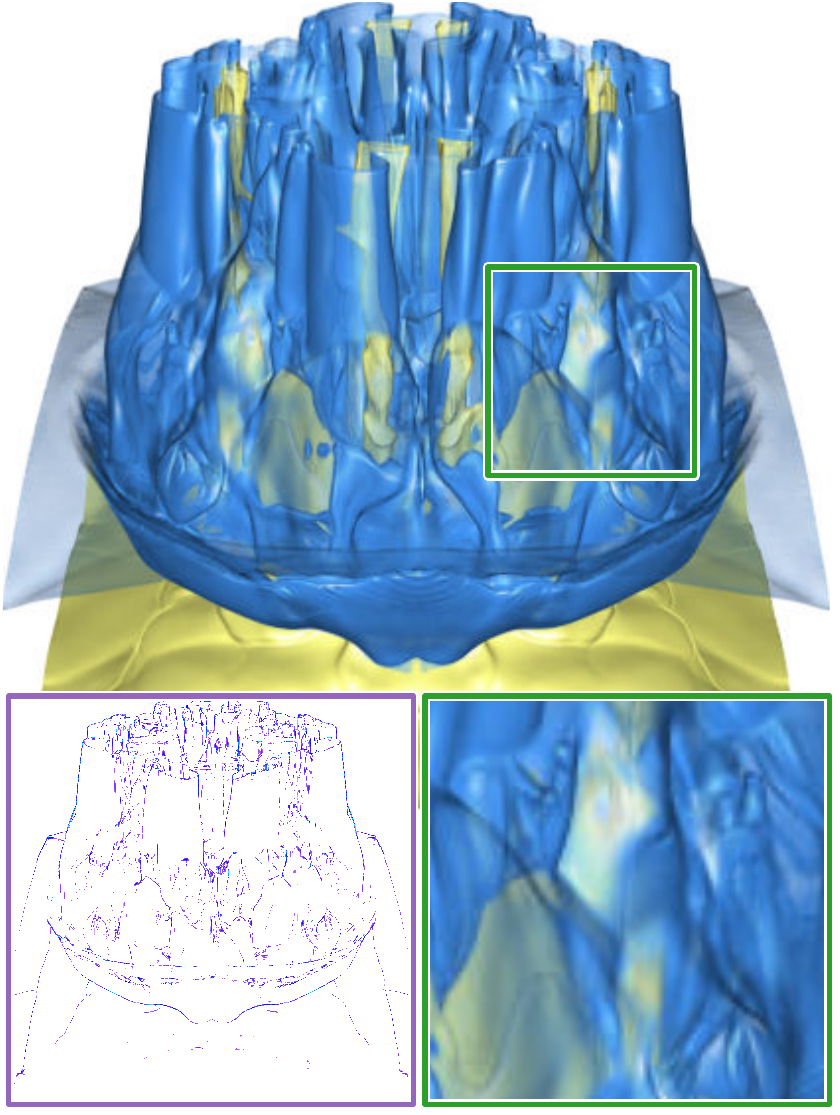}&
\includegraphics[height=1.6in]{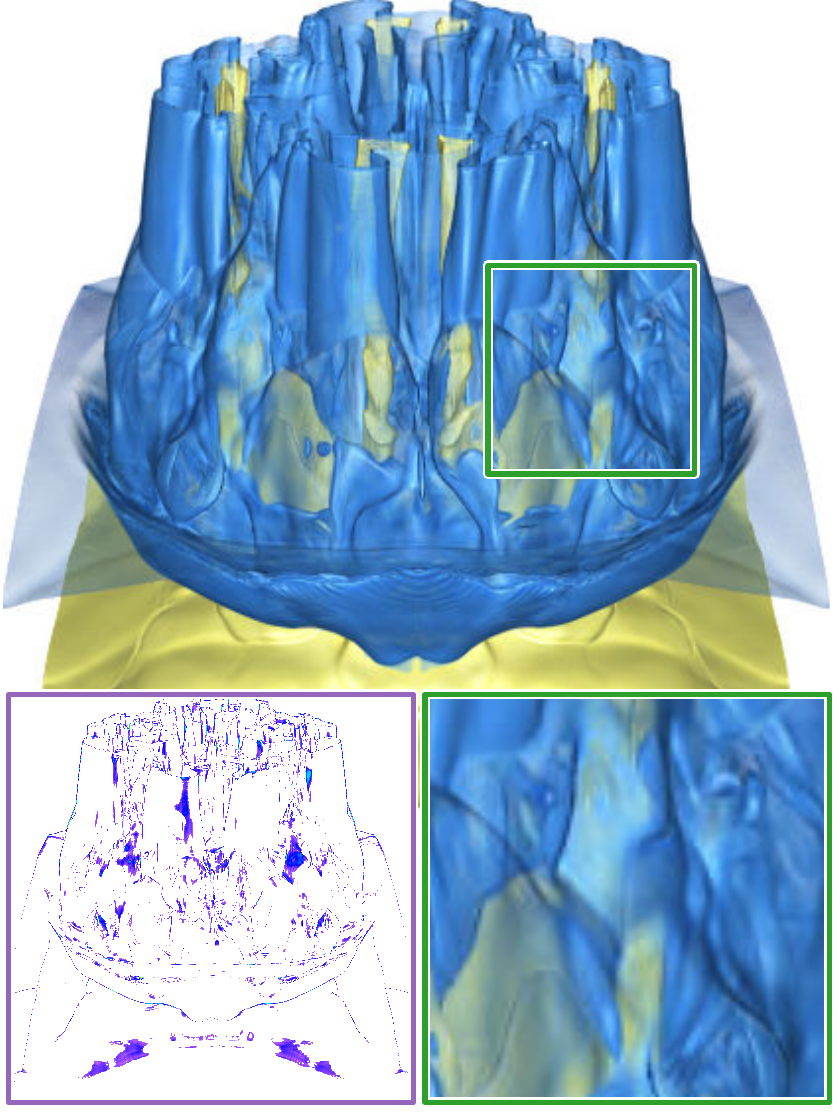}&
\includegraphics[height=1.6in]{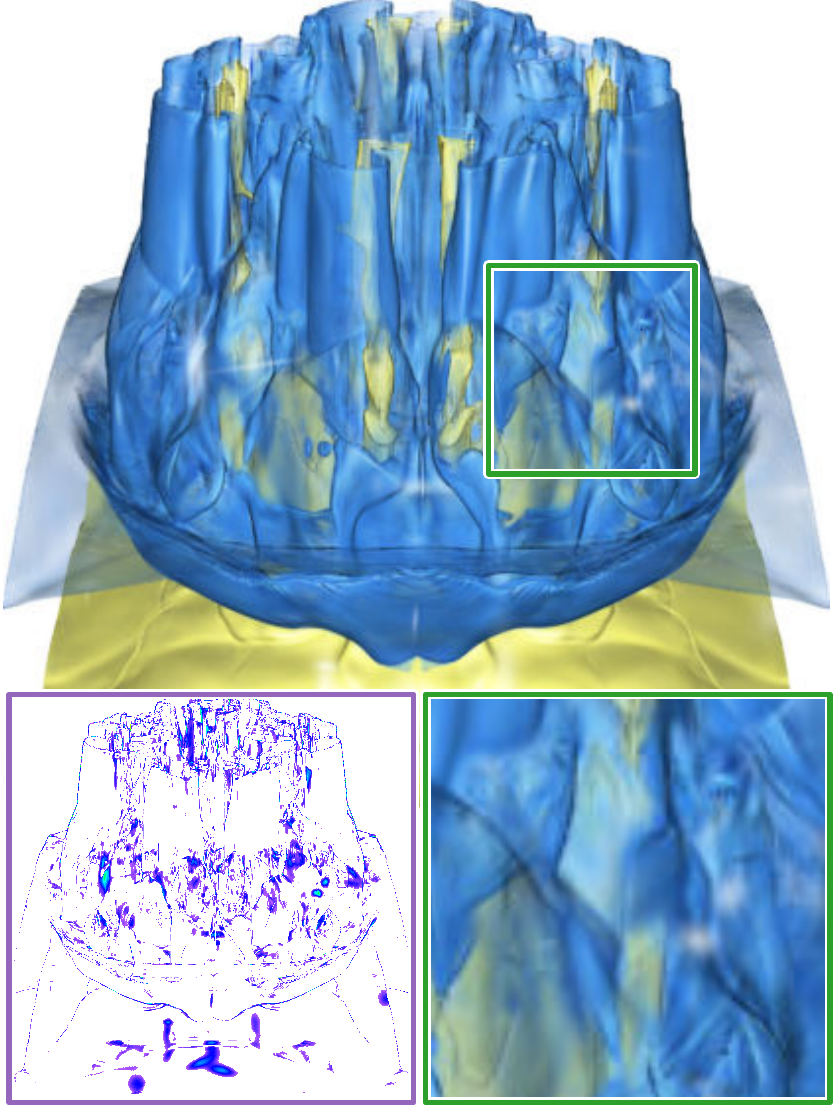}&
\includegraphics[height=1.6in]{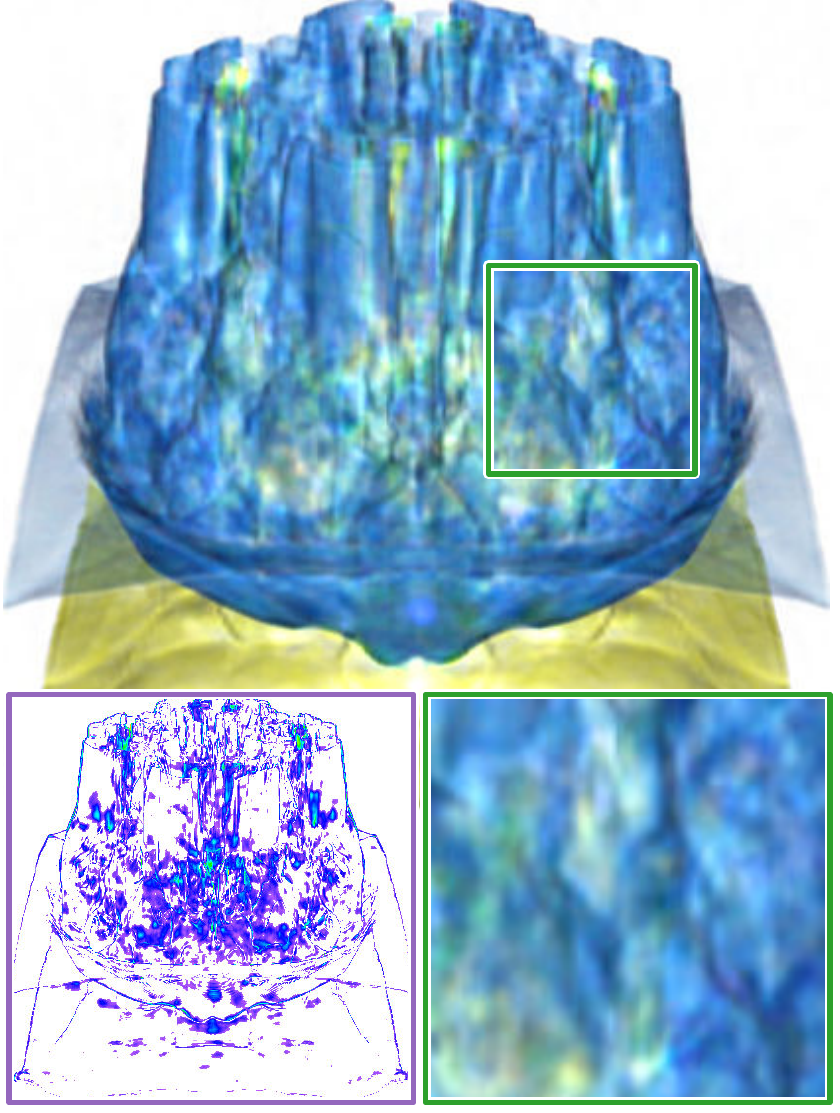}\\

\includegraphics[height=1.85in]{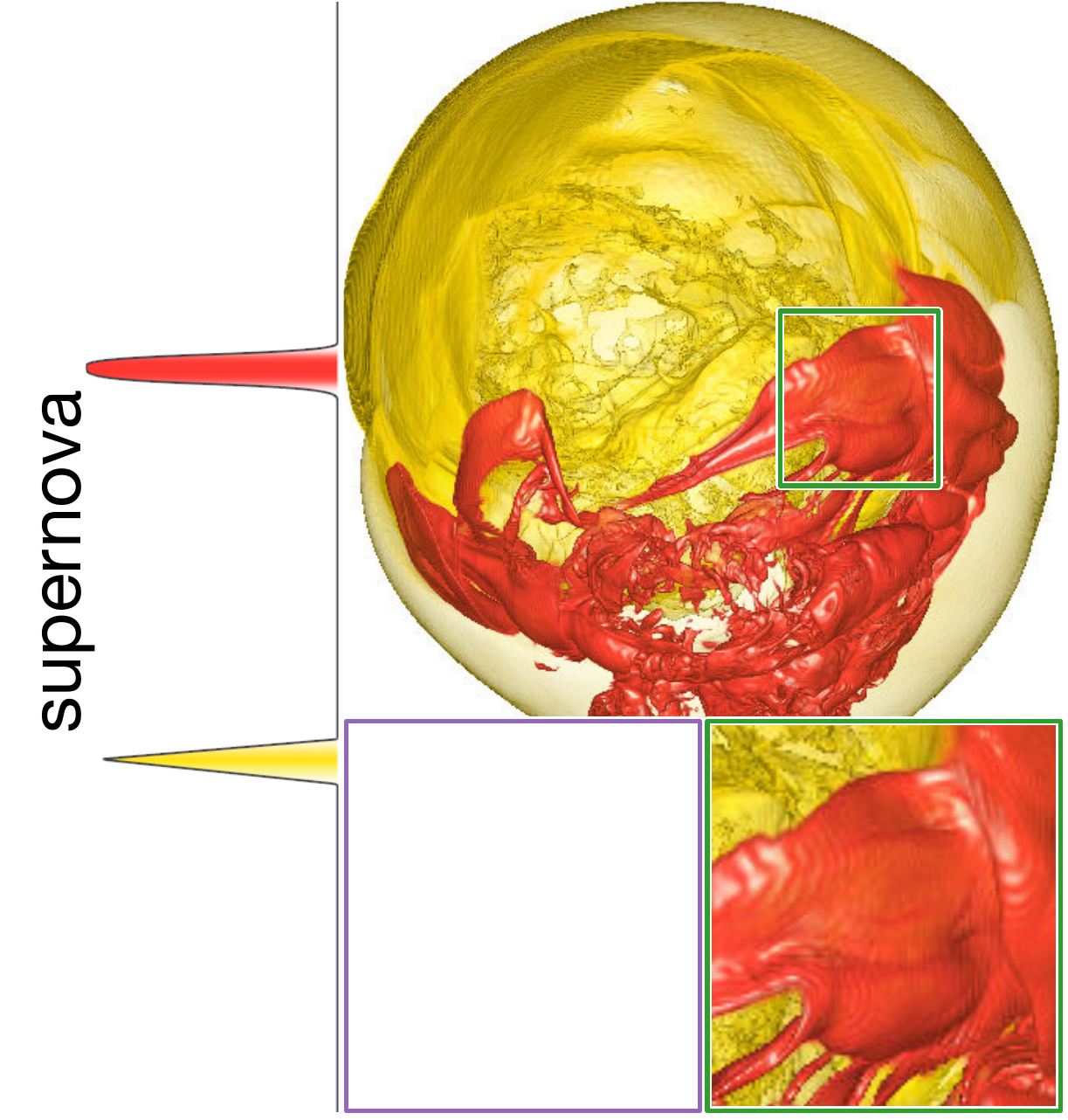}&
\includegraphics[height=1.85in]{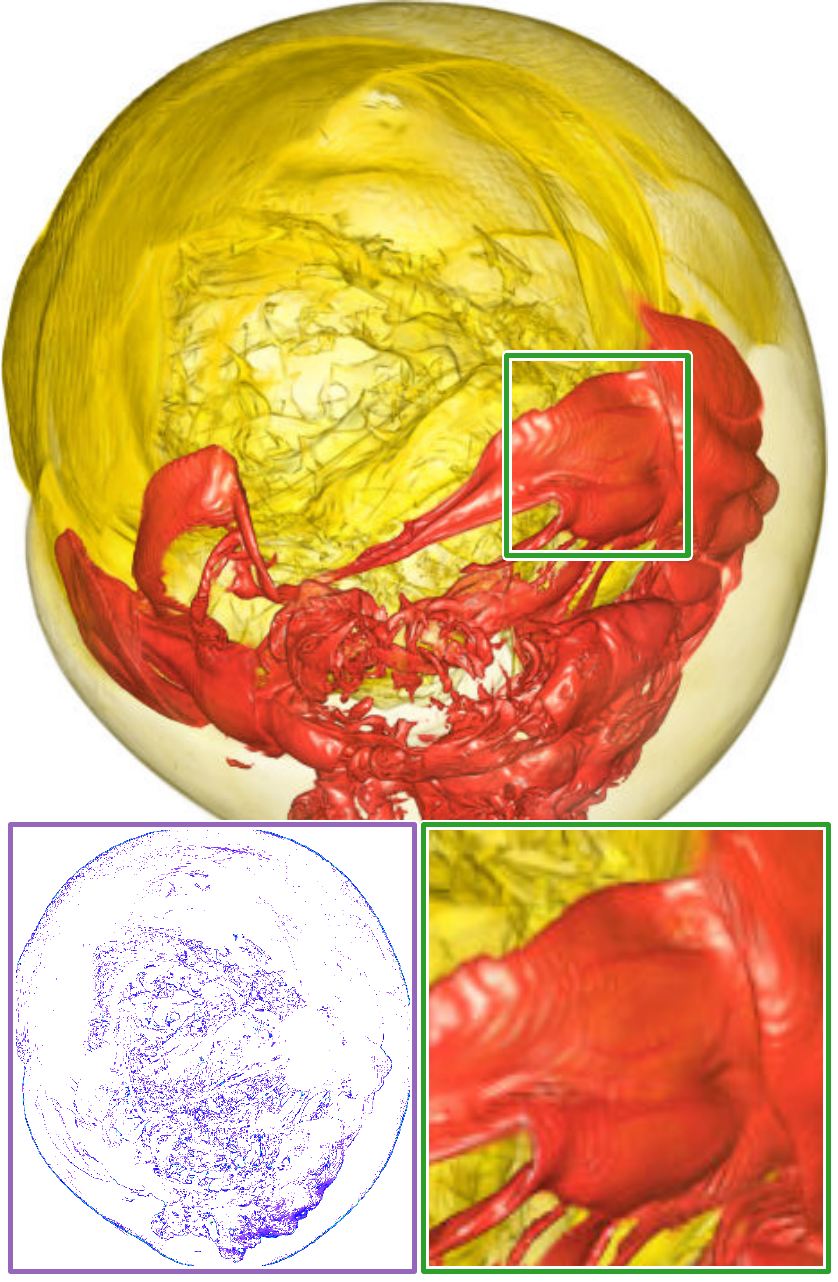}&
\includegraphics[height=1.85in]{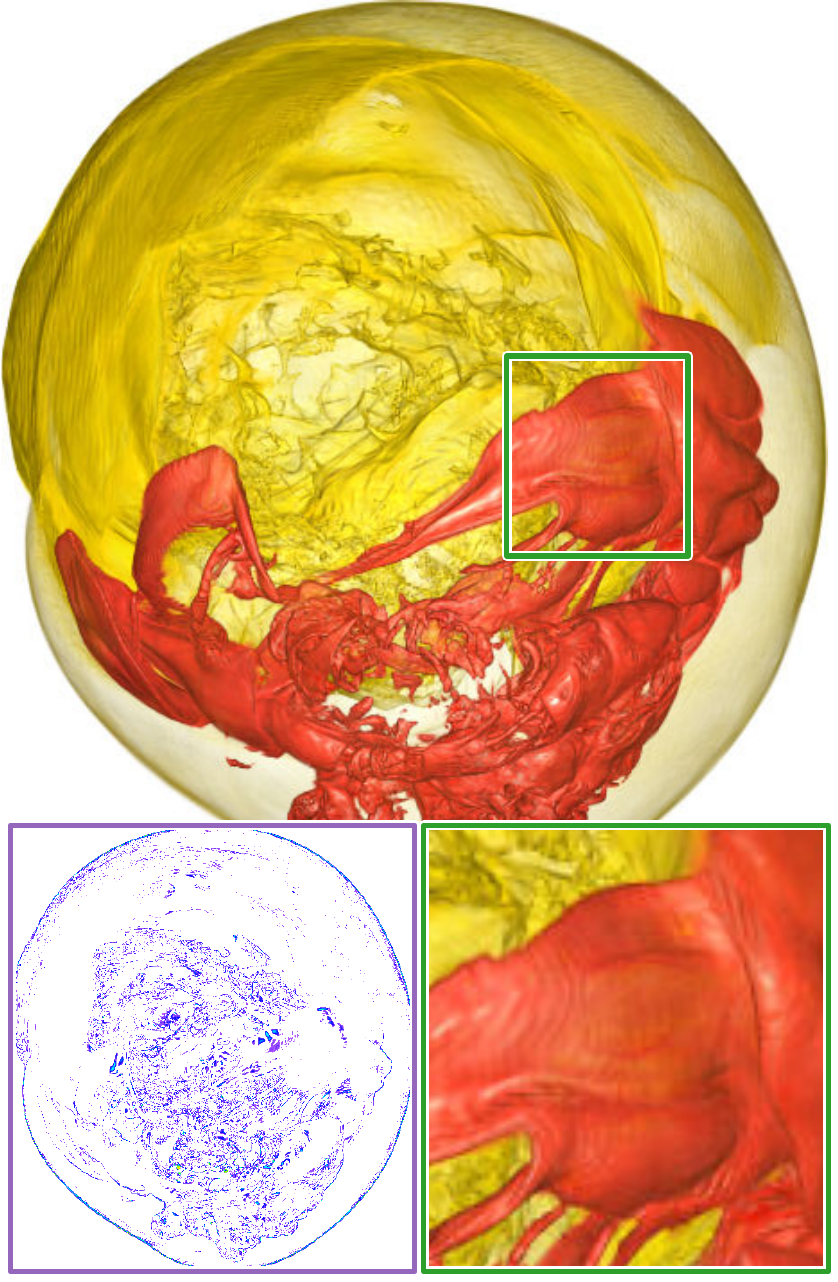}&
\includegraphics[height=1.85in]{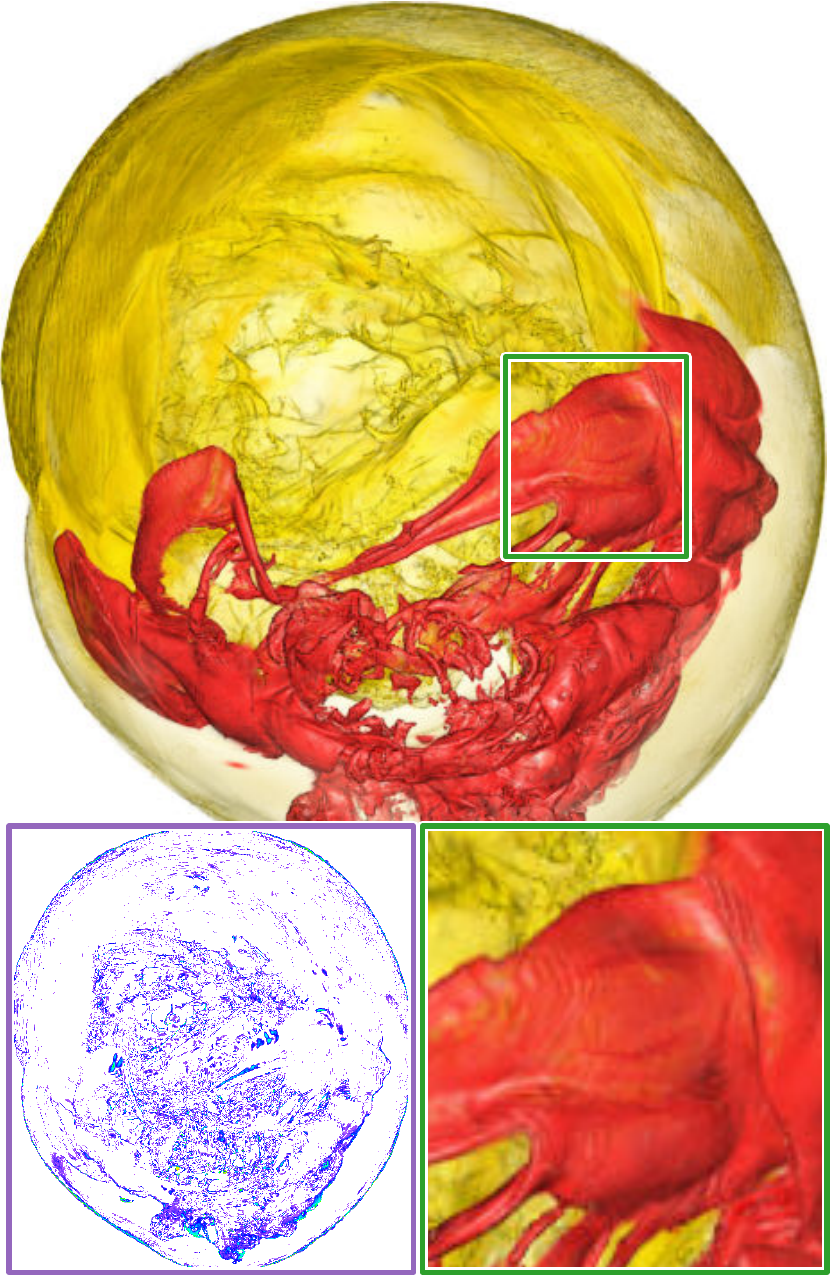}&
\includegraphics[height=1.85in]{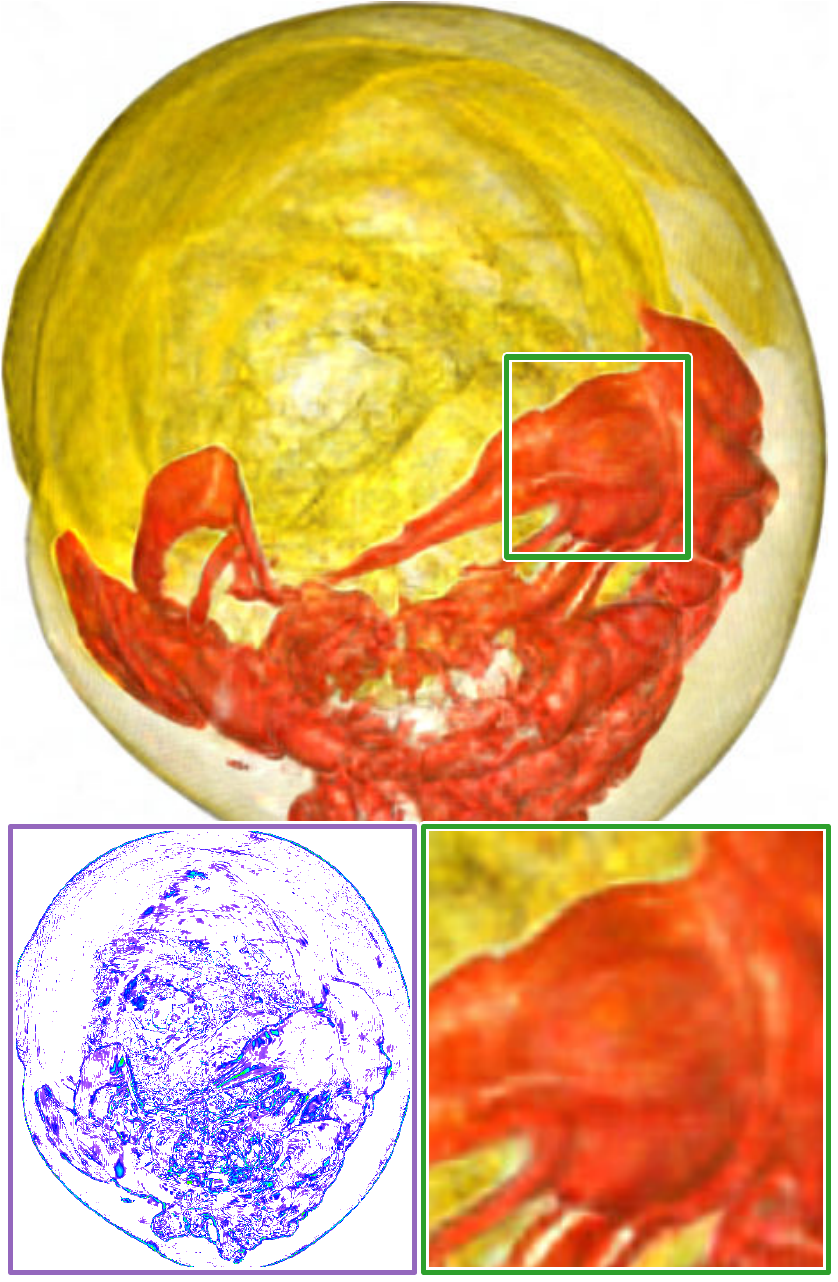}\\

\includegraphics[height=1.5in]{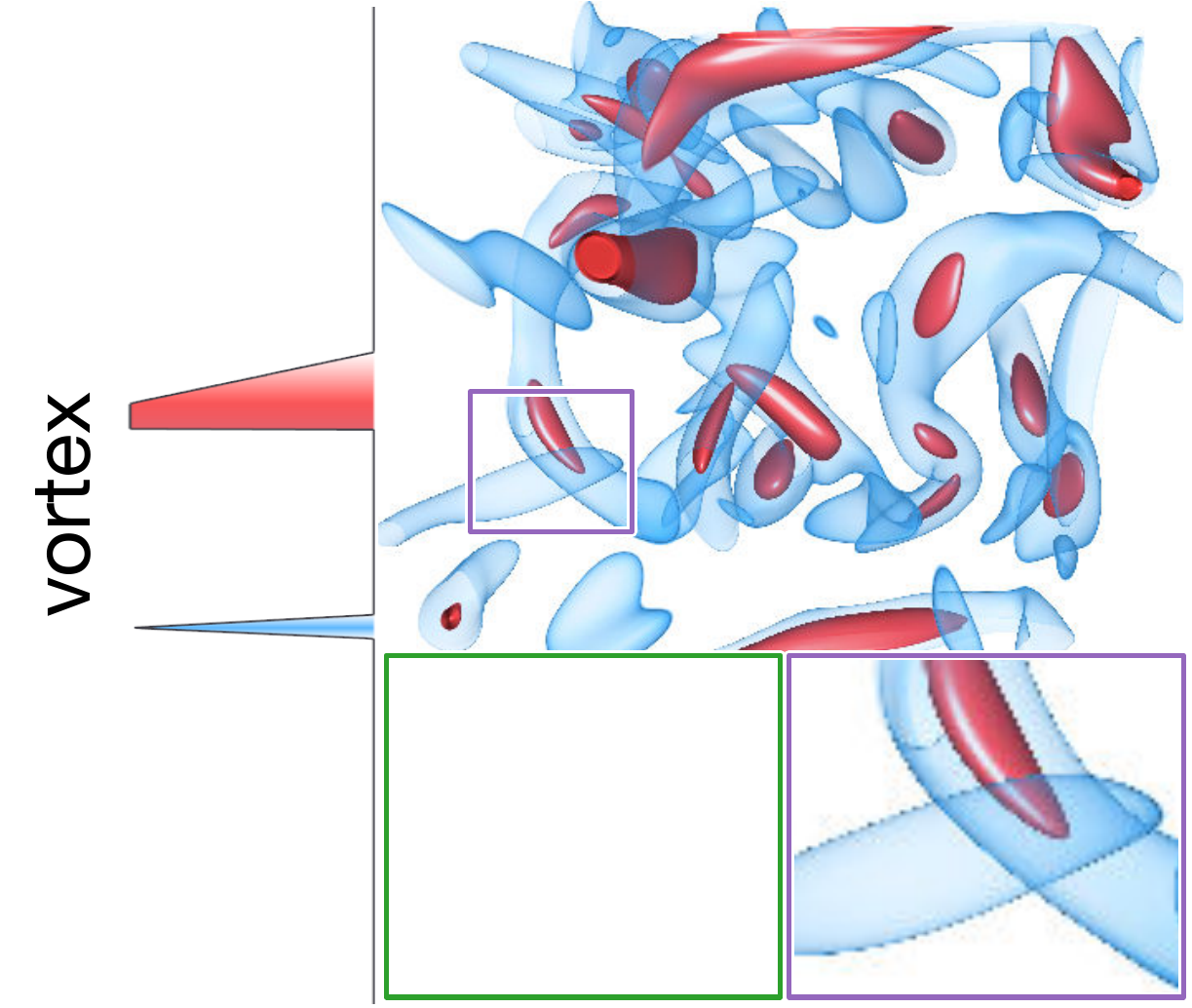}&
\includegraphics[height=1.5in]{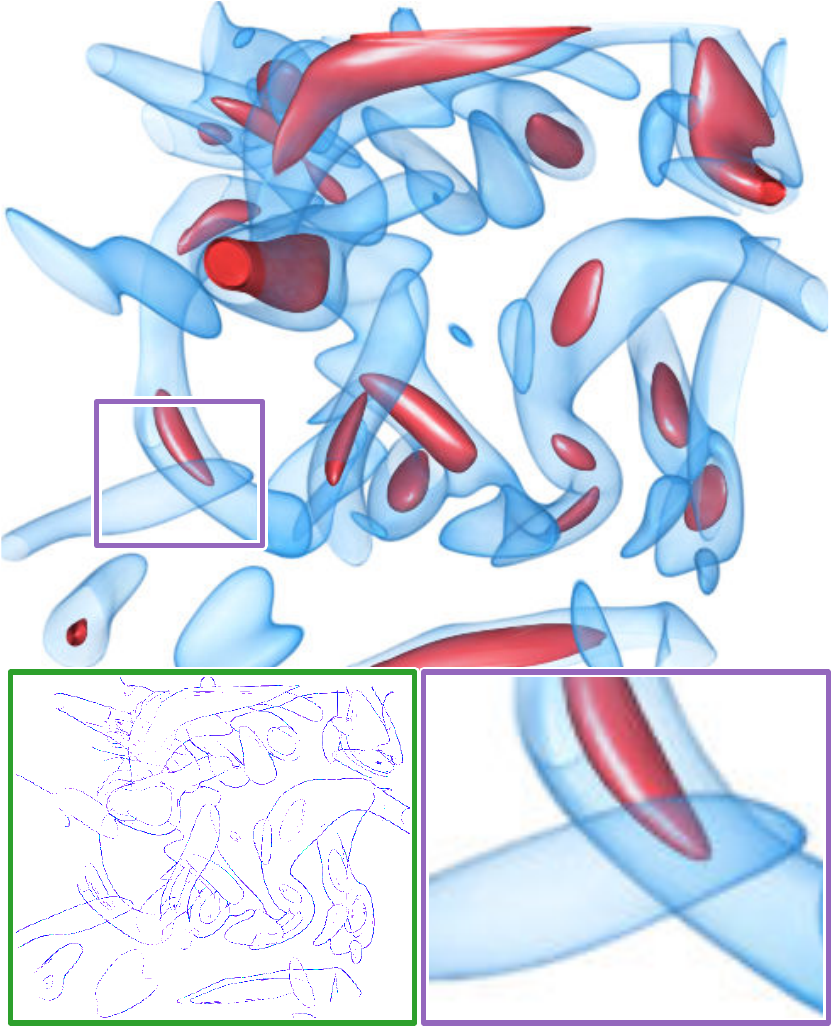}&
\includegraphics[height=1.5in]{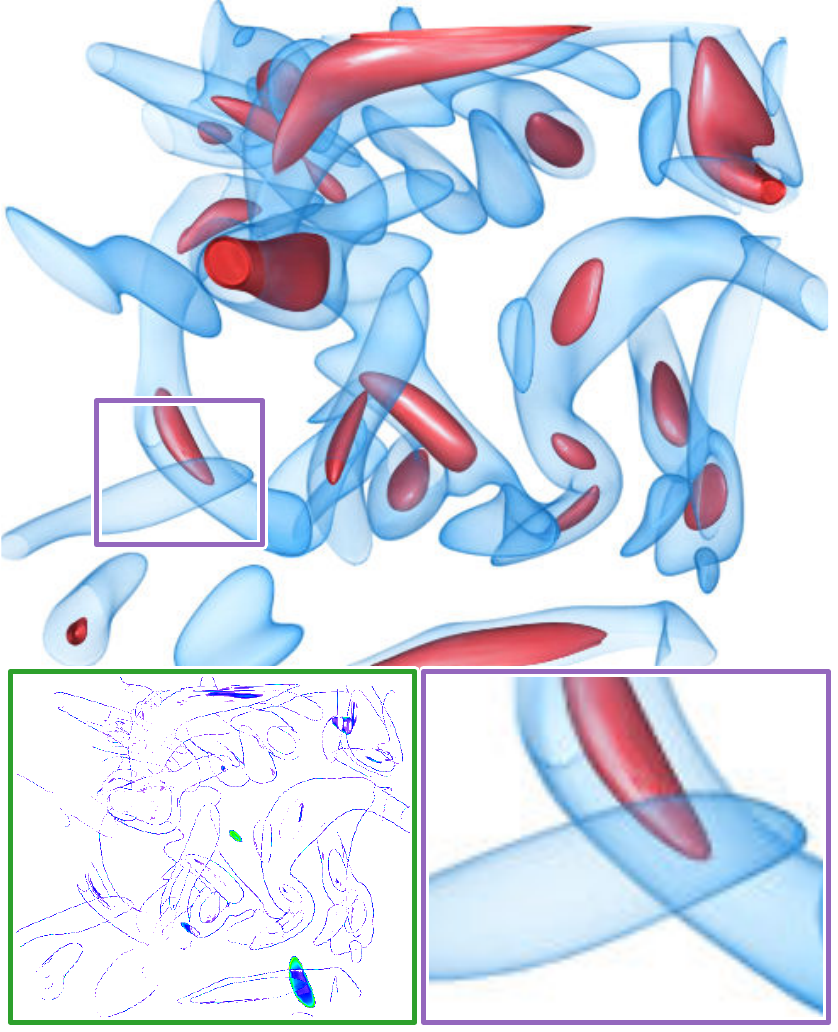}&
\includegraphics[height=1.5in]{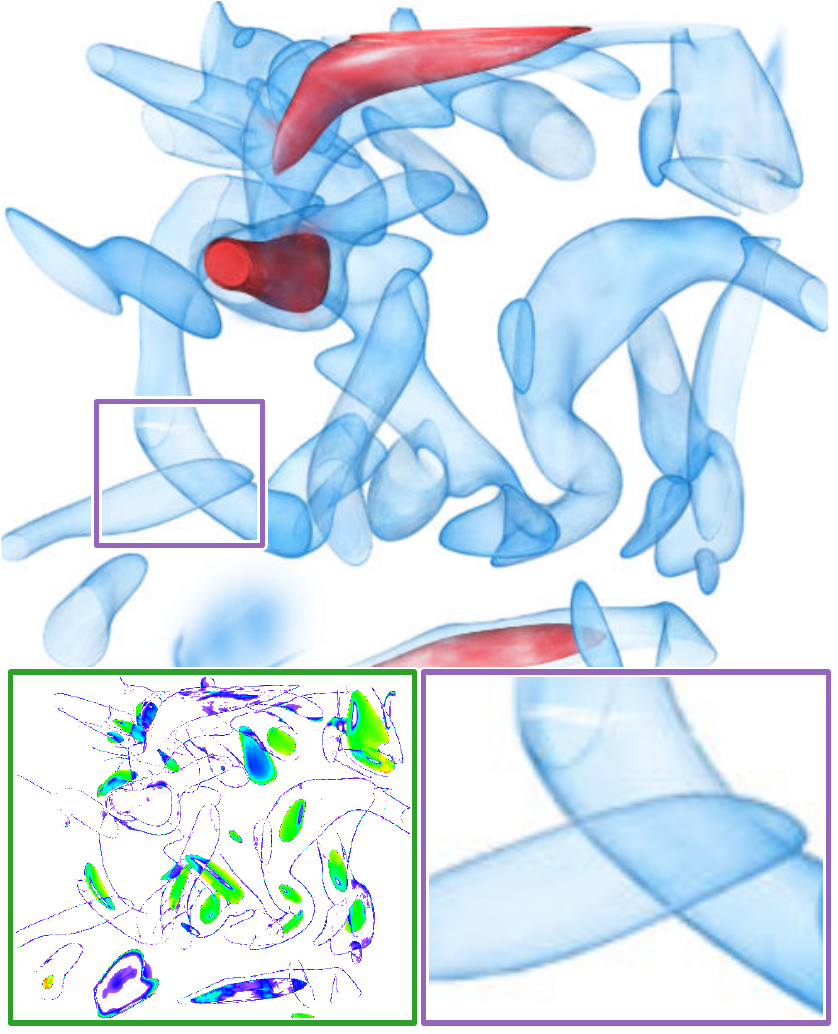}&
\includegraphics[height=1.5in]{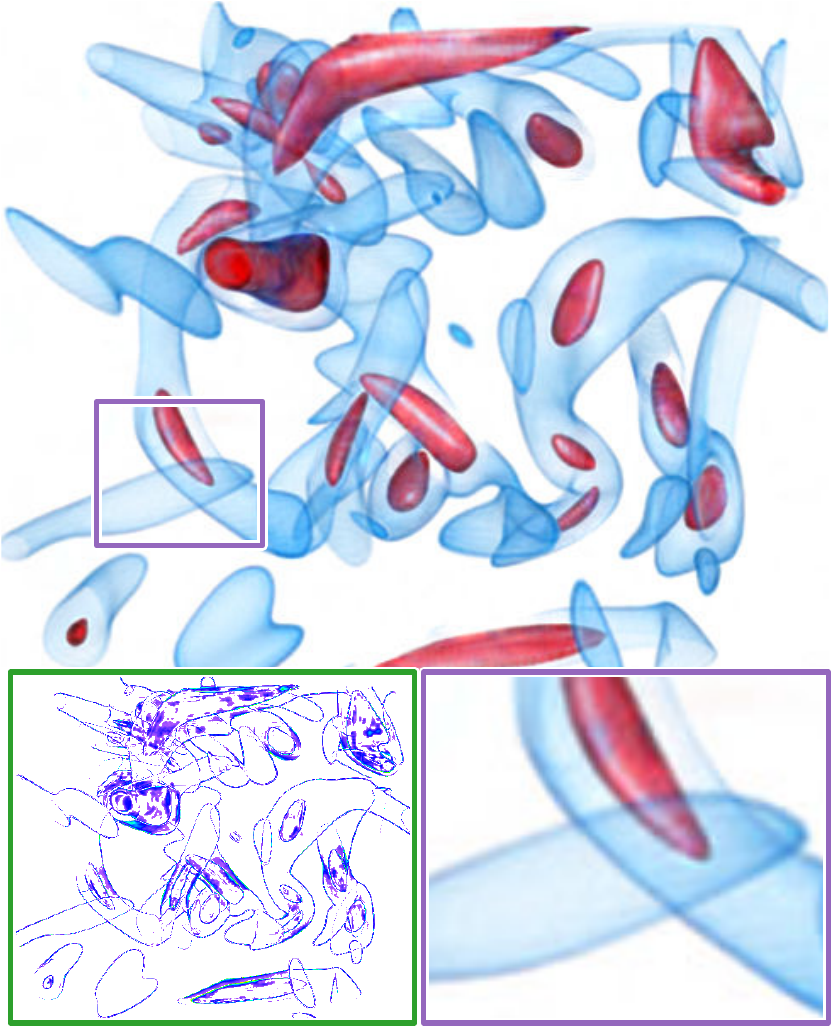}\\

\mbox{\small (a) GT} &
\mbox{\small (b) ECoNGS} &
\mbox{\small (c) 3DGS} &
\mbox{\small (d) Scaffold-GS} &
\mbox{\small (e) Plenoxels}
\end{array}$
%}
\end{center}
\vspace{-.25in} 
\caption{Comparing scene composing results of four methods w.r.t.\ GT.
While Plenoxels, Scaffold-GS, and 3DGS focus on reconstruction, only ECoNGS supports scene editing.
The difference image in the bottom-left corner shows the pixel-wise perceptible difference (blue to red indicates low to high) in the CIELUV color space. 
%( T ) and ( MF ) remove space, or use English version of ()
}
\label{fig:baseline-uncompress-results}
\vspace{-.1in}
\end{figure*}
%--------------------------------------

%--------------------------------------
\begin{figure*}[!t]
 \begin{center}
%\resizebox{\textwidth}{!}{
$\begin{array}{c@{\hspace{0.05in}}c@{\hspace{0.05in}}c@{\hspace{0.05in}}c@{\hspace{0.05in}}c}
\includegraphics[height=1.9in]{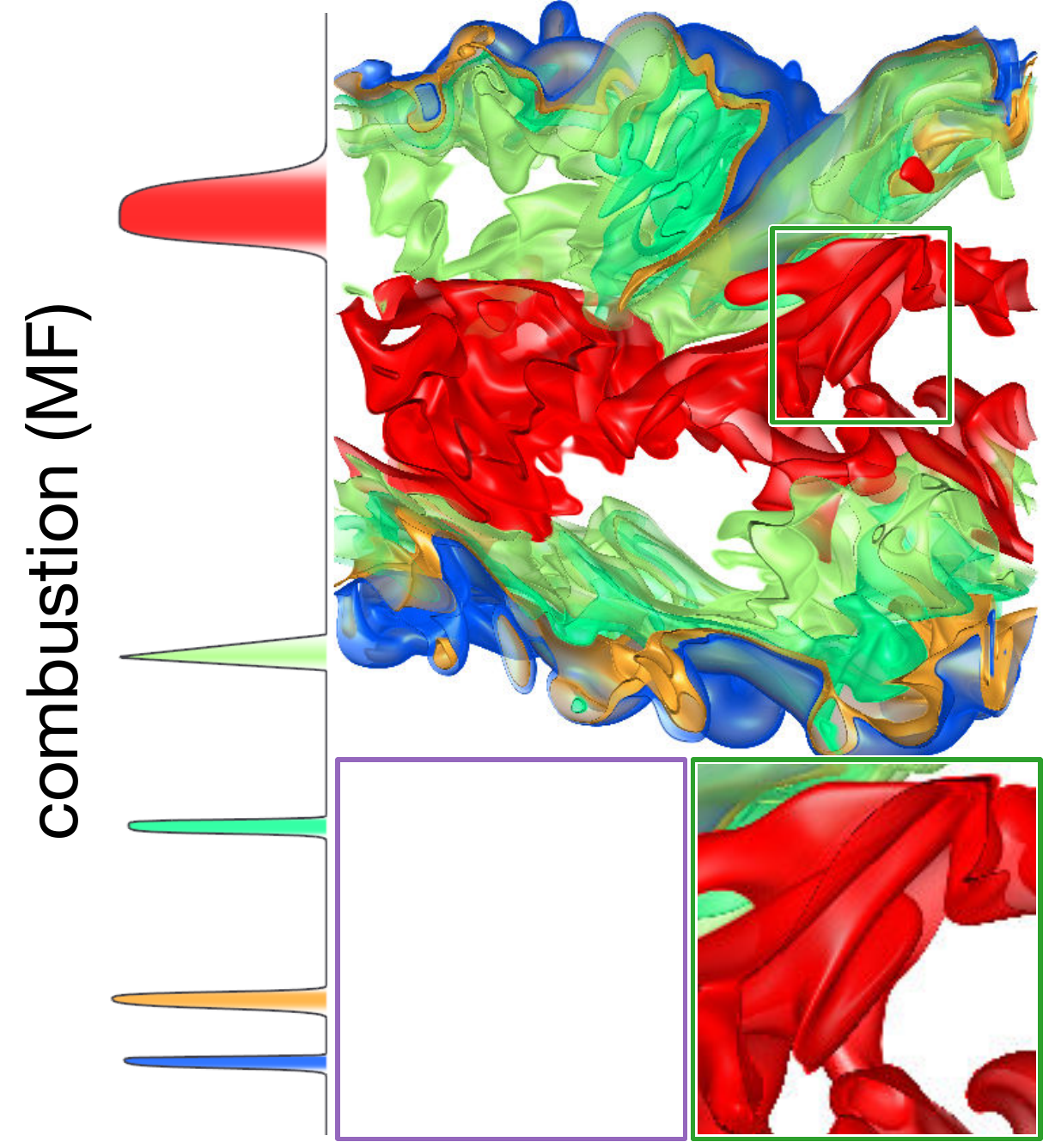}&
\includegraphics[height=1.9in]{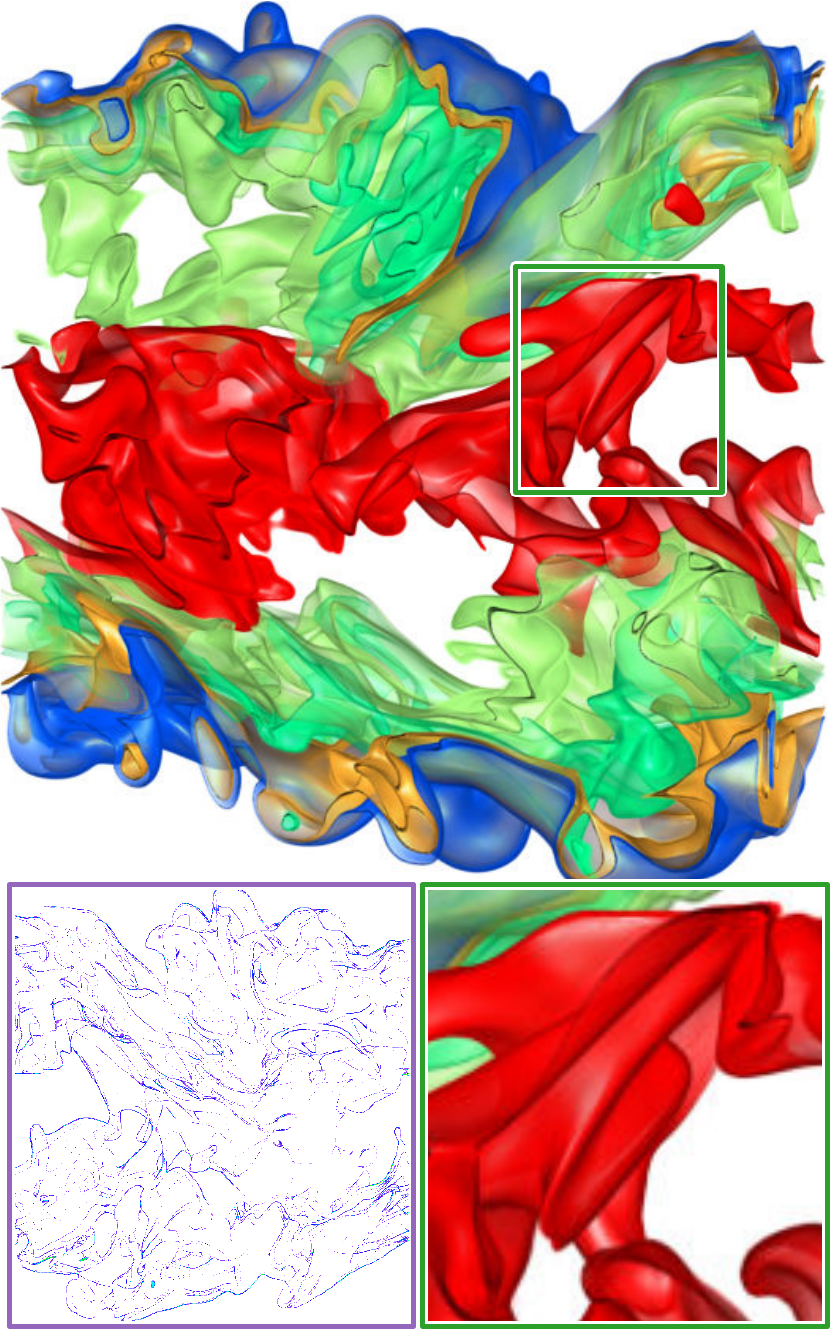}&
\includegraphics[height=1.9in]{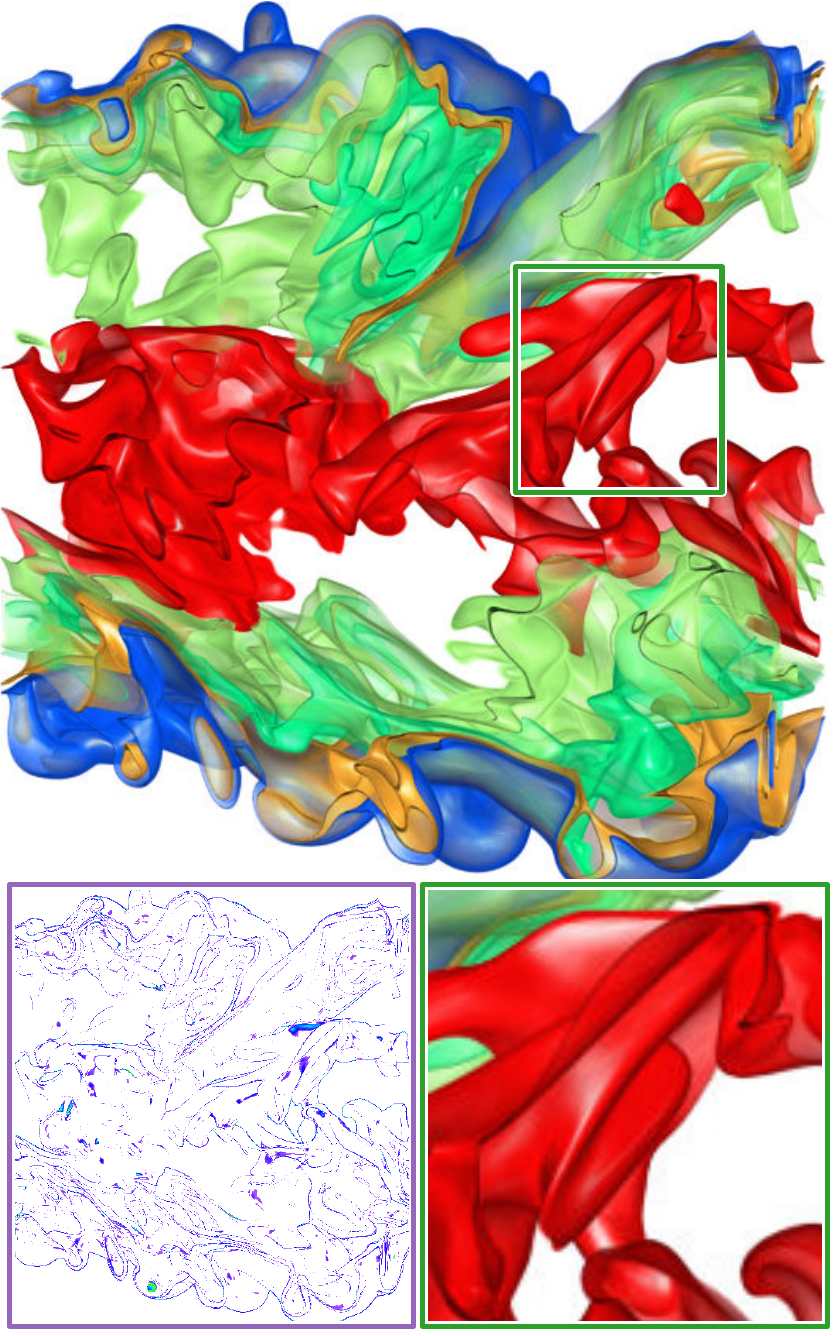}&
\includegraphics[height=1.9in]{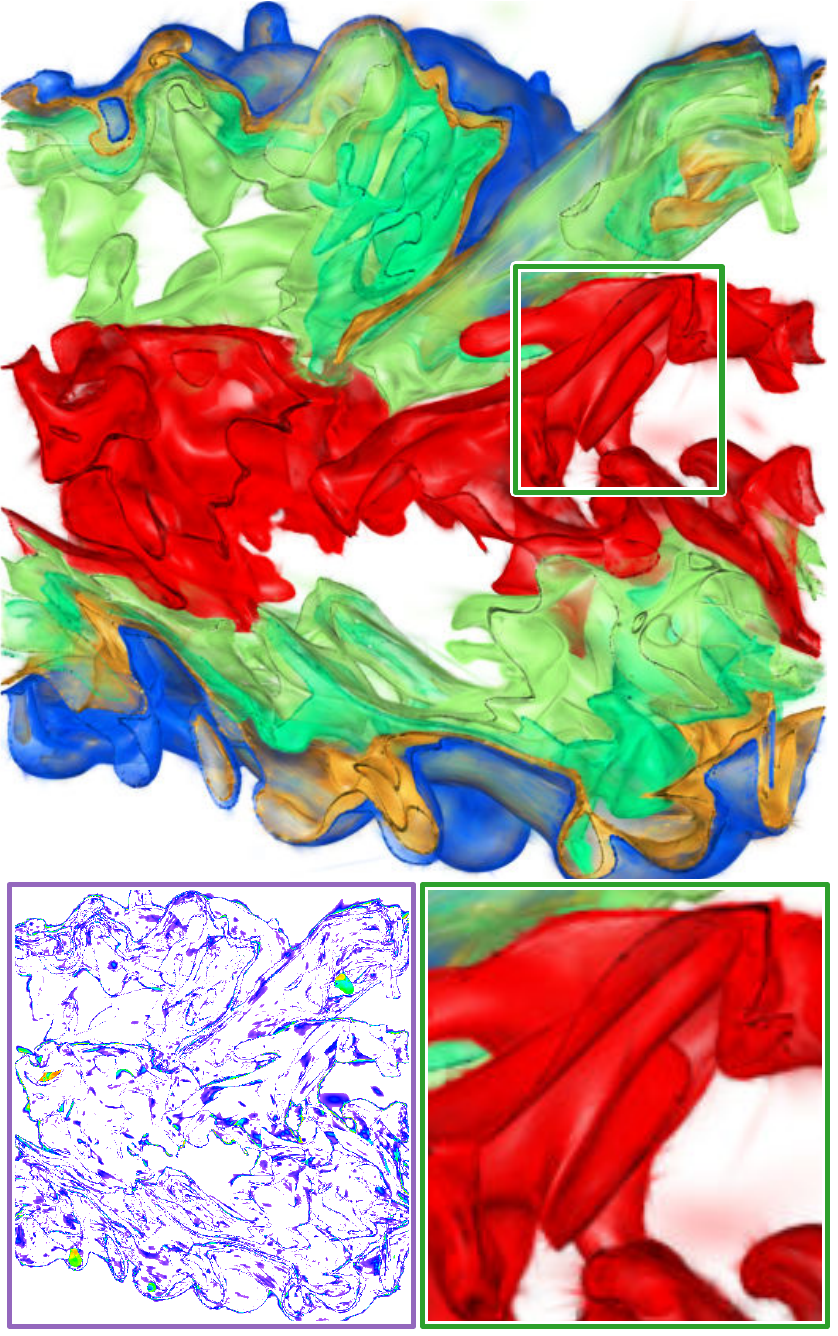}&
\includegraphics[height=1.9in]{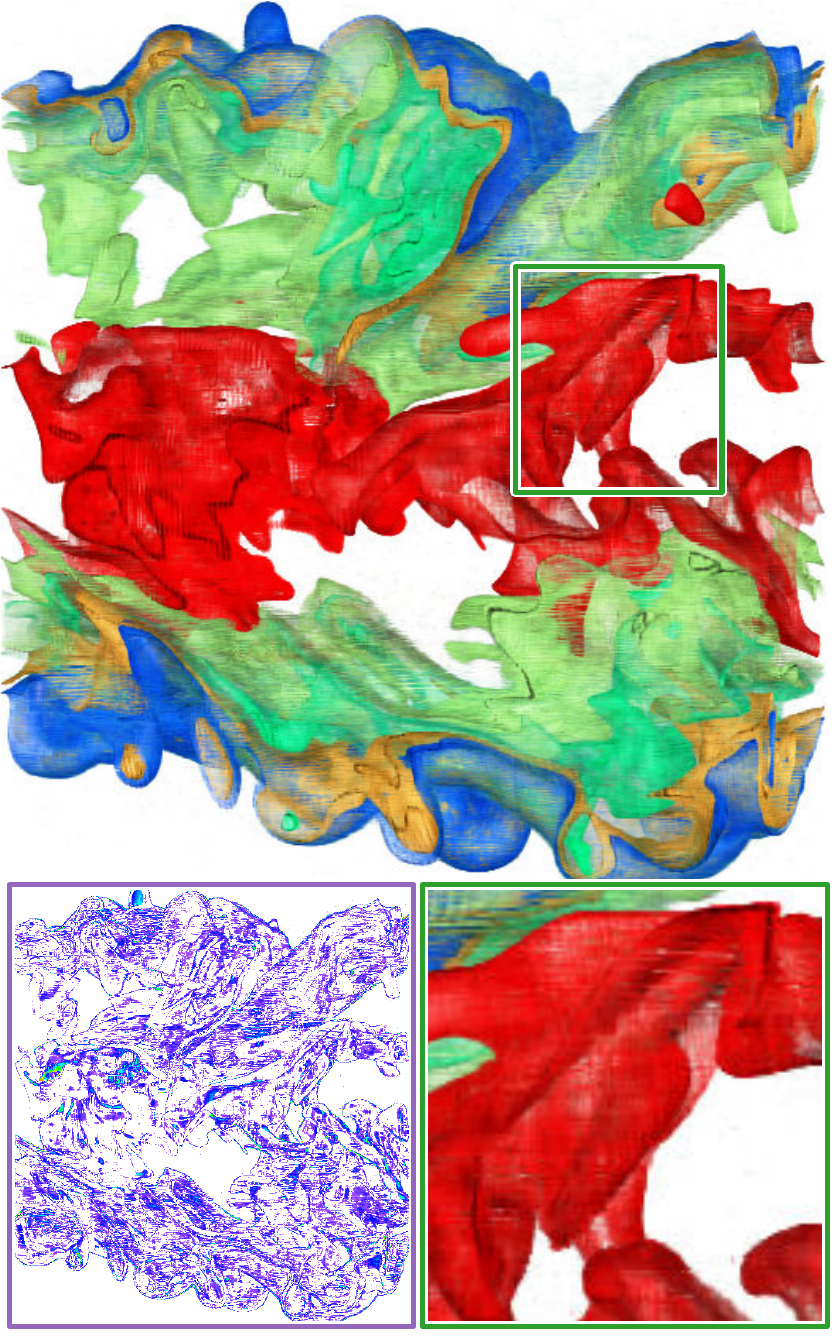}\\

\includegraphics[height=1.19in]{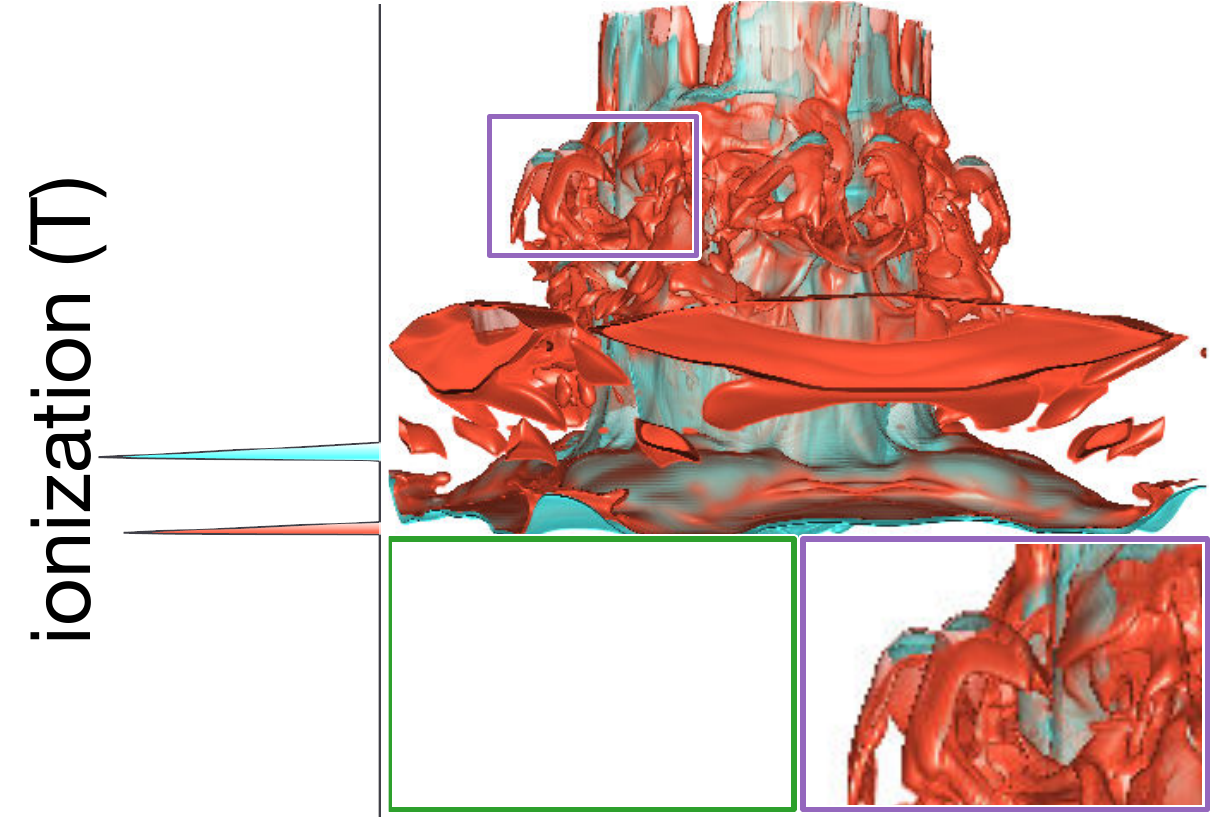}&
\includegraphics[height=1.19in]{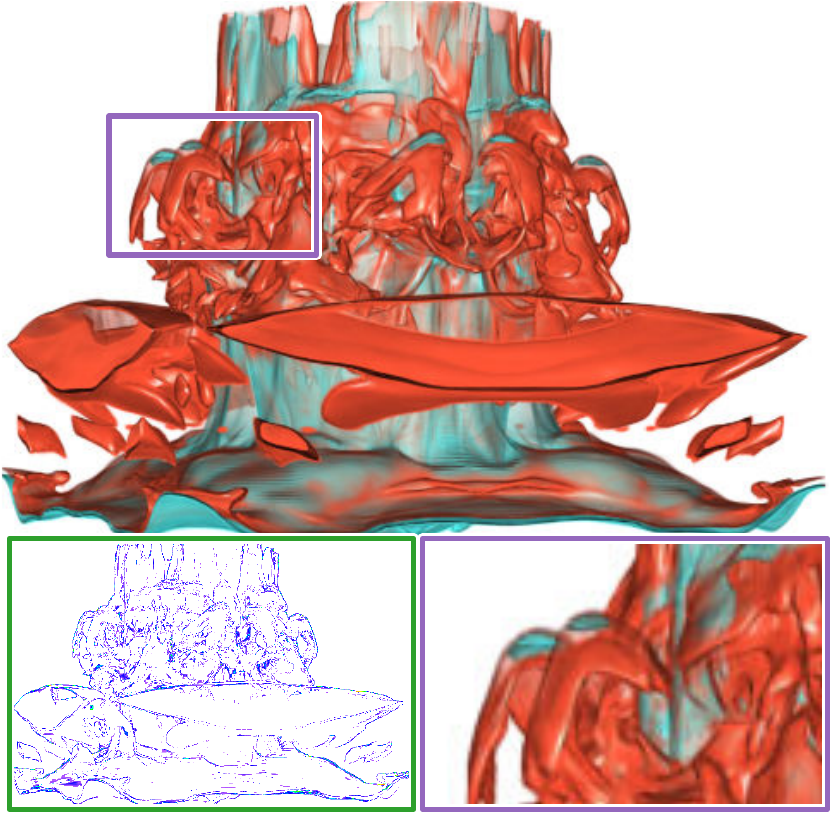}&
\includegraphics[height=1.19in]{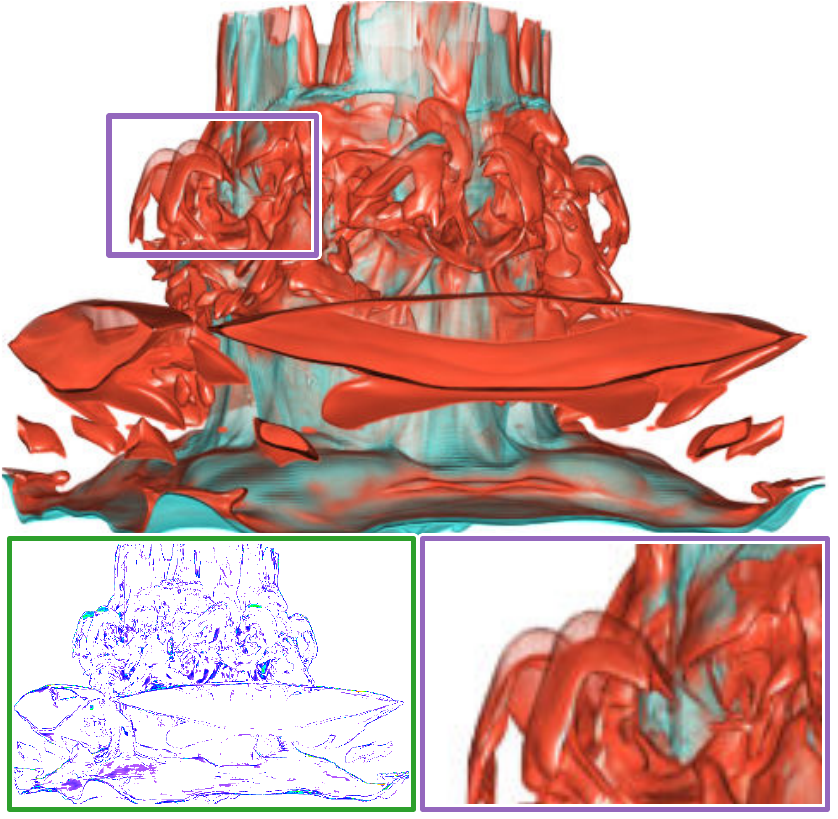}&
\includegraphics[height=1.19in]{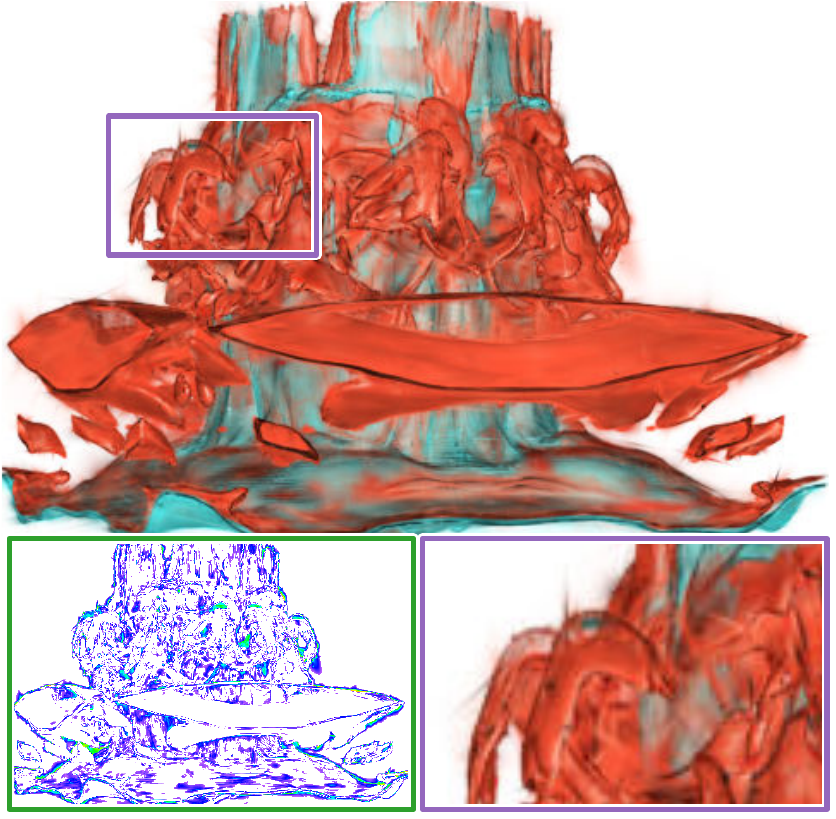}&
\includegraphics[height=1.19in]{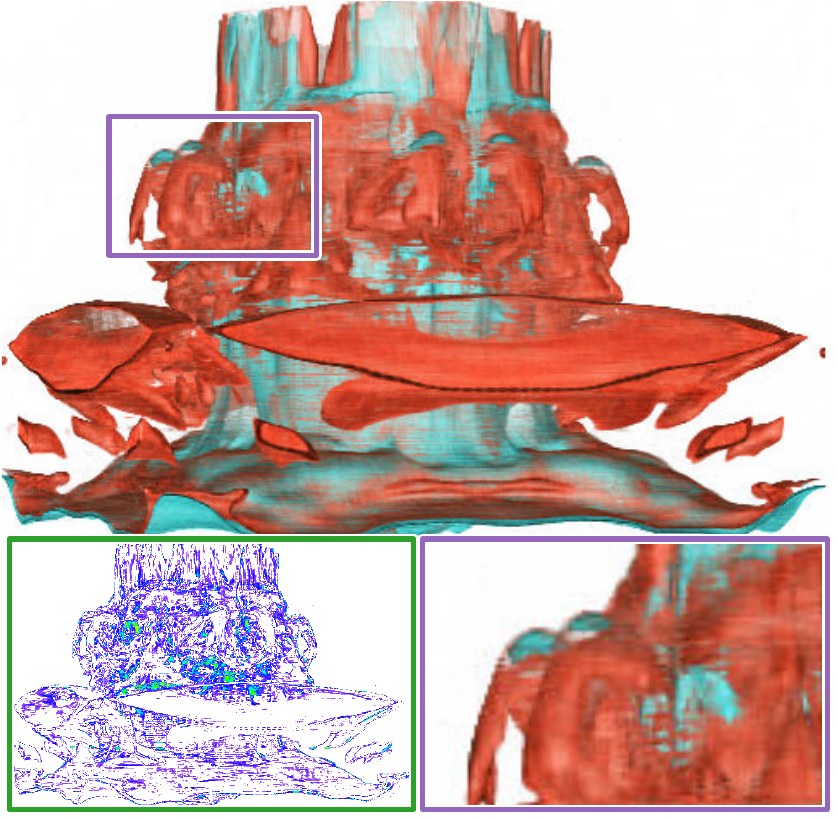}\\

\includegraphics[height=1.85in]{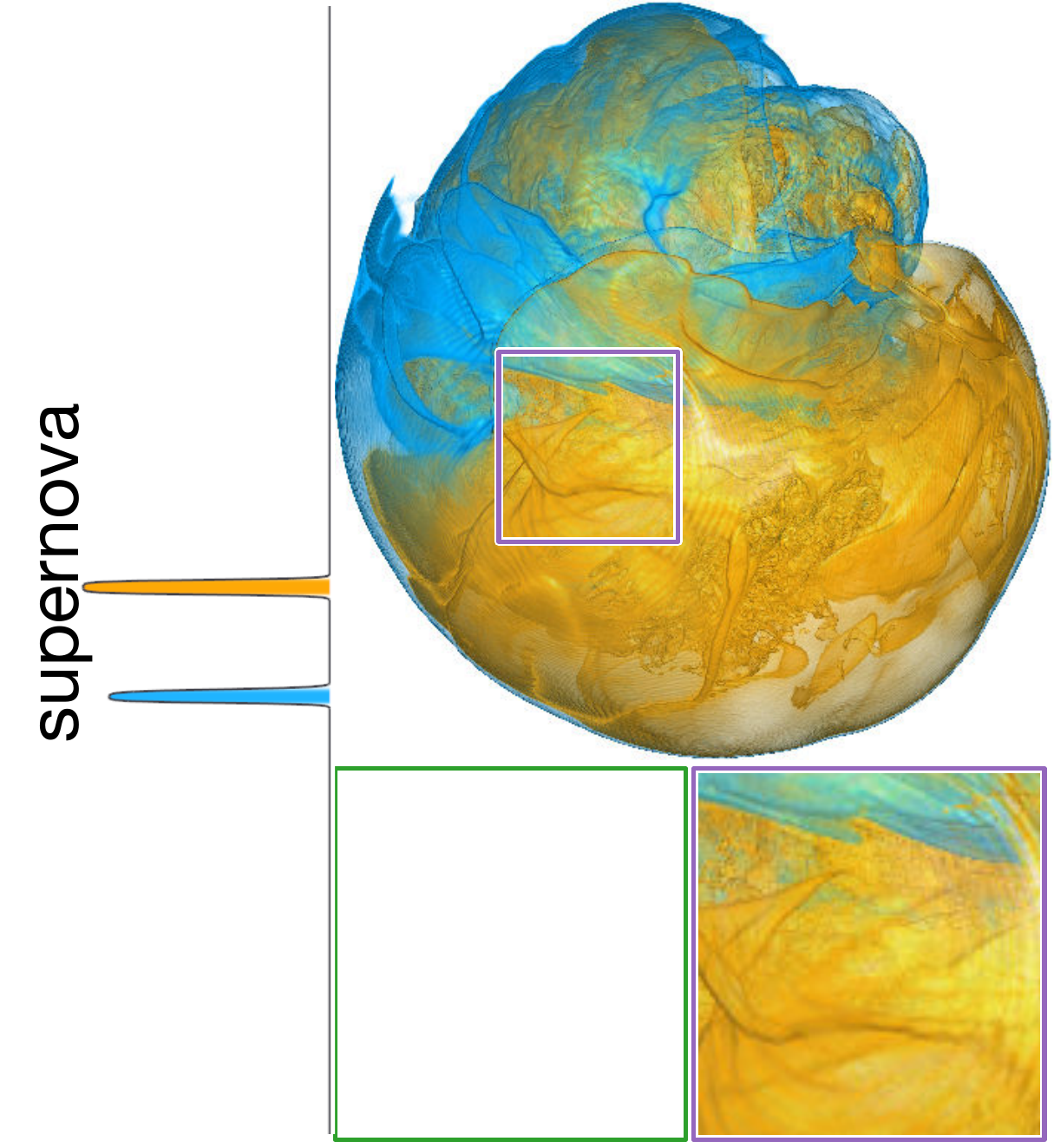}&
\includegraphics[height=1.85in]{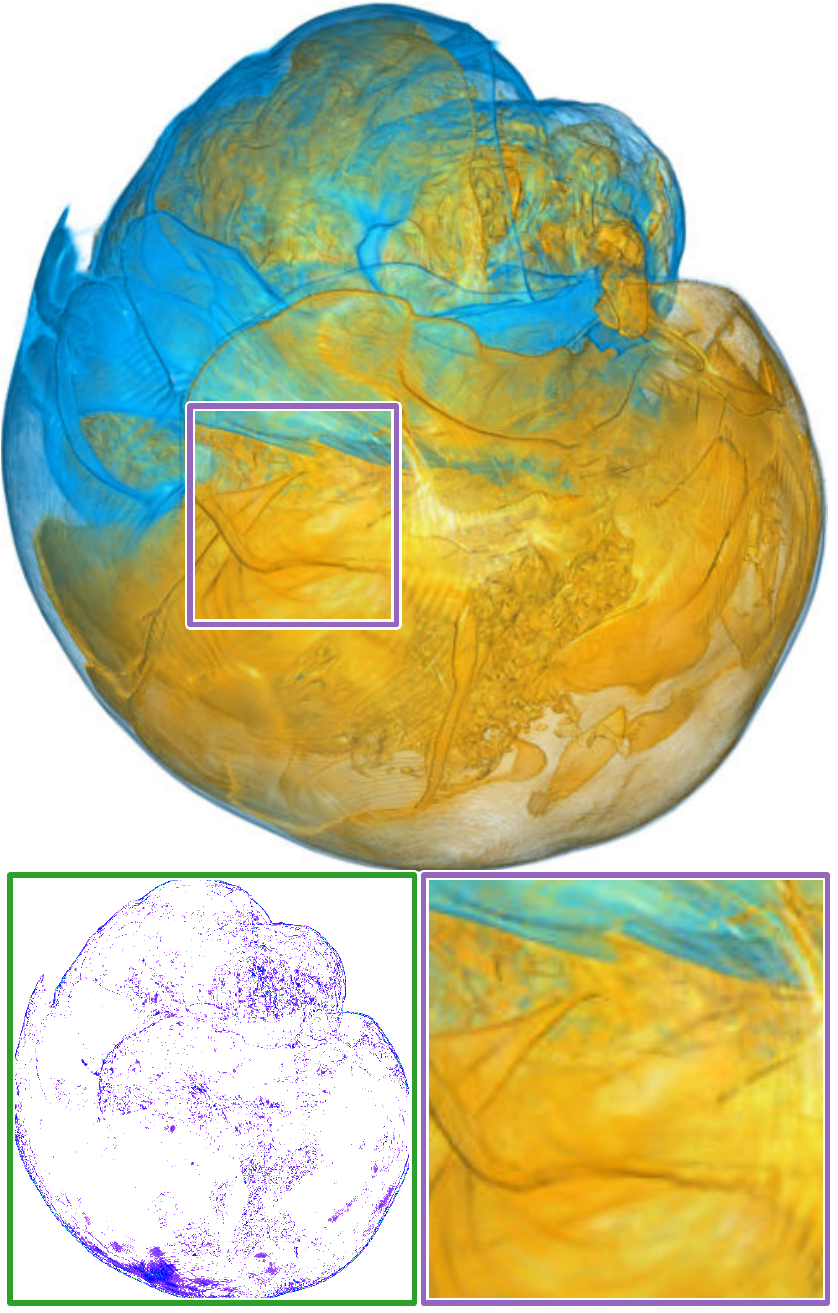}&
\includegraphics[height=1.85in]{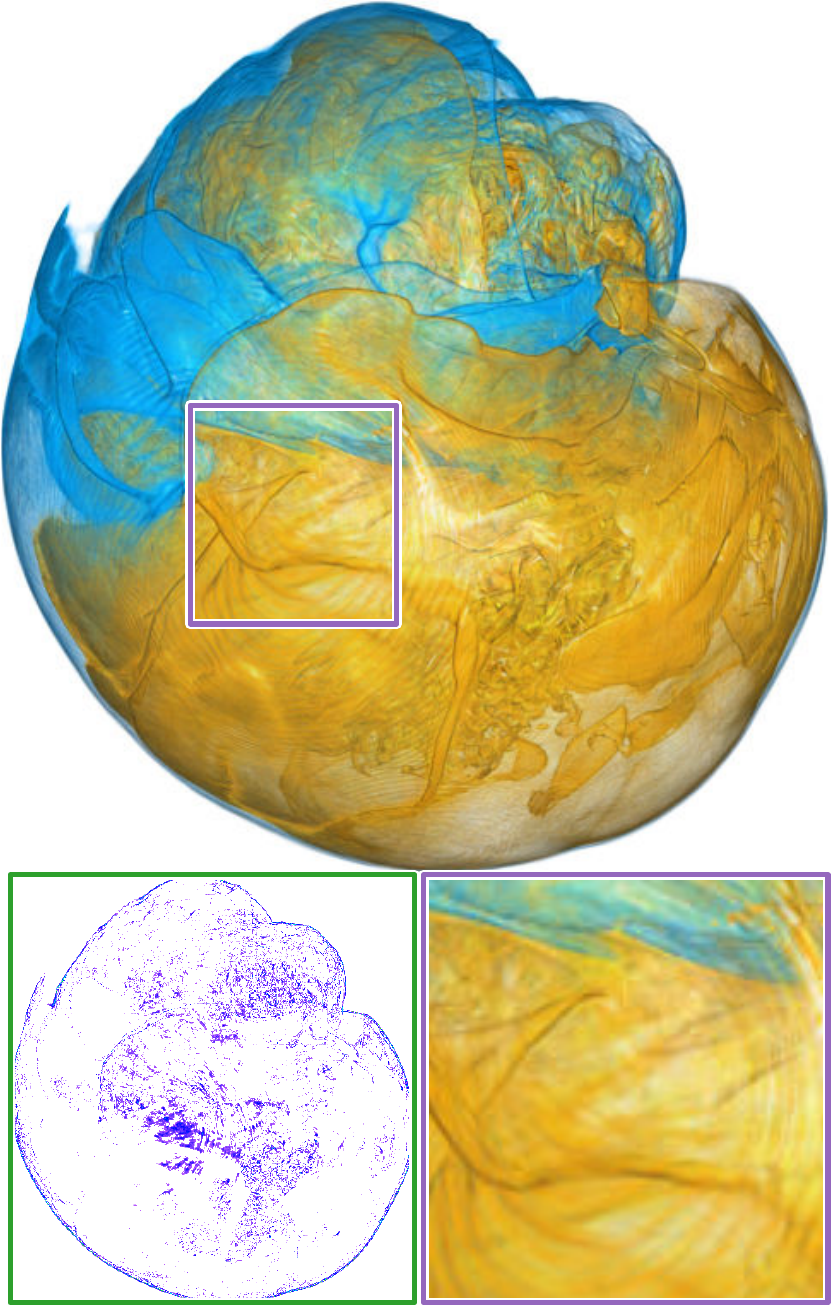}&
\includegraphics[height=1.85in]{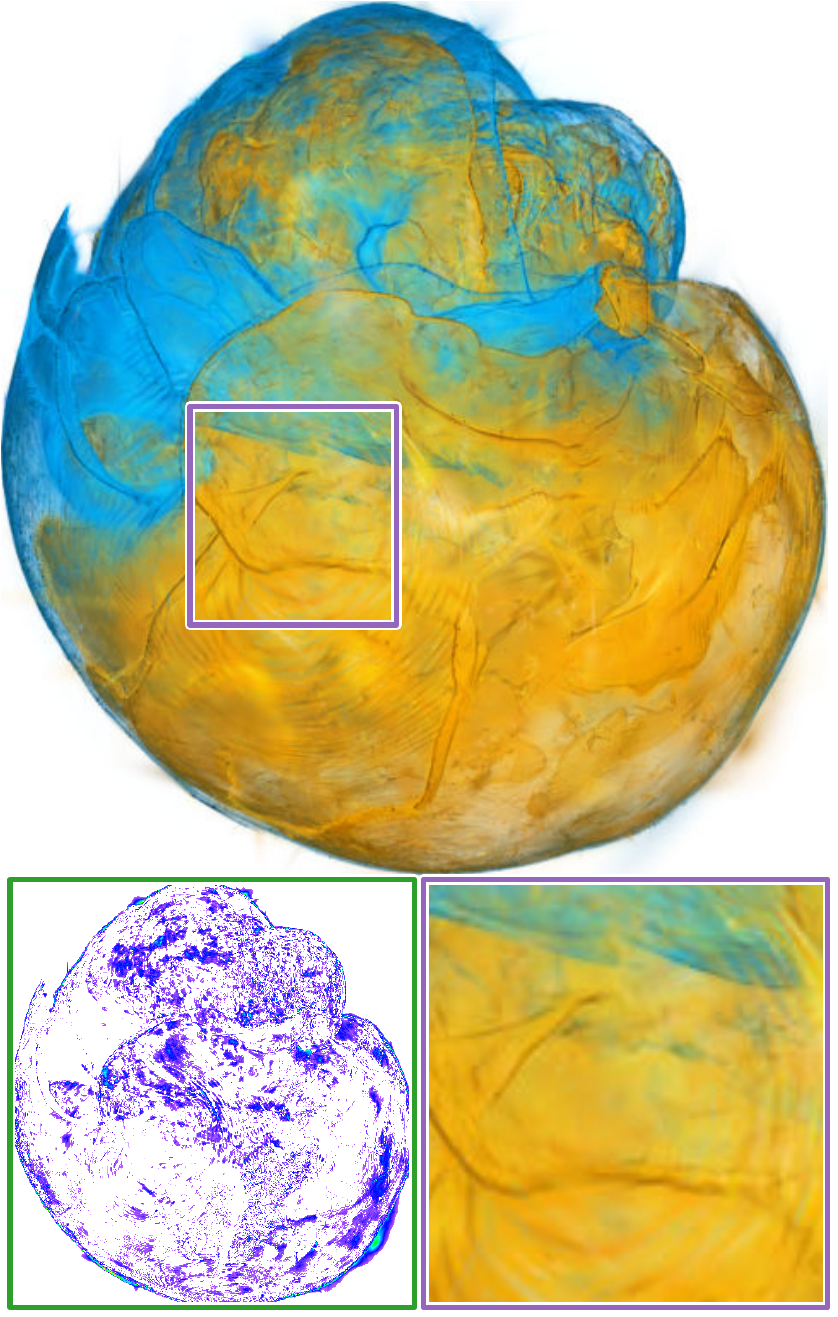}&
\includegraphics[height=1.85in]{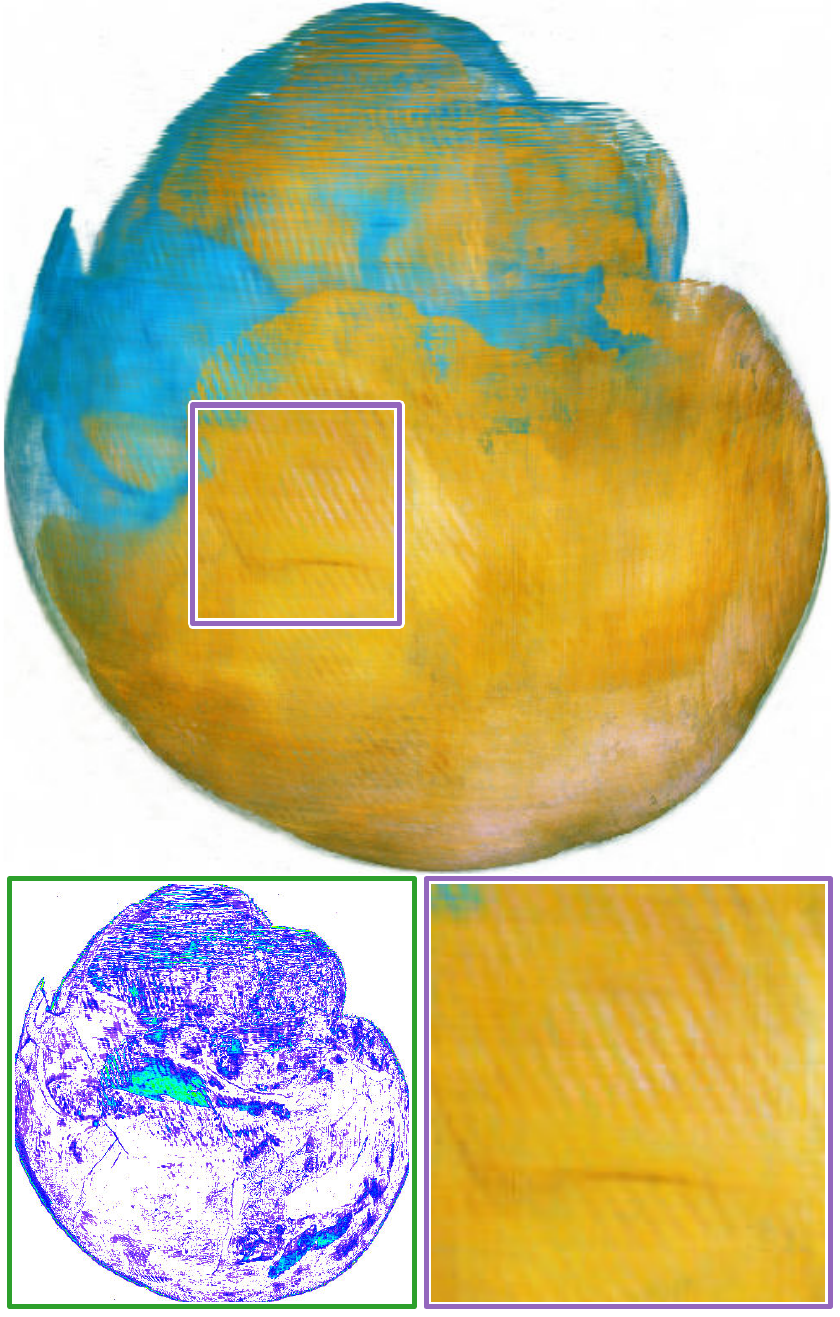}\\

\includegraphics[height=1.8in]{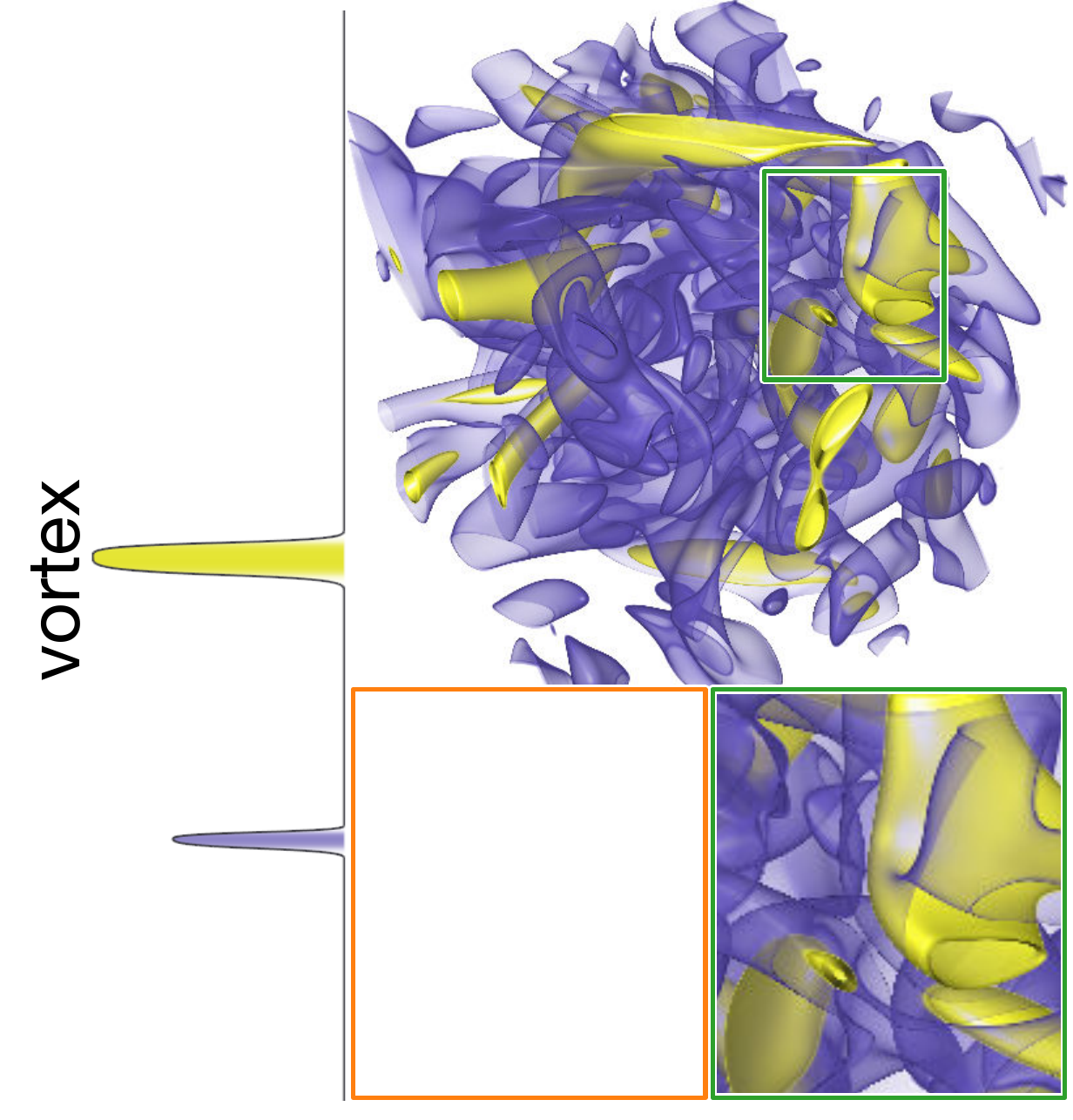}&
\includegraphics[height=1.8in]{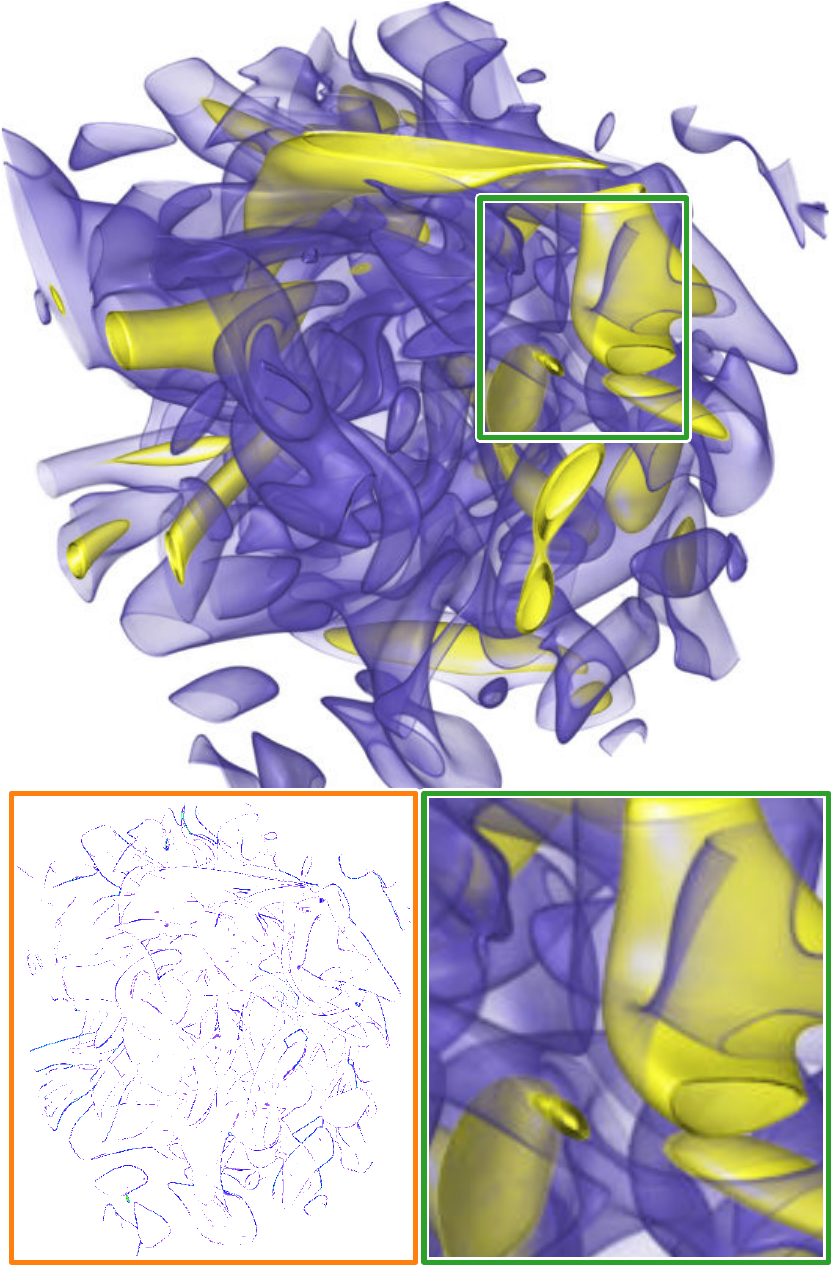}&
\includegraphics[height=1.8in]{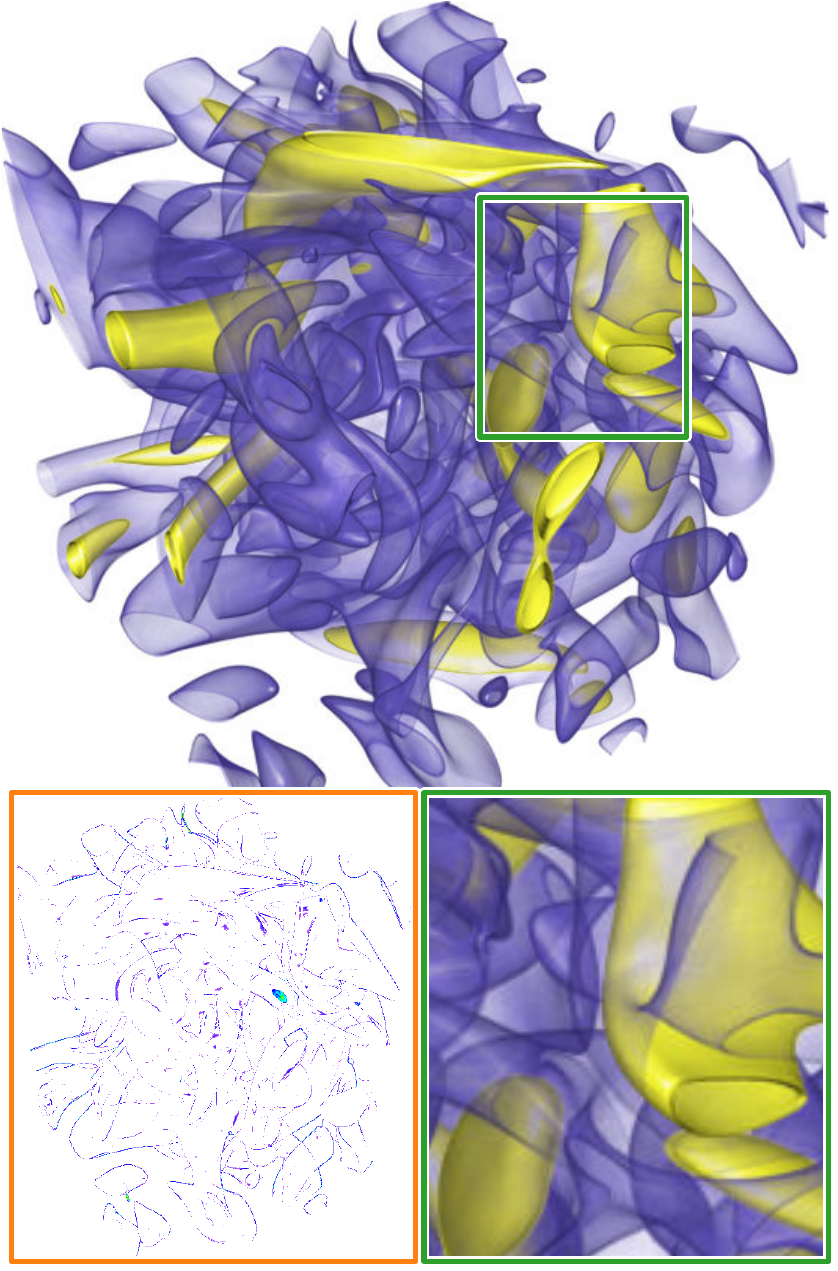}&
\includegraphics[height=1.8in]{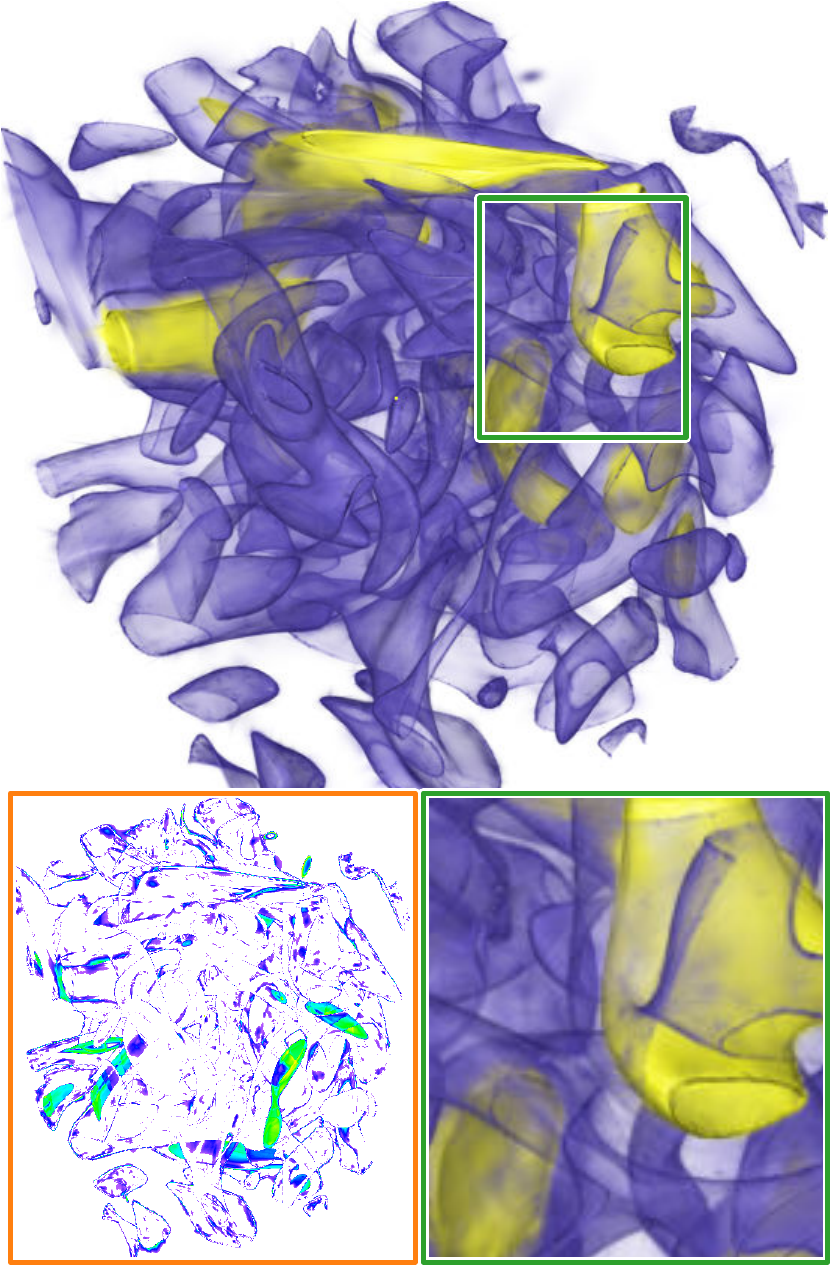}&
\includegraphics[height=1.8in]{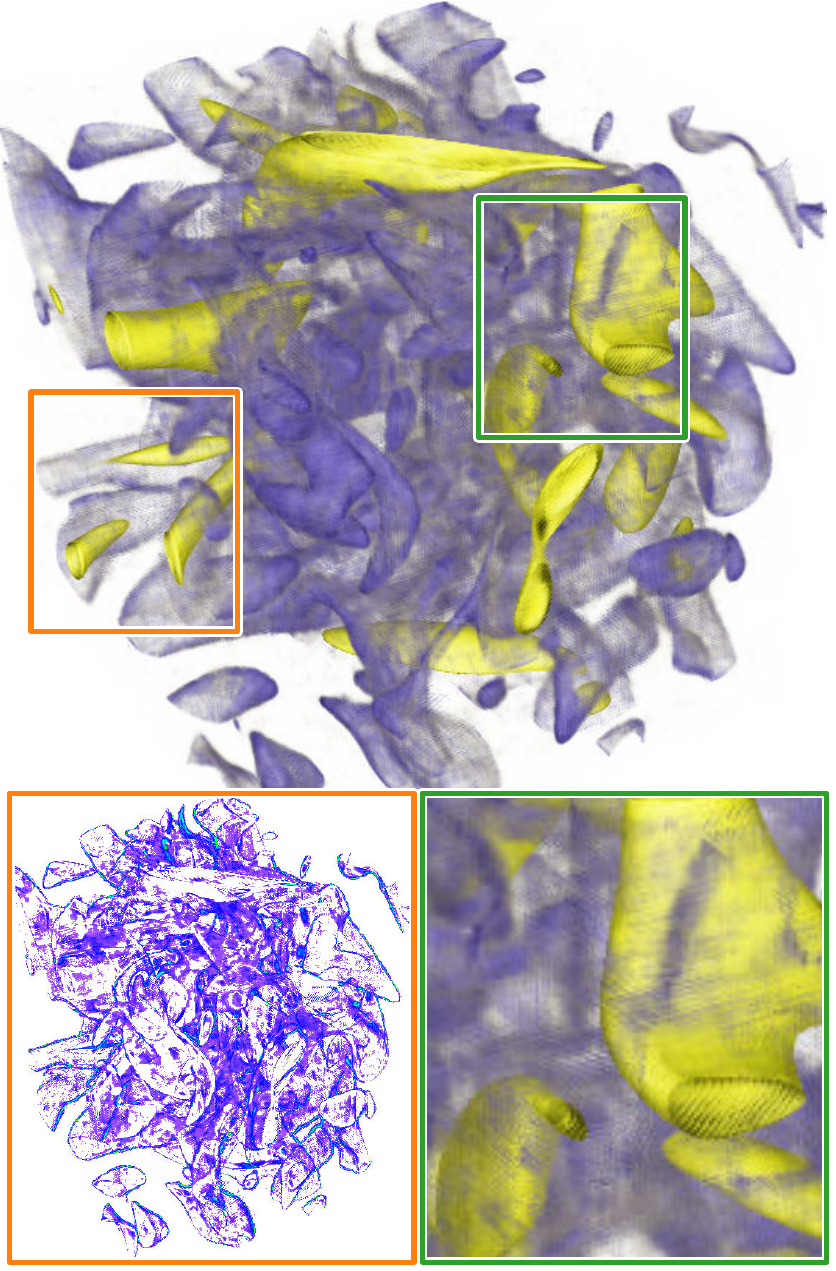}\\

\mbox{\small (a) GT} &
\mbox{\small (b) ECoNGS} &
\mbox{\small (c) iVR-GS} &
\mbox{\small (d) HAC} &
\mbox{\small (e) CCNeRF}
\end{array}$
%}
\end{center}
\vspace{-.25in}
\caption{Comparing scene composing results of four methods w.r.t.\ GT.
CCNeRF and HAC bake TF information and lighting into their representations, so they cannot perform scene editing.
In contrast, both iVR-GS and ECoNGS enable scene editing.
%The difference image in the bottom-left corner shows the pixel-wise perceptible difference (blue to red indicates low to high) in the CIELUV color space.
%do the same as Figure 4
}
\label{fig:baseline-compress-results}
\vspace{-.1in}
\end{figure*}
%--------------------------------------

\section{Results and Discussion}

\subsection{Experiment Setup}

{\bf Training and inference configuration.}
We implemented ECoNGS in PyTorch, using CUDA kernels optimized for efficient rasterization.
The Adam optimizer was leveraged to update learnable parameters. 
We set the learning rates for the anchor feature, opacity, scaling, rotation, and TF embeddings to 0.01 and 0.002, respectively, and decayed them exponentially as training progressed.
The palette color was updated with a fixed learning rate $1\times10^{-4}$.
%For other anchor attributes and MLP parameters, we used the same learning rates as those of other neural Gaussian methods. 
We set $\lambda_e$ to 0.0001.
\hot{We set the number of neural editable Gaussians derived from each anchor point to $K=10$.}
Tables~\ref{tab:univariate-datasets}~and~\ref{tab:multivariate-datasets} list all the univariate and multivariate datasets used for experiments.
We manually set the basic TF for each dataset, with appropriate colors and opacities to highlight different internal structures.
\hot{In practice, when a target TF is not known in advance, as in iVR-GS~\cite{Tang-TVCG25}, users can also select several basic TFs whose opacity bumps are non-overlapping in voxel values to cover the full scalar range and compose them for exploration.}
For each scene, we applied icosphere sampling to uniformly sample 92 multi-view training images within ParaView using the NVIDIA IndeX plugin \hot{with depth enhancement~\cite{Zheng-TVCG13}}.
During rendering, we added a headlight with diverse lighting parameters for each dataset and rendered the images with 800$\times$800 resolution.

All volume datasets were rendered on a high-performance cluster utilizing an NVIDIA A40 GPU with 48 GB of memory.
We recorded CPU/GPU runtime, memory usage, and average DVR rendering FPS across various basic scenes.
Note that the FPS may vary slightly when the rendering TF is adjusted, as different TFs alter the number of visible voxels, thereby affecting the ray-integration workload.
In contrast, the runtime memory remains nearly constant regardless of TF changes, since the entire volume and related resources are pre-loaded into GPU memory, and the memory footprint is determined by the volume resolution rather than the rendered content.

{\bf Baselines and evaluation metrics.}
We compared ECoNGS with three non-compressed and three compressed NVS methods.
%\vspace{-0.025in}
%\begin{myitemize}
%  \item \textbf{Plenoxels}~\cite{Fridovich-Keil-CVPR22} is a NeRF-based method representing a scene with an explicit sparse 3D grid. Its composability is realized by summing the parameters of multiple grids from basic models. 
%  \item \textbf{3DGS}~\cite{Kerbl-TOG23} represents a scene with a collection of anisotropic 3D Gaussian primitives equipped with differentiable attributes.
%  \item \textbf{Scaffold-GS}~\cite{Lu-CVPR24} introduces a structured scaffold representation on 3D Gaussian splatting, leveraging hierarchical sparsity and view-adaptive rendering for scalable training. 
%  \item \textbf{CCNeRF}~\cite{Tang-neurips22} is a compressed NeRF-based approach that encodes a scene into multiple compact low-rank tensor components. In our evaluation, we adopt the CCNeRF-CP model variant for its compactness.
%  \item \textbf{HAC}~\cite{Chen-ECCV24} is a neural Gaussian compression method that leverages neural entropy coding to reduce size while maintaining rendering quality. 
%  \item \textbf{iVR-GS}~\cite{Tang-TVCG25} is a compressed NVS method that leverages pruning and quantization techniques to reduce the parameter size of editable Gaussian primitives.
%\end{myitemize}
The non-compressed category comprises the NeRF-based method Plenoxels~\cite{Fridovich-Keil-CVPR22}, vanilla Gaussian method 3DGS~\cite{Kerbl-TOG23}, and neural Gaussian method Scaffold-GS~\cite{Lu-CVPR24}. 
The compressed methods include CCNeRF \cite{Tang-neurips22}, a tensor-decomposition-based NeRF compression method (we used the CCNeRF-CP variant for its compactness); HAC \cite{Chen-ECCV24}, a neural Gaussian compression method using neural entropy coding; and iVR-GS \cite{Tang-TVCG25}, which prunes and quantizes editable Gaussians for compact scene representation.
%\vspace{-0.15in}
For each univariate or multivariate volume dataset, after optimization, we evaluated ECoNGS and other baseline methods on 181 unseen testing views that enclose each basic scene. 
In the quantitative analysis, we reported average performances across different basic scenes and testing views.
In addition to evaluating each basic scene, we combined the basic TFs from all basic scenes within a dataset to render ground-truth (GT) images of a composed VolVis scene. 
We then assessed the composability of NVS methods by merging the NVS models trained on each basic scene.
\hot{For rendering results of different views, we refer readers to the supplementary video.}
Training and inference for all methods were conducted on a local workstation with an NVIDIA RTX 4090 GPU.
We leveraged \textit{peak signal-to-noise ratio} (PSNR) and \textit{learned perceptual image patch similarity} (LPIPS)~\cite{Zhang2018} to measure reconstruction quality.

\vspace{-0.075in}
\subsection{Evaluation on Univariate Datasets}

{\bf Quantitative analysis.}
Table~\ref{tab:univariate-baseline} summarizes the average reconstruction quality, training time, and model size across all basic scenes for different methods on four univariate volume datasets.
Overall, ECoNGS consistently achieves the best reconstruction quality while using a noticeably smaller model size than other methods.

When compared against uncompressed baselines (Plenoxels, 3DGS, and Scaffold-GS), our method achieves higher PSNR while using only a fraction of the storage.
For example, on the vortex dataset, ECoNGS surpasses 3DGS by 2.76~dB in PSNR while using over 60$\times$ less storage, and it renders at over 500~FPS---slower than 3DGS but far exceeding Plenoxels and Scaffold-GS.
This performance gain can be attributed to the hybrid structure and neural entropy coding of ECoNGS, which represent multiple local Gaussian primitives with a single compressed anchor point, resulting in a more compact scene representation.

When evaluated against compressed baselines (CCNeRF, HAC, and iVR-GS), our method achieves higher PSNR and a smaller model size while significantly reducing training time.
Compared with iVR-GS, the closest competitor in reconstruction quality, ECoNGS improves PSNR by 0.9--2.2~dB across all datasets while reducing model size by 3.9--6.1$\times$ and training time by 2.4--5.7$\times$.
HAC, which compresses each basic scene independently, fails to exploit cross-scene redundancy, resulting in lower compression ratios and noticeable quality degradation in complex scenes such as vortex.
CCNeRF achieves compact models but suffers from limited reconstruction quality and extremely slow rendering (below 1~FPS) due to NeRF-based volume sampling.
Although neural entropy coding introduces additional computational overhead during training, joint learning and sparse point cloud initialization greatly improve parallelism and convergence speed, resulting in a substantial reduction in total training time.

%--------------------------------------
\begin{figure*}[!htb]
 \begin{center}
 $\begin{array}{c@{\hspace{0.05in}}c@{\hspace{0.05in}}c@{\hspace{0.05in}}c@{\hspace{0.05in}}c}
 \includegraphics[width=0.185\linewidth]{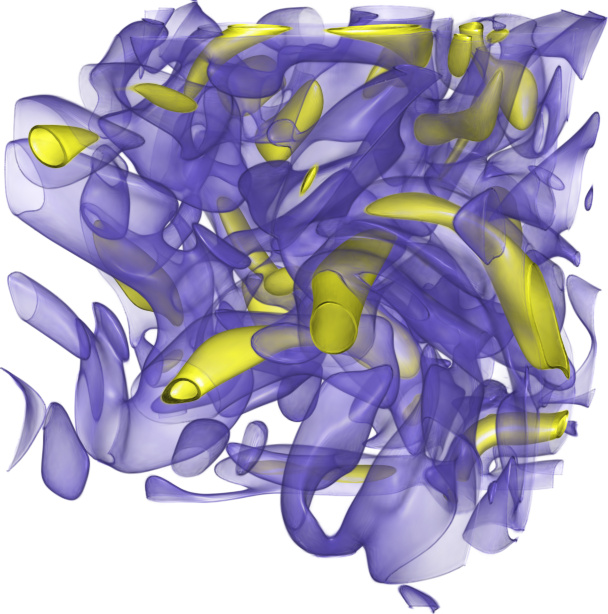}&
 \includegraphics[width=0.185\linewidth]{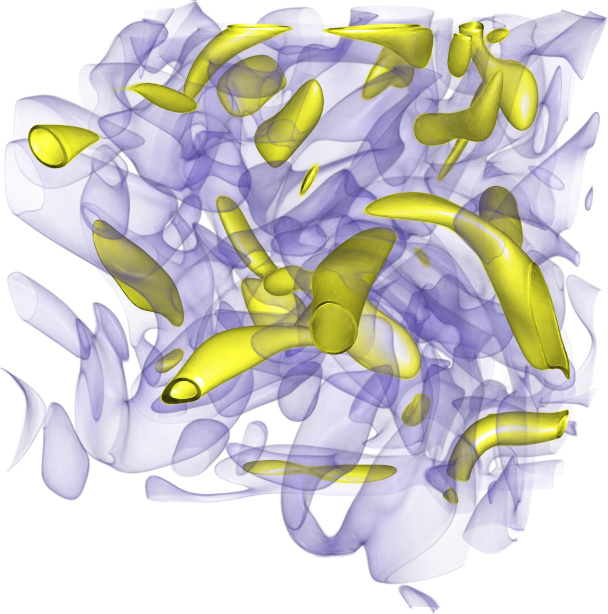}&
 \includegraphics[width=0.185\linewidth]{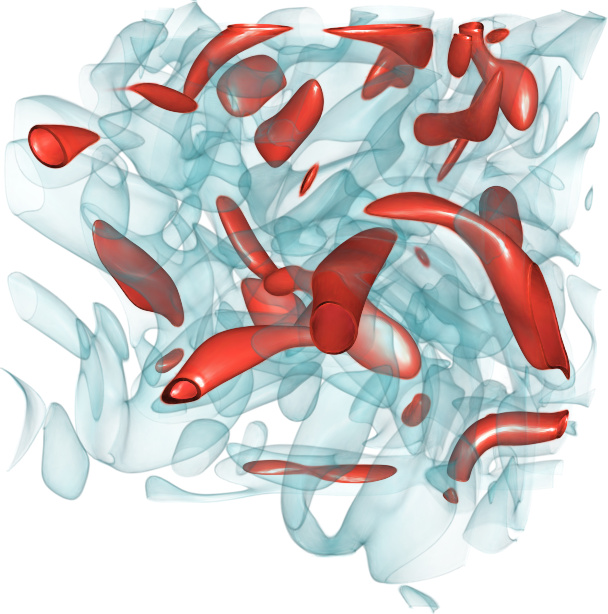}&
 \includegraphics[width=0.185\linewidth]{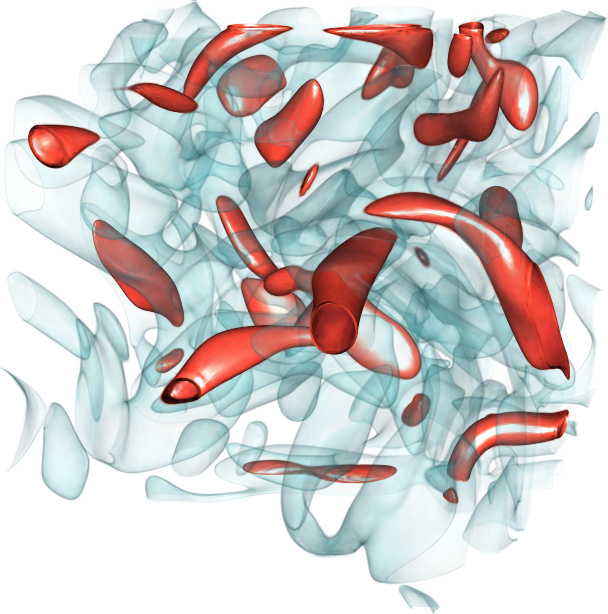}&
\includegraphics[width=0.185\linewidth]{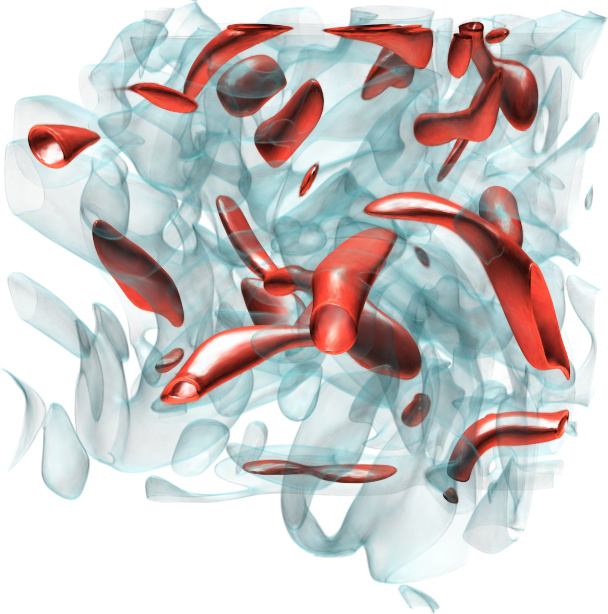}\\
\mbox{\small (a) original scene} & \mbox{\small (b): (a) + opacity change}& \mbox{\small (c): (b) + color change} & \mbox{\small (d): (c) + light mag.\ change} & \mbox{\small (e): (d) + light dir.\ change}\\
 \includegraphics[width=0.185\linewidth]{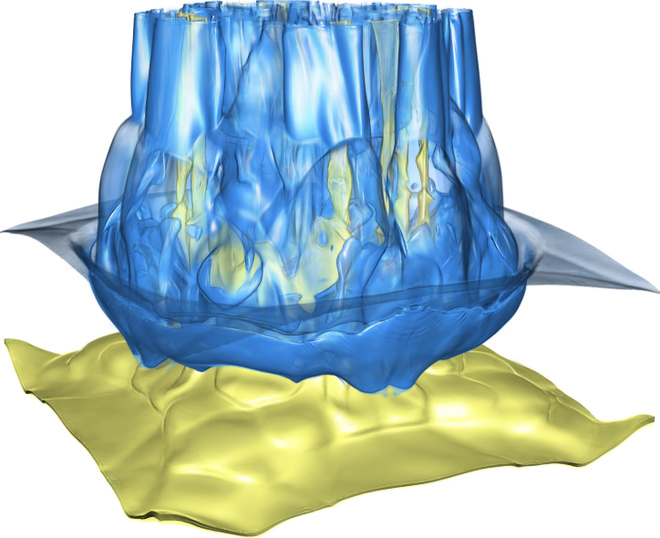}&
  \includegraphics[width=0.185\linewidth]{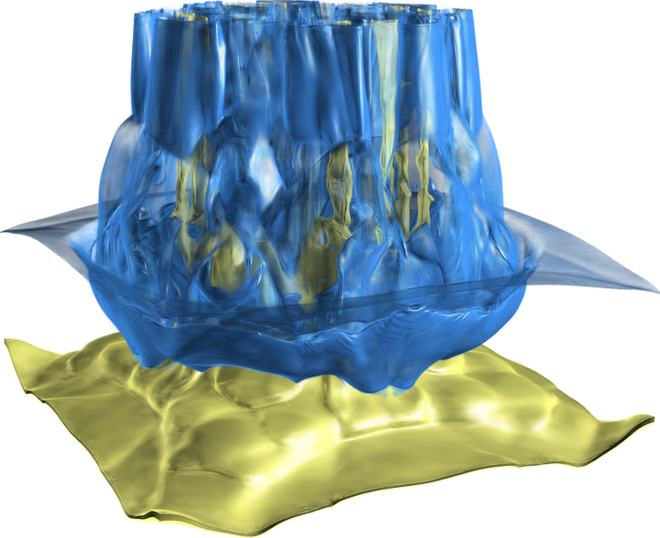}&
  \includegraphics[width=0.185\linewidth]{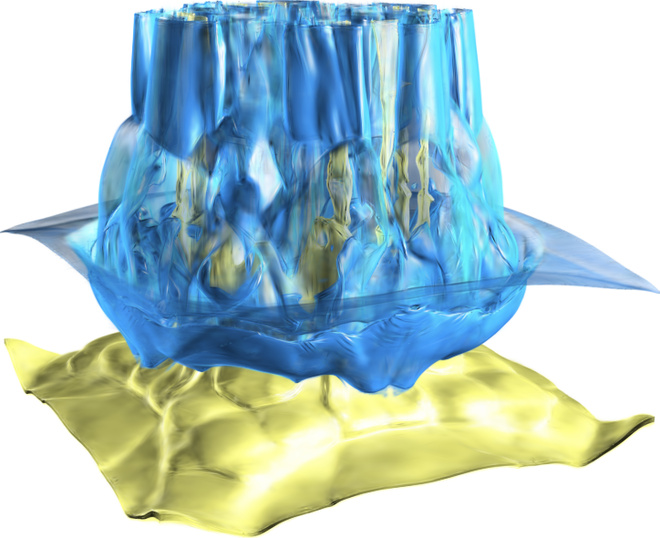}&
 \includegraphics[width=0.185\linewidth]{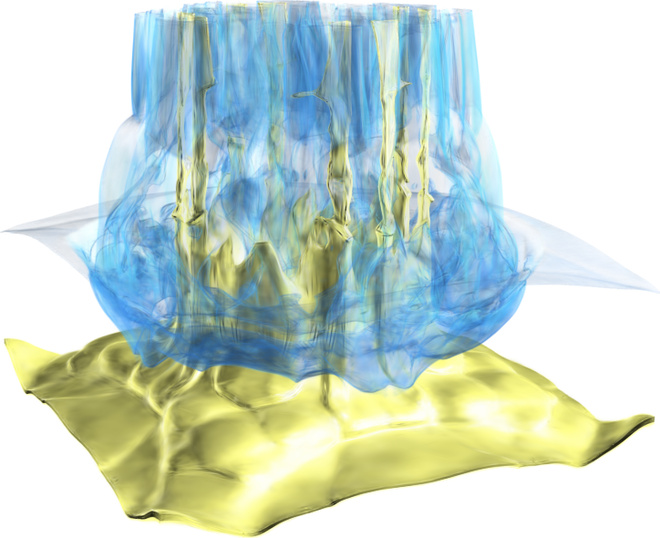}&
\includegraphics[width=0.185\linewidth]{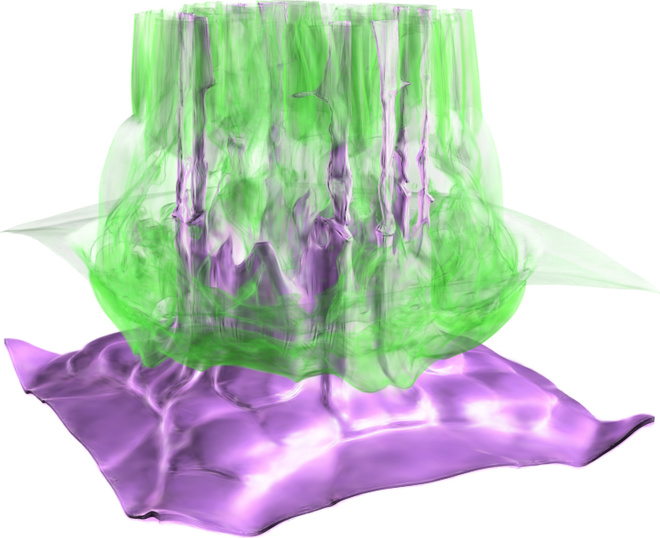}\\
\mbox{\small (f) selection of basic TFs} & \mbox{\small (g): (f) + light dir.\ change }& \mbox{\small (h): (g) + light mag.\ change} & \mbox{\small (i): (h) + opacity change} & \mbox{\small (j): (i) + color change}\\
\end{array}$
\end{center}
\vspace{-.25in} 
\caption{Examples of iterative scene editing results with composed ECoNGS models on the vortex and ionization (T) datasets. 
The light source moves from the front to the top of the scene from (d) to (e), and from the front to the right of the scene from (f) to (g).} 
\label{fig:scene-editing}
\vspace{-.1in}
\end{figure*}
%--------------------------------------

%--------------------------------------
\begin{figure}[!ht]
 \begin{center}
$\begin{array}{c@{\hspace{0.05in}}c}

\includegraphics[height=0.95in]{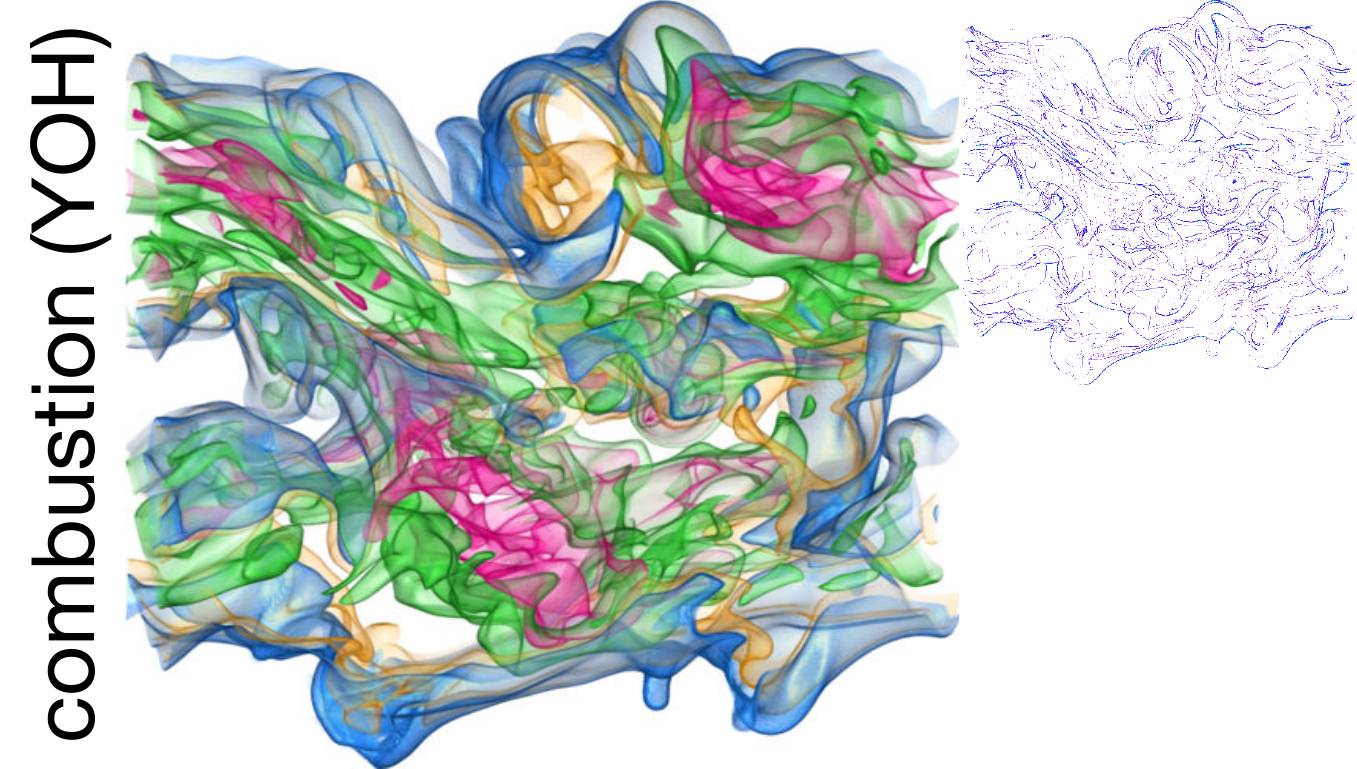} &
\includegraphics[height=0.95in]{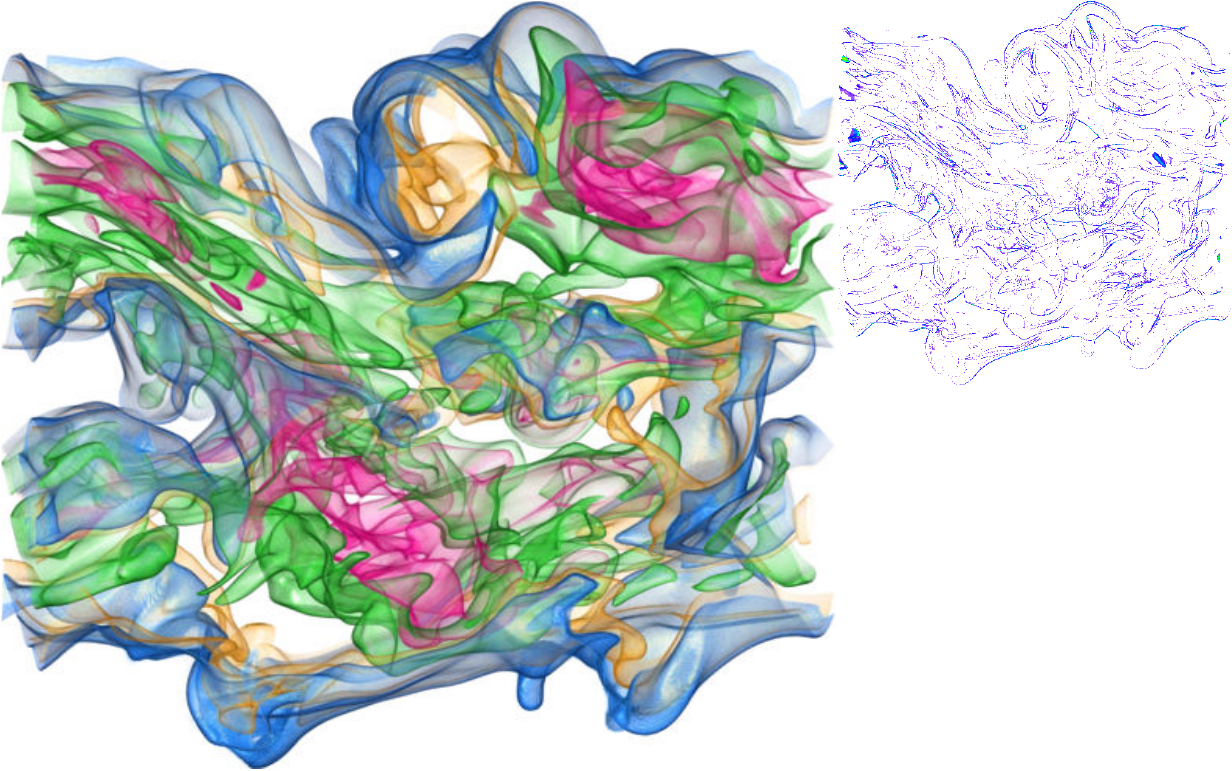} \\

\includegraphics[height=0.85in]{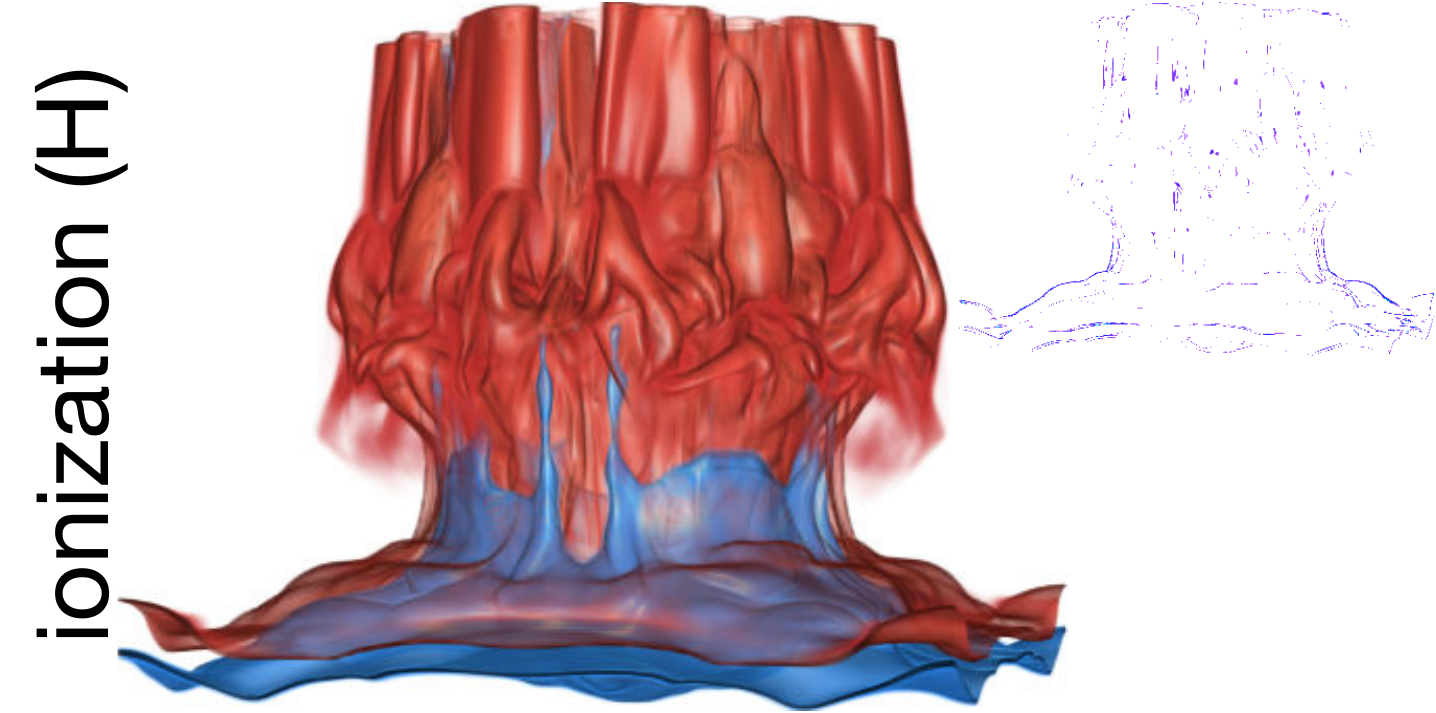} &
\includegraphics[height=0.85in]{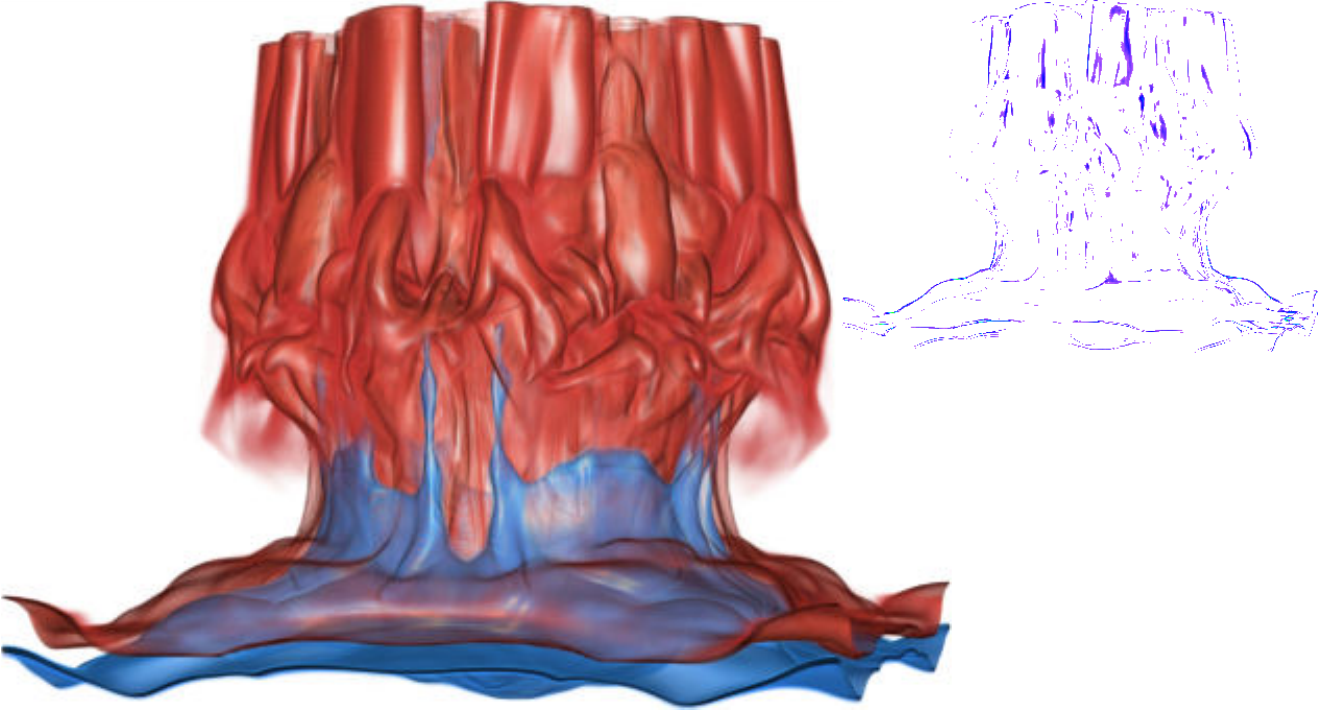} \\

\includegraphics[height=0.85in]{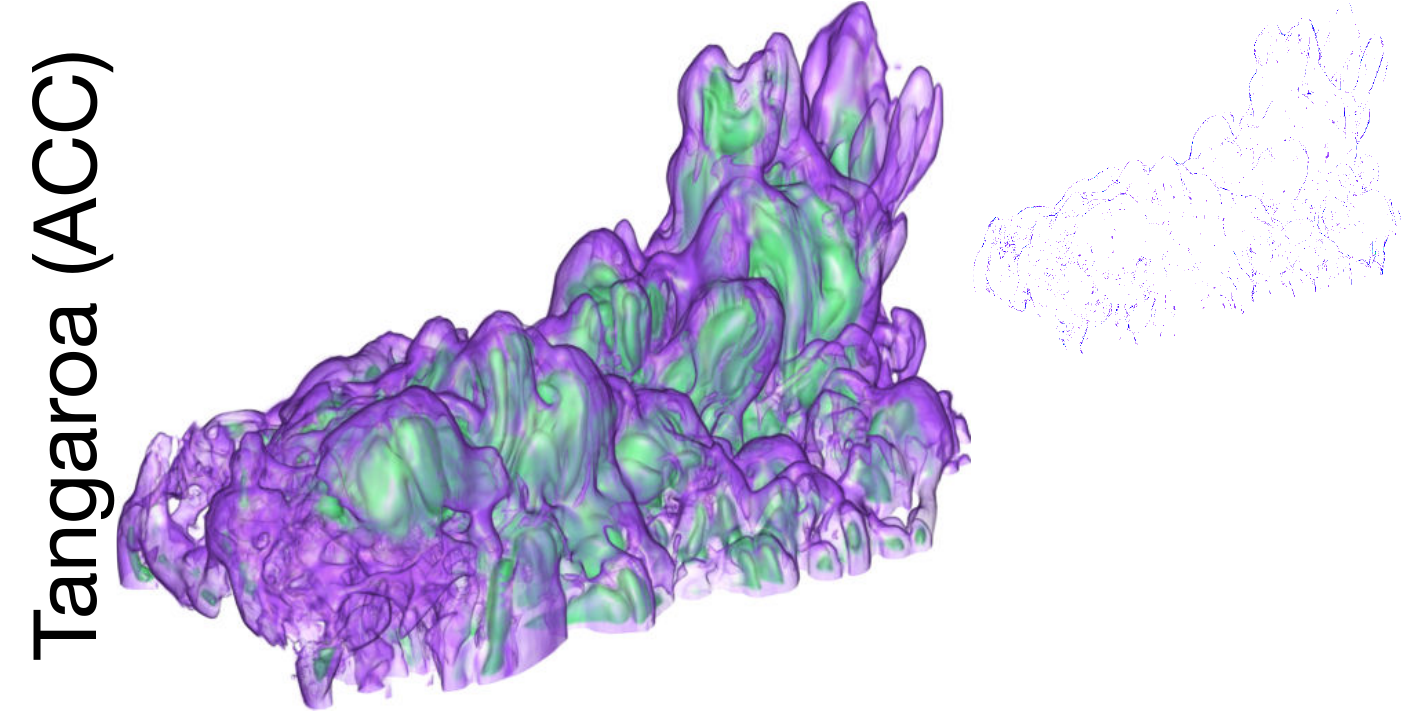} &
\includegraphics[height=0.85in]{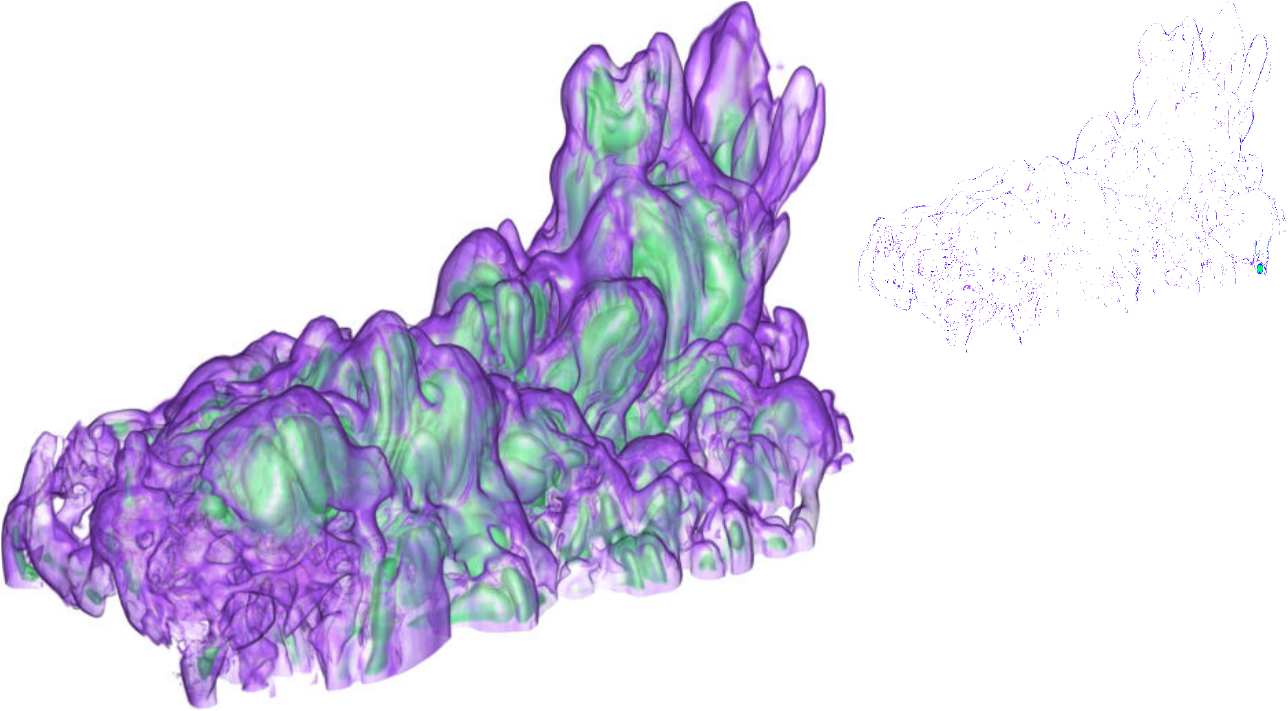} \\

\mbox{\small (a) ECoNGS} &
\mbox{\small (b) iVR-GS}

\end{array}$

\end{center}
\vspace{-.25in}
\caption{Comparing reconstruction results of ECoNGS and iVR-GS on multivariate datasets, with difference images shown in the top-right corner.
ECoNGS achieves 3.8--5.9$\times$ faster training and 4.2--6.0$\times$ smaller model size than iVR-GS across all multivariate datasets. %use a larger font for labels, they should look like the same size in the PDF
}
\label{fig:multivariate-baseline}
\vspace{-.1in}
\end{figure}

{\bf Qualitative analysis.}
To avoid visual occlusion across different basic scenes, we split the basic scenes in each dataset listed in Table~\ref{tab:univariate-datasets} into two groups for rendering and evaluation, as shown in Figures~\ref{fig:baseline-uncompress-results} and~\ref{fig:baseline-compress-results}, respectively.
The left side of each rendering displays the composed TF, with each opacity bump corresponding to the underlying basic TF from its respective basic scene.
% Figures~\ref{fig:baseline-uncompress-results} and~\ref{fig:baseline-compress-results} present the rendering results of all methods on novel views after composing the independently optimized models of each basic scene.
%\hot{Explain how the basic TFs are split between these two figures, referencing Table~\ref{tab:univariate-datasets}.} 

In Figure~\ref{fig:baseline-uncompress-results}, we compare ECoNGS with uncompressed baselines.
All models render considerably faster than DVR, with 3DGS achieving the highest frame rate.
This is due to the absence of the Blinn-Phong shading model in 3DGS, which makes it slightly faster than ECoNGS.
In addition, ECoNGS produces models that are much smaller than those from other methods.
Regarding reconstruction quality, ECoNGS achieves visually higher fidelity than other baseline approaches.
Plenoxels exhibits noticeable blurring of delicate structures due to its limited parameter grid resolution.
Scaffold-GS tends to overfit some training views when the scene is complex, leading to missing content in novel views (e.g., vortex), whereas 3DGS, which models view-dependent colors via SH coefficients instead of the Blinn-Phong shading model, performs less well in specular regions (e.g., the zoom-in areas of supernova and vortex) compared with ECoNGS.

Figure~\ref{fig:baseline-compress-results} compares the rendering results of ECoNGS with the compressed baselines.
ECoNGS achieves faster rendering and higher reconstruction quality while maintaining a smaller model size than all baselines.
Among the competing methods, CCNeRF, as a NeRF-based approach, suffers from noticeable blurring and stripe-like artifacts due to low-rank tensor decomposition.
Although employing neural Gaussians and neural entropy coding, HAC exhibits overfitting issues similar to those of Scaffold-GS on certain training views.
Furthermore, because each basic-scene model in HAC is independently compressed, redundant parameters across scenes reduce the overall compression ratio.
While iVR-GS produces results comparable to ECoNGS, the zoom-in regions reveal that iVR-GS performs less accurately in specular highlights.
This difference arises because the shading MLP in ECoNGS has greater capacity, enabling finer adjustment of the decoded shading attributes and thus more faithful rendering of lighting effects.

%--------------------------------------
\begin{table}[htb]
\caption{Average PSNR (dB), LPIPS, rendering framerate (FPS), total training time (TT, in hours), and model size (MS, in MB) on multivariate datasets. The best results are highlighted in bold.}
\vspace{-0.1in}
\centering
\resizebox{0.9\linewidth}{!}{%
\begin{tabular}{l  l c c c c c}
%\toprule
dataset & method & PSNR $\uparrow$ & LPIPS $\downarrow$ & FPS $\uparrow$ & TT $\downarrow$ & MS $\downarrow$ \\
\hline %\midrule
\multirow{2}{*}{combustion}
&iVR-GS&32.88&0.035&177.91&7.1&95.79 \\
&ECoNGS&\textbf{34.13}&\textbf{0.030}&\textbf{615.67}&\textbf{1.2}&\textbf{15.99} \\
\hline %\midrule
\multirow{2}{*}{ionization}
&iVR-GS&34.84&0.020&172.79&4.0&53.41 \\
&ECoNGS&\textbf{35.60}&\textbf{0.019}&\textbf{590.15}&\textbf{0.7}&\textbf{12.67} \\
\hline %\midrule
\multirow{2}{*}{Tangaroa}
&iVR-GS&32.84&0.021&130.66&1.5&33.13 \\
&ECoNGS&\textbf{34.08}&\textbf{0.019}&\textbf{553.74}&\textbf{0.4}&\textbf{5.89} \\
%\bottomrule
\end{tabular}
}
\label{tab:multi-variate-baseline}
\vspace{-.1in}
\end{table}
%--------------------------------------

\vspace{-0.075in}
\subsection{Iterative Scene Editing}

Like previous methods that represent VolVis scenes with editable Gaussian primitives, ECoNGS supports real-time editing of color, opacity, and lighting for VolVis scenes extracted from large-scale volume datasets.
Figure~\ref{fig:scene-editing} presents iterative scene editing results on the vortex and ionization (T) datasets.
After training, users can adjust various rendering settings during inference to achieve desired visualization effects and explore different appearance configurations interactively.

\vspace{-0.05in}
\subsection{Evaluation on Multivariate Datasets}

VolVis scenes with similar spatial structures can benefit from joint learning, thereby enhancing model performance.
This observation also holds for multivariate datasets, where different variables often share similar geometric and structural patterns.
To verify this, we compare ECoNGS with iVR-GS, another method based on editable Gaussian primitives, on the multivariate datasets.
Table~\ref{tab:multi-variate-baseline} reports the quantitative results, showing that ECoNGS consistently outperforms iVR-GS across all evaluation metrics.
On the largest multivariate combustion dataset (41 basic scenes, 5 variables), ECoNGS improves PSNR by 1.25~dB over iVR-GS while achieving 5.9$\times$ faster training and 6.0$\times$ smaller model size.
Similar advantages are observed for multivariate ionization and Tangaroa datasets, where ECoNGS consistently delivers higher reconstruction quality with significantly reduced training cost and storage requirements.
Moreover, the performance gap widens as the number of basic scenes increases.
This improvement stems from a larger set of similar basic scenes, enabling joint learning to enhance parallel optimization further and reduce redundant primitives.
ECoNGS also renders 3--4$\times$ faster than iVR-GS, thanks to the reduced number of anchors from joint learning.
Figure~\ref{fig:multivariate-baseline} presents the corresponding rendering results, where both ECoNGS and iVR-GS achieve high-fidelity reconstructions.
However, ECoNGS produces finer local details, a smaller model size, and a faster rendering framerate, demonstrating its efficiency and scalability for modeling complex multivariate volume datasets.

%--------------------------------------

\vspace{-0.075in}
\subsection{Ablation Studies}
\label{subsec:ablation}

\begin{figure}[!t]
 \begin{center}
$\begin{array}{c@{\hspace{0.2in}}c@{\hspace{0.2in}}c}
 \includegraphics[width=0.25\linewidth]{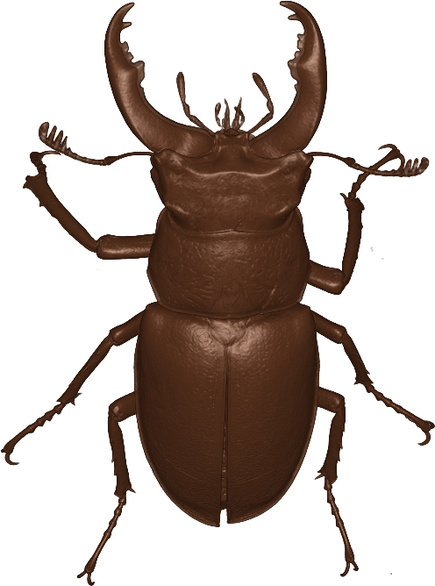}&
 \includegraphics[width=0.25\linewidth]{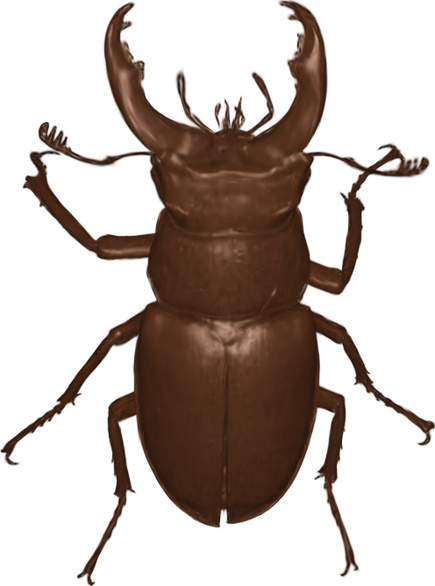}&
 \includegraphics[width=0.25\linewidth]{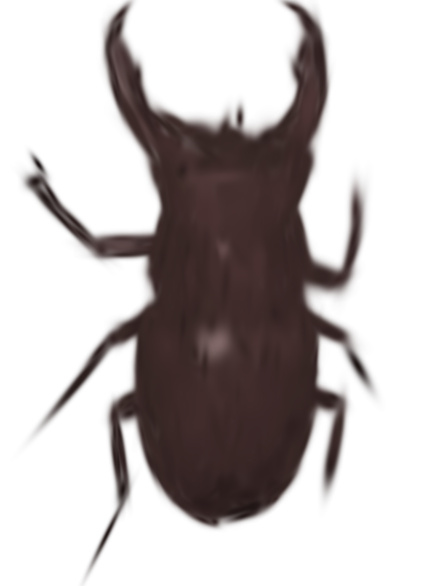}\\
 \mbox{\small (a) GT} & \mbox{\small (b) with our init} & \mbox{\small (c) with random init}
\end{array}$
\includegraphics[width=0.6\linewidth]{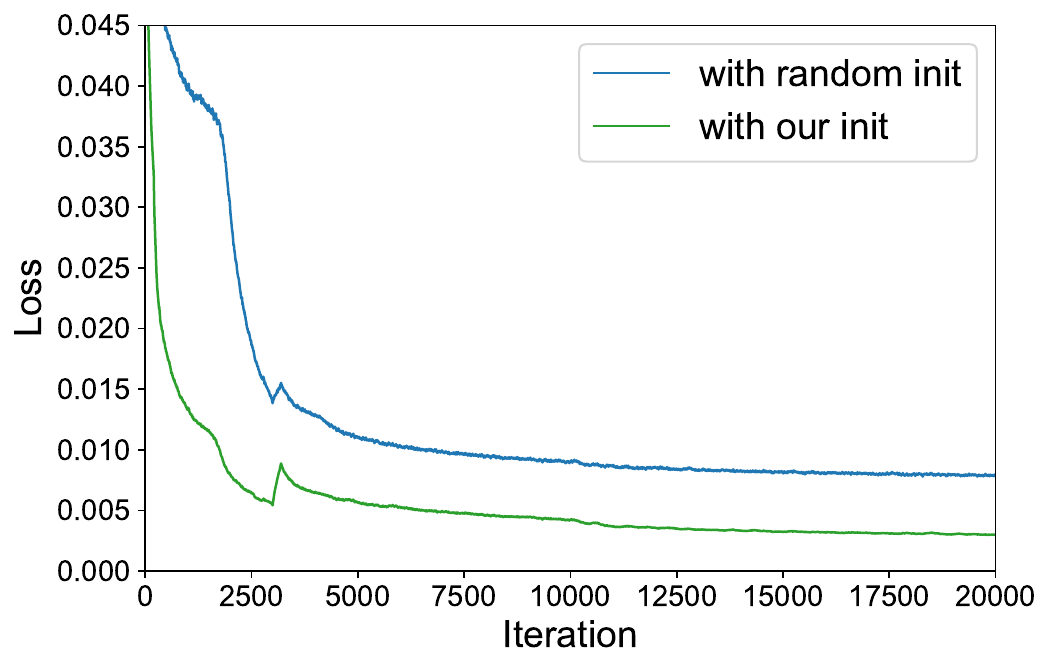}\\
 \mbox{\small (d) loss curves}\\
\end{center}
\vspace{-0.25in}
\caption{Comparing different initialization strategies on the beetle dataset. 
(b) and (c) are results from 1,000 iterations using our initialization and random initialization. 
(d) Loss curves of different initialization strategies, where our initialization (green curve) converges faster and achieves lower loss compared to random initialization (blue curve).}
\label{fig:ablation-point-cloud}
\vspace{-.1in}
\end{figure}

{\bf Evaluation on point cloud initialization.}
Before optimization, ECoNGS samples a sparse point cloud to initialize the anchor positions and partial attributes.
We conduct experiments on the beetle dataset to investigate how different point cloud generation strategies affect model initialization.
\hot{Tables~\ref{tab:ablation-point-cloud} and~\ref{tab:ablation-num-points} report the quantitative results of varying initialization settings and the number of sampled points, respectively.}
\hot{For initialization attributes, the position is the voxel position, and the color for each point is derived from the TF look-up.}
We observe that initializing all attributes yields the best reconstruction performance.
\hot{We further compare against random initialization, the default in 3DGS, at varying numbers of sampled points from 1k to 500k. As shown in Table~\ref{tab:ablation-num-points}, random initialization reconstruction accuracy saturates at around 32.6~dB PSNR regardless of the point count, whereas our point cloud initialization reaches up to 35.96~dB.}
In addition,
% the storage overhead remains negligible, as only 10k points are stored and the point cloud is saved in a binary-compressed format.
Moreover, increasing the number of points does not improve reconstruction accuracy but does increase the point cloud size.
\hot{In practice, we retain only 10k points to keep the storage overhead negligible while improving reconstruction accuracy.}
Figure~\ref{fig:ablation-point-cloud} illustrates the reconstructed results after 1,000 iterations and the corresponding training loss curves under different initialization strategies.
\hot{The temporary loss increase at 3,000 iterations is caused by adding the uniform quantization noise of Equation~\ref{eq:quantization_train}.}
It can be seen that point cloud initialization with position and color accelerates convergence and yields visually better reconstructions during the early training stage, providing a more stable foundation for subsequent context model learning.

%--------------------------------------
\begin{table}[!t]
\centering
\caption{\hot{Average PSNR (dB), LPIPS, and initialization file size (MB) on the beetle dataset under different initialization attribute settings.}
%The best PSNR and LPIPS are highlighted in bold.
}
\vspace{-0.1in}
\label{tab:ablation-point-cloud}
\resizebox{0.85\linewidth}{!}{%
\begin{tabular}{llccc}
 & setting & init file size $\downarrow$ & PSNR $\uparrow$ & LPIPS $\downarrow$ \\
\hline %\midrule
\multirow{3}{*}{init attributes}
 & \hot{random init}   & 0    & 32.57   &0.033 \\
 & + position  & 0.12 & 35.88  & 0.026\\
 & + color     & 0.15 & \textbf{35.96}   & \textbf{0.026} \\
\end{tabular}
}
\vspace{-.1in}
\end{table}
%--------------------------------------

%--------------------------------------
\begin{table}[!t]
\centering
\caption{\hot{Average PSNR (dB), LPIPS, and initialization file size (MB) on the beetle dataset under random initialization and our initialization with varying numbers of sampled points.}
%The best PSNR and LPIPS are highlighted in bold.
}
\vspace{-0.1in}
\label{tab:ablation-num-points}
\resizebox{0.85\linewidth}{!}{%
\begin{tabular}{llccc}
 & \hot{setting} & \hot{init file size $\downarrow$} & \hot{PSNR $\uparrow$} & \hot{LPIPS $\downarrow$} \\
\hline %\midrule
\multirow{2}{*}{\hot{500k points}} & \hot{random init} & \hot{0}    & \hot{32.63} & \hot{0.030} \\
                             & \hot{our init}   & \hot{7.2}  & \hot{35.41} & \hot{\textbf{0.024}} \\ \midrule
\multirow{2}{*}{\hot{100k points}} & \hot{random init} & \hot{0}    & \hot{32.67} & \hot{0.031} \\
                             & \hot{our init}   & \hot{1.4}  & \hot{35.59} & \hot{0.025} \\ \midrule
\multirow{2}{*}{\hot{10k points}}  & \hot{random init} & \hot{0}    & \hot{32.57} & \hot{0.033} \\
                             & \hot{our init}   & \hot{0.15} & \hot{\textbf{35.96}} & \hot{0.026} \\ \midrule
\multirow{2}{*}{\hot{1k points}}   & \hot{random init} & \hot{0}    & \hot{32.55} & \hot{0.034} \\
                             & \hot{our init}   & \hot{0.01} & \hot{35.88} & \hot{0.028} \\
\end{tabular}
}
\vspace{-.1in}
\end{table}
%--------------------------------------

%--------------------------------------
%--------------------------------------

%--------------------------------------
\begin{table}[htb]
\centering
\caption{Evaluation of the variant model design with the multivariate combustion dataset.
Average PSNR (dB), LPIPS, rendering framerate (FPS), total training time (TT, in minutes), total model size (MS, in MB), and total number of anchors are reported.}
\vspace{-0.1in}
\resizebox{1.0\linewidth}{!}{%
%\resizebox{\columnwidth}{!}{
%\setlength{\tabcolsep}{5pt}
\begin{tabular}{lcccccc}
model variant & PSNR $\uparrow$ &  LPIPS $\downarrow$ & FPS $\uparrow$ & TT $\downarrow$ & MS $\downarrow$ & \# anchors \\
\hline
full ECoNGS &34.13&0.030& 615.67 &72.0&15.99 & 415,275\\
w/o $\mathbf{v}_{\text{CA}}$ input &33.59&0.039& 614.82 &70.5&15.72 &420,822\\
w/o joint learning &34.60&0.031& 453.36 &156.9&45.75 &1,049,074\\
w/o context model &34.42&0.037& 605.66 &54.5&205.38 &409,398\\
\end{tabular}
}
\label{tab:model-design}
\vspace{-.1in}
\end{table}
%--------------------------------------

{\bf Evaluation on variant model design.}
We conduct an ablation study on the multivariate combustion dataset to evaluate the contribution of different components in the ECoNGS framework.
Table~\ref{tab:model-design} reports the quantitative results.
When the view-direction $\mathbf{v}_{\text{CA}}$ input is removed from the lightweight MLPs, PSNR drops by 0.54 dB, and LPIPS increases by 30\%, while the model size remains comparable, confirming that view-dependent decoding is critical for reconstruction accuracy.
In addition, without joint learning, training time increases by 2.2$\times$, the total number of anchors grows by 2.5$\times$, and the model size increases by 2.9$\times$, demonstrating the effectiveness of joint optimization in accelerating convergence and reducing redundant primitives.
Finally, when the context model for neural entropy coding is disabled, the model size increases by 12.8$\times$, confirming the crucial role of neural entropy coding in achieving a compact representation.
\hot{Overall, these components target complementary aspects of the framework and are therefore largely orthogonal: sparse point cloud initialization mainly accelerates convergence and improves reconstruction quality while barely affecting the model size (Table~\ref{tab:ablation-point-cloud} and Figure~\ref{fig:ablation-point-cloud}); joint learning mainly reduces the model size and rendering complexity by lowering the number of explicit anchors; and the context model mainly compresses the explicit anchor attributes, shrinking the model size by over $10\times$. Consequently, point cloud initialization and neural entropy coding can each be applied in isolation, whereas joint learning relies on a shared-MLP hybrid representation and similar scene clustering based on the point cloud.}

%--------------------------------------
\subsection{Rate-Distortion Analysis}
\label{subsec:rate-distortion}

%--------------------------------------
\begin{figure}[!t]
\centering
$\begin{array}{c@{\hspace{0.025in}}c}
\includegraphics[width=0.475\linewidth]{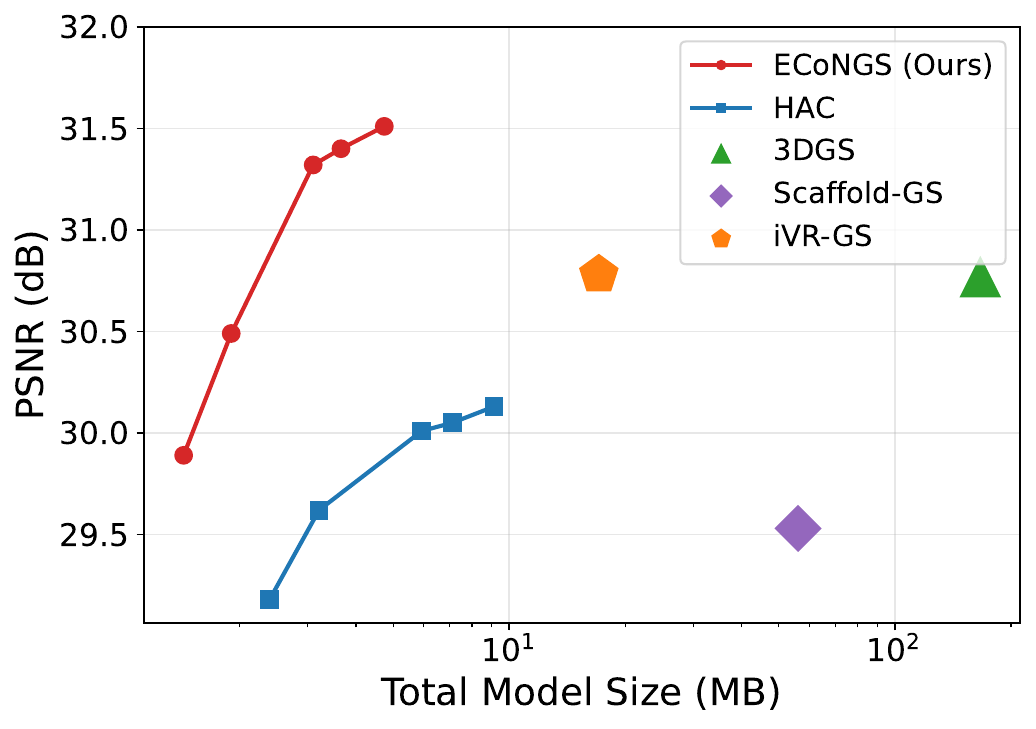}&
\includegraphics[width=0.475\linewidth]{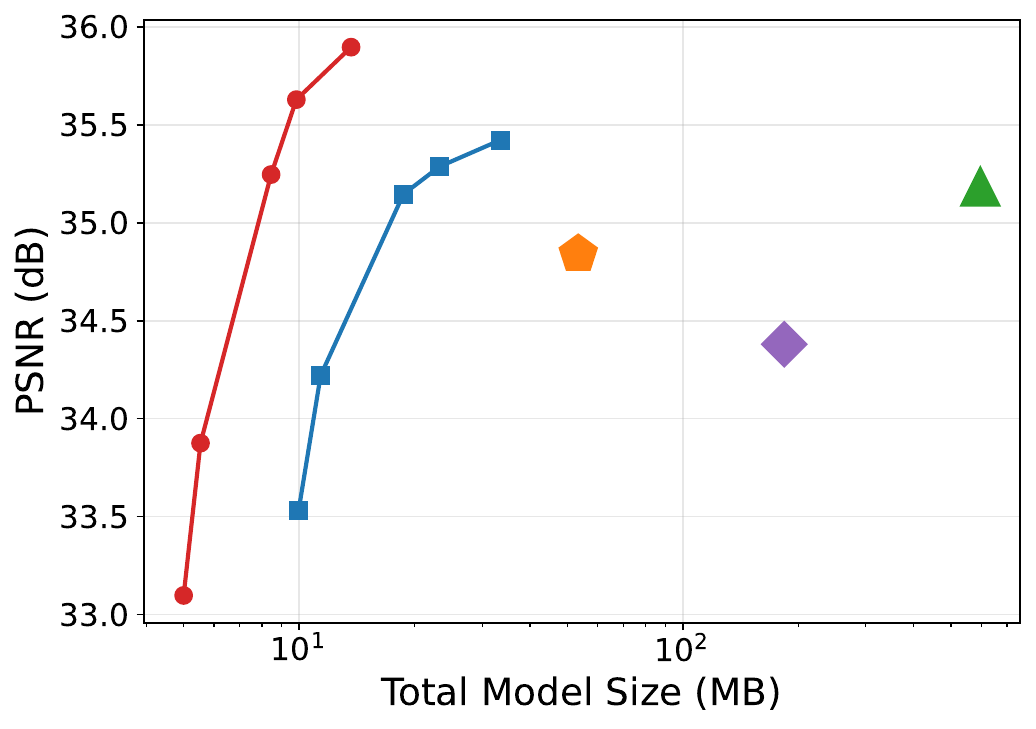}\\
\includegraphics[width=0.475\linewidth]{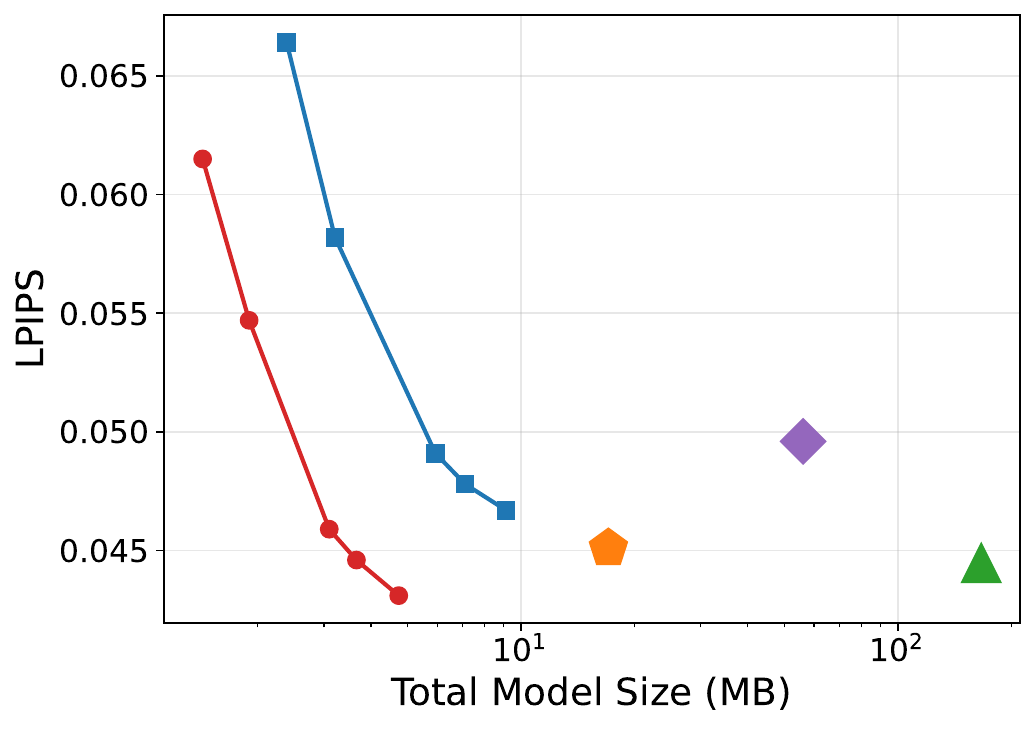}&
\includegraphics[width=0.475\linewidth]{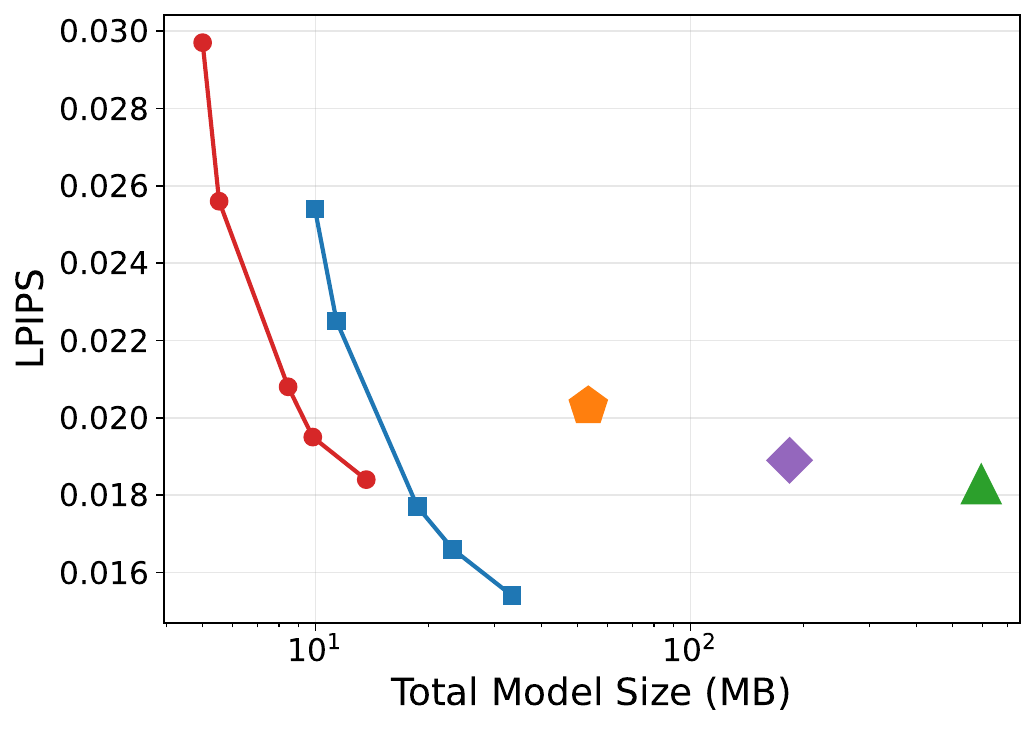}\\
 \mbox{\small supernova}&
 \mbox{\small ionization}\\
\end{array}$
\vspace{-.125in}
\caption{Rate-distortion curves of different GS-based methods on the univariate supernova and multivariate ionization datasets. The total model size is shown on a log scale for better visualization.}
\label{fig:RD-curves}
\end{figure}
%--------------------------------------

To evaluate the tradeoff between model size and reconstruction quality, we conduct a rate-distortion analysis by varying the entropy regularization weight $\lambda_e$ across five levels: 0.0001, 0.0005, 0.001, 0.005, and 0.01 for ECoNGS.
We then compare it with all GS-based methods, including uncompressed baselines (3DGS and Scaffold-GS), the fixed-rate compressed method iVR-GS, and HAC, which also supports tunable compression via entropy regularization, on the univariate supernova and multivariate ionization datasets.
Figure~\ref{fig:RD-curves} plots the average PSNR versus total model size at each compression level.
ECoNGS generally achieves higher PSNR than HAC at comparable model sizes across both datasets.
This is largely due to joint learning, which shares MLP parameters across basic scenes, thereby reducing redundant anchors.
At the lowest compression level, ECoNGS also outperforms the uncompressed 3DGS baseline in PSNR while using a much smaller model.
Under higher compression, ECoNGS still maintains reasonable reconstruction quality, suggesting that neural entropy coding is effective at preserving important scene information at lower bitrates.

\vspace{-0.075in}
\section{Conclusions and Future Work}

We have presented ECoNGS, an efficient compressive neural Gaussian splatting framework for VolVis.
ECoNGS introduces a hybrid neural-explicit representation that leverages lightweight MLPs to decode neural editable Gaussians from compact anchor points, greatly reducing redundancy while preserving fast training and rendering performance.
Through joint learning strategy and neural entropy coding, ECoNGS achieves superior reconstruction quality, smaller model size, and faster convergence than existing VolVis scene representation methods. %(iVR-GS, ScaffoldGS, and HAC).
In addition, our analysis of point cloud initialization provides new insights into the role of geometric priors in accelerating training stability and enhancing reconstruction fidelity.

% In the future, we plan to extend ECoNGS toward dynamic VolVis scenes, enabling shared neural Gaussians across time-varying datasets using a deformation network~\cite{Yang-CVPR2024} or 4D Gaussian primitive~\cite{Yang-ICLR2024}.
% We also aim to integrate ECoNGS with multimodal large language models to achieve more expressive scene editing or semantic control of volumetric scenes.
% Finally, deploying ECoNGS on resource-constrained or immersive visualization platforms (e.g., web and VR environments) will further demonstrate its potential for scalable, interactive exploration of large-scale volumetric data.

\hot{In the future, we plan to extend ECoNGS to dynamic VolVis scenes with shared neural Gaussians across time-varying datasets via a deformation network~\cite{Yang-CVPR2024} or 4D Gaussian primitives~\cite{Yang-ICLR2024}, integrate it with multimodal large language models for more expressive scene editing and semantic control, and deploy it on resource-constrained or immersive platforms such as web and virtual reality environments~\cite{Jeon-VIS26}.}

\appendix % You can use the `hideappendix` class option to skip everything after \appendix
\crefalias{section}{appendix} % this is to make sure that cleverref switches to referring to Appx. X from here on
%\newpage  % removed for arXiv layout: references now come after the appendix,
%\clearpage % so the appendix continues right after the acknowledgments

\setcounter{section}{0}
\setcounter{figure}{0}
\setcounter{table}{0}
%\setcounter{page}{1}

%\vspace{-0.05in}
%\section*{Appendix}

%\section{Additional Results}
%\label{sec:appendix-results}

\section{Scalability of Joint Learning}
\label{sec:scalability}

A key advantage of ECoNGS is its ability to jointly learn multiple basic scenes.
\hot{To analyze the scalability of this strategy, we study it from four aspects: the effect of the clustering threshold that controls how scenes are grouped, the behavior of joint learning across unrelated datasets, the clustering overhead as the number of basic scenes grows, and the joint learning gain across datasets with different numbers of basic scenes.}

%--------------------------------------
\begin{table}[htb]
\centering
\caption{Scalability analysis of joint learning on the multivariate combustion dataset. Average PSNR (dB), LPIPS, total training time (TT, in minutes), total model size (MS, in MB), and peak training GPU memory (GB) are reported under varying clustering thresholds for scene grouping.}
\vspace{-0.1in}
\resizebox{0.9\linewidth}{!}{%
\begin{tabular}{ccccccc}
 &  &  &  &  &  & peak GPU \\
threshold & \# groups & PSNR $\uparrow$ & LPIPS $\downarrow$ & TT $\downarrow$ & MS $\downarrow$ & memory \\
\hline
 0    & 41 & 34.60 & 0.031 & 156.9 & 45.75 & 6.1\\
 0.005 & 19 & 34.30 & 0.030 & 95.8 & 19.03 & 12.8 \\
 0.01 & 10 & 34.13 & 0.030 & 70.5 & 15.72 & 16.1 \\
 0.02 & 7  & 34.02 & 0.035 & 67.6 & 14.72 & 23.3 \\
 0.04 & 3 (OOM) & --- & --- & --- & --- & $>$24 \\
\end{tabular}
}
\label{tab:scalability}
\vspace{-.1in}
\end{table}
%--------------------------------------

\hot{\noindent\textbf{Effect of the clustering threshold.}}
Table~\ref{tab:scalability} presents scalability results for the multivariate combustion dataset across varying Chamfer distance clustering thresholds.
A larger threshold groups more basic scenes together, reducing the number of jointly trained groups.
As the clustering threshold increases from 0 to 0.02, the number of groups decreases from 41 to 7, resulting in a 2.3$\times$ reduction in training time and a 3.1$\times$ reduction in model size, while PSNR drops slightly.
This indicates that the shared MLP has sufficient capacity to accommodate diverse basic scenes from multivariate volumetric data within each group without obvious degradation in quality.
However, when the clustering threshold is set too high (e.g., 0.04) and joint learning is applied across more than 20 basic scenes, it becomes difficult to simultaneously load all anchor and optimizer parameters, as well as training images, into GPU memory, leading to out-of-memory (OOM) errors during training.
Note that the reported peak GPU memory corresponds to the joint training phase, which is the most memory-intensive stage. 
During inference, the rendering cost is significantly lower (less than 2 GB), since only the trained model needs to be loaded.

%--------------------------------------
\begin{table}[htb]
\centering
\caption{Cross-dataset joint learning. Per-dataset PSNR (dB), LPIPS, and total training time (TT, in minutes) are reported.}
\vspace{-0.1in}
\resizebox{0.9\linewidth}{!}{%
\begin{tabular}{lccccc}
 & \multicolumn{2}{c}{PSNR $\uparrow$} & \multicolumn{2}{c}{LPIPS $\downarrow$} & \\
training setup & beetle & argon bubble & beetle & argon bubble & TT $\downarrow$ \\
\hline
beetle              & 35.96 & ---   & 0.026 & ---   & 4.2 \\
argon bubble        & ---   & 31.76 & ---   & 0.015 & 4.8 \\
joint  & 34.26 & 16.82 & 0.027 & 0.225 & 5.2 \\
\end{tabular}
}
\label{tab:cross-dataset}
\vspace{-.1in}
\end{table}
%--------------------------------------

\hot{\noindent\textbf{Cross-dataset joint learning.}}
To further examine the scalability of joint learning, we test whether scenes from entirely unrelated datasets can be trained together.
We jointly train basic scenes from the beetle and argon bubble datasets, and compare with training each dataset separately.
As shown in Table~\ref{tab:cross-dataset}, joint learning across unrelated datasets results in a degradation in quality.
This is because the shared MLPs lack the capacity to simultaneously encode fundamentally different volume structures, confirming that scene grouping based on Chamfer distance is essential for effective joint learning.

%--------------------------------------
\begin{table}[htb]
\centering
\caption{\hot{Joint learning vs.\ separate (per-scene) optimization on different datasets. Average PSNR (dB), LPIPS, total training time (TT, in minutes), total model size (MS, in MB), and total number of anchors are reported.}}
\vspace{-0.1in}\hot{
\resizebox{\linewidth}{!}{%
\begin{tabular}{lcccccccc}
dataset & \# scenes & method & PSNR $\uparrow$ & LPIPS $\downarrow$ & TT $\downarrow$ & MS $\downarrow$ & \# anchors $\downarrow$ \\
\hline
\multirow{2}{*}{ionization (T)} & \multirow{2}{*}{4} & separate & \textbf{34.53} & \textbf{0.017} & 15.9 & 4.39 & 98,786 \\
                                    & & joint    & 34.34 & 0.019 & \textbf{10.36} & \textbf{3.32} & \textbf{55,306} \\
\hline
\multirow{2}{*}{combustion (MF)} & \multirow{2}{*}{9} & separate & \textbf{35.04} & \textbf{0.019} & 34.4 & 9.93 & 188,990 \\
                                    & & joint    & 34.70 & 0.021 & \textbf{18.04} & \textbf{4.94} & \textbf{98,174} \\
\hline
\multirow{2}{*}{ionization} & \multirow{2}{*}{24} & separate & \textbf{35.95} & \textbf{0.016} & 85.2 & 20.83 & 515,049 \\
                                    & & joint    & 35.60 & 0.019 & \textbf{42.8} & \textbf{12.67} & \textbf{257,943} \\
\hline
\multirow{2}{*}{combustion} & \multirow{2}{*}{41} & separate & \textbf{34.60} & 0.031 & 156.9 & 45.75 & 1,049,074 \\
                                    & & joint    & 34.13 & \textbf{0.030} & \textbf{72.0} & \textbf{15.99} & \textbf{415,275} \\
\end{tabular}
}}
\label{tab:joint-gain}
\vspace{-.1in}
\end{table}
%--------------------------------------

\hot{\noindent\textbf{Joint learning gain vs.\ number of basic scenes.}
To examine how the benefit of joint learning depends on the number of basic scenes, we further compare joint learning against the separate baseline (each basic scene optimized independently with the same hyperparameters) on four datasets of increasing scale. Table~\ref{tab:joint-gain} reports the average PSNR, LPIPS, total training time, total model size, and total number of anchors. Across all datasets, joint learning substantially reduces training time, model size, and the number of anchors while keeping reconstruction quality nearly unchanged. These savings grow with the number of basic scenes: the training-time reduction increases monotonically from $35\%$ (4 scenes) to $48\%$ (9 scenes), $50\%$ (24 scenes), and $54\%$ (41 scenes), and the anchor reduction grows from $44\%$ to $60\%$, since more scenes expose more inter-scene redundancy for the shared MLPs and context model to exploit.
Even when the number of basic scenes is small, joint learning still reduces training time, model size, and the number of anchors, indicating that the strategy is broadly beneficial rather than effective only at the VolVis scene with a large number of basic scenes.}

%--------------------------------------
\begin{table}[htb]
\centering
\caption{\hot{Clustering overhead of joint learning across datasets of increasing scale. We report the number of basic scenes, clustering time (CT, in seconds), and total training time (TT, in minutes).}}
\vspace{-0.1in}\hot{
\resizebox{0.6\linewidth}{!}{%
\begin{tabular}{lccc}
dataset & \# basic scenes & CT  & TT \\
\hline
vortex      & 4   & 0.01 & 9.3 \\
Tangaroa    & 8   & 0.03 & 20.9 \\
ionization  & 24  & 0.18 & 42.9 \\
combustion  & 41  & 0.49 & 72.0 \\
\end{tabular}
}}
\label{tab:clustering-cost}
\vspace{-.1in}
\end{table}
%--------------------------------------

\hot{\noindent\textbf{Clustering overhead vs. number of basic scenes.}
Before joint learning, we group geometrically similar scenes by computing pairwise Chamfer distances between their point clouds and applying hierarchical clustering. This step requires a quadratic number of pairwise comparisons in the number of basic scenes. However, the point clouds used for clustering and initialization are randomly subsampled to at most 10K points, thereby reducing the clustering cost in practice. To quantify this overhead, Table~\ref{tab:clustering-cost} reports the clustering time together with the total training time for datasets of increasing scale. Even for the multivariate combustion dataset with 41 basic scenes, clustering completes in less than half a second, accounting for less than $0.02\%$ of the total training time. These results indicate that the clustering overhead is negligible in our experiments and is well justified by the training-time and model-size reductions enabled by joint learning.}

%--------------------------------------
\begin{table*}[htb]
\centering
\caption{\hot{Comparing ECoNGS with triangular mesh on the isosurfaces extracted from the combustion (MF) volume at three isovalues. We report the corresponding model size (MB), rendering frame rate (FPS), and render-time GPU/host memory usage (GB) for both methods, as well as the number of triangles (in millions) for the triangle mesh, and PSNR (in dB) and LPIPS for ECoNGS. 
The three basic scenes are trained jointly; each basic-scene MS includes the shared decoder MLPs, so all basic model sizes do not sum to the composed MS.}}
\vspace{-0.1in}\hot{
\resizebox{0.75\linewidth}{!}{%
\begin{tabular}{l|ccccc|cccccc}
 \multicolumn{1}{c}{}& \multicolumn{4}{c}{Triangular mesh} & \multicolumn{6}{c}{ECoNGS} \\
scene & \# triangles & MS $\downarrow$  & FPS $\uparrow$ & GPU & host & PSNR $\uparrow$ & LPIPS $\downarrow$ & MS $\downarrow$ & FPS $\uparrow$ & GPU & host \\
\hline
iso $=-0.5$ & 17.1 & 317.5 & 101 & 0.31 & 1.86 & 36.21 & 0.014 & 0.65 & 468 & 3.30 & 7.61 \\
iso $=0.0$  & 20.9 & 388.6 & 101 & 0.33 & 1.87 & 34.16 & 0.016 & 0.66 & 470 & 3.29 & 7.61 \\
iso $=+0.5$ & 15.9 & 296.8 & 100 & 0.32 & 1.82 & 35.63 & 0.011 & 0.64 & 471 & 3.27 & 7.60 \\
\hline
composed & 53.9 & 1002.9 & 101 & 0.95 & 4.92 & 32.20 & 0.025 & 1.57 & 331 & 3.43 & 7.63 \\
\end{tabular}
}}
\label{tab:iso-mesh}
\vspace{-.1in}
\end{table*}
%--------------------------------------

%--------------------------------------
\begin{figure}[htb]
 \begin{center}
$\begin{array}{c@{\hspace{0.02in}}c@{\hspace{0.02in}}c@{\hspace{0.02in}}c@{\hspace{0.02in}}c}

 &
\mbox{\small \hot{317.5\,MB}} &
\mbox{\small \hot{388.6\,MB}} &
\mbox{\small \hot{296.8\,MB}} &
\mbox{\small \hot{1002.9\,MB}} \\

\rotatebox{90}{\makebox[0.215\linewidth][c]{\small \hot{Triangular mesh}}} &
\includegraphics[width=0.225\linewidth]{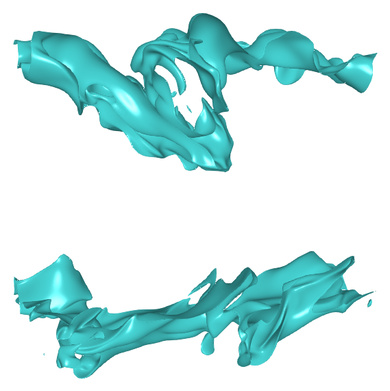} &
\includegraphics[width=0.225\linewidth]{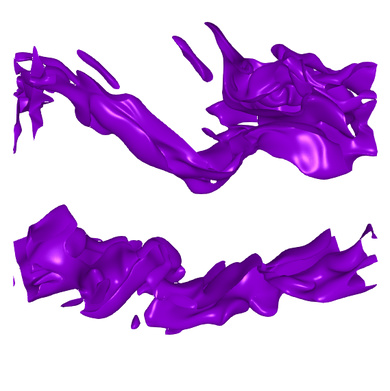} &
\includegraphics[width=0.225\linewidth]{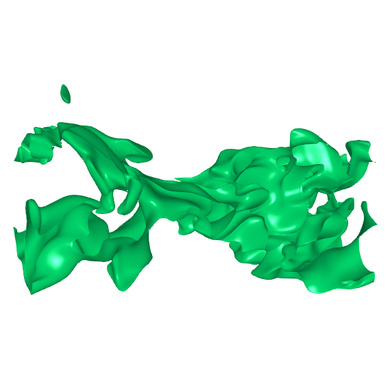} &
\includegraphics[width=0.225\linewidth]{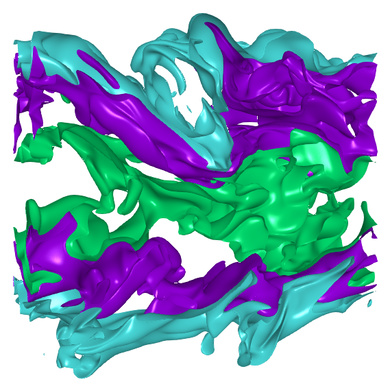} \\

 &
\mbox{\small \hot{0.65\,MB}} &
\mbox{\small \hot{0.66\,MB}} &
\mbox{\small \hot{0.64\,MB}} &
\mbox{\small \hot{1.57\,MB}} \\

\rotatebox{90}{\makebox[0.215\linewidth][c]{\small \hot{ECoNGS}}} &
\includegraphics[width=0.225\linewidth]{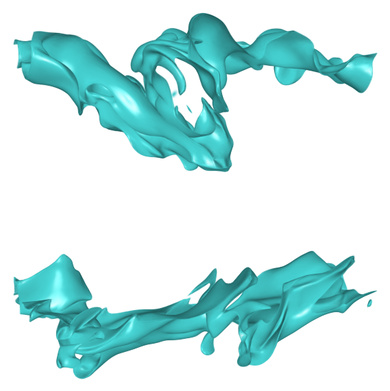} &
\includegraphics[width=0.225\linewidth]{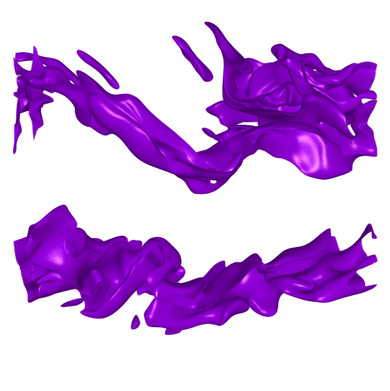} &
\includegraphics[width=0.225\linewidth]{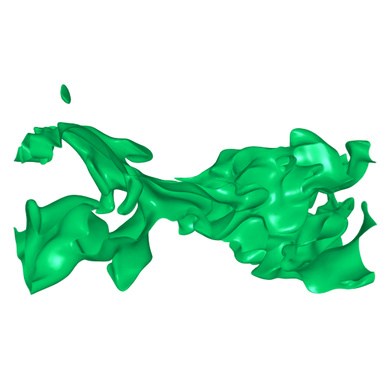} &
\includegraphics[width=0.225\linewidth]{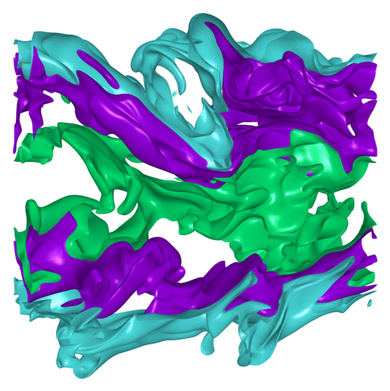} \\

 &
\mbox{\small \hot{(a) iso\,$=-0.5$}} &
\mbox{\small \hot{(b) iso\,$=0.0$}} &
\mbox{\small \hot{(c) iso\,$=+0.5$}} &
\mbox{\small \hot{(d) composed}}
\end{array}$
\end{center}
\vspace{-.25in}
\caption{\hot{Comparing ECoNGS with triangular meshes on the isosurface extracted from the combustion (MF) volume.  (a)--(c) the three basic scenes extracted with different isovalues, and (d) is the composed scene. ECoNGS preserves high visual quality but is more compact than the original triangular mesh representation.}}
\label{fig:iso-mesh}
\vspace{-.1in}
\end{figure}
%--------------------------------------

%--------------------------------------
\begin{table}[htb]
\centering
\caption{\hot{Average PSNR (dB), LPIPS, total model size (MS, in MB), and rendering framerate (FPS) for two fuzzy basic scenes and the composed scene of the vortex dataset.}}
\vspace{-0.1in}\hot{
\resizebox{0.6\linewidth}{!}{%
\begin{tabular}{lcccc}
scene & PSNR  & LPIPS  & MS & FPS  \\
\hline
TF-blue & 47.12 & 0.029 & 0.33 & 491 \\
TF-red  & 50.77 & 0.008 & 0.31 & 491 \\
\hline
composed & 45.84 & 0.024 & 0.45 & 492 \\
\end{tabular}
}}
\label{tab:fuzzy-tf}
\vspace{-.1in}
\end{table}
%--------------------------------------

%--------------------------------------
\begin{figure}[htb]
 \begin{center}
%\resizebox{\linewidth}{!}{%
$\begin{array}{c@{\hspace{0.02in}}c@{\hspace{0.02in}}c@{\hspace{0.02in}}c}

\rotatebox{90}{\makebox[0.215\linewidth][c]{\small \hot{GT}}} &
\includegraphics[height=0.7in]{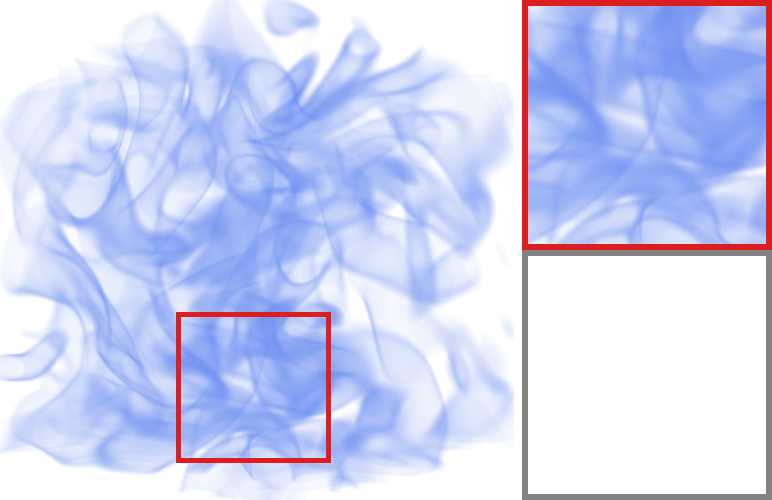} &
\includegraphics[height=0.7in]{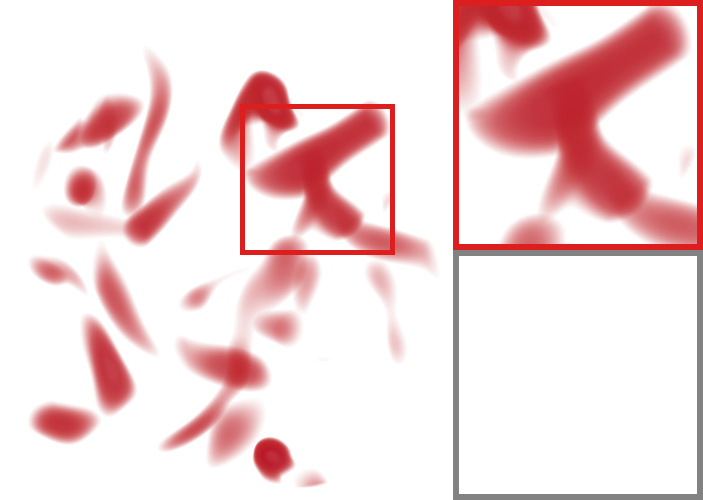} &
\includegraphics[height=0.7in]{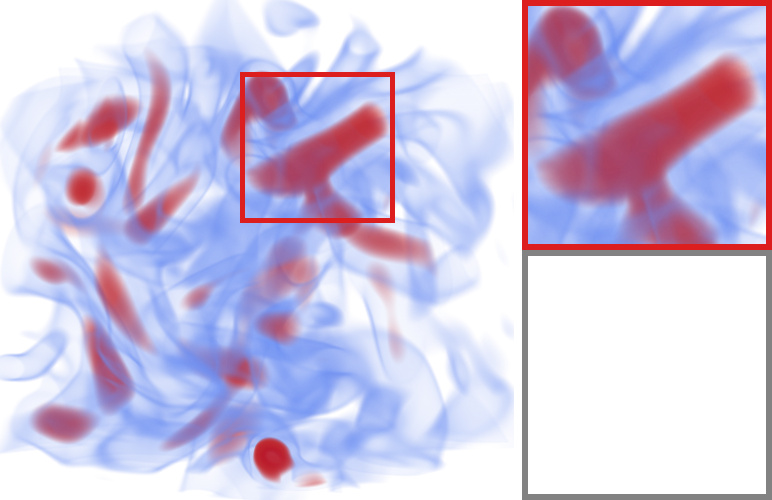} \\

\rotatebox{90}{\makebox[0.215\linewidth][c]{\small \hot{ECoNGS}}} &
\includegraphics[height=0.7in]{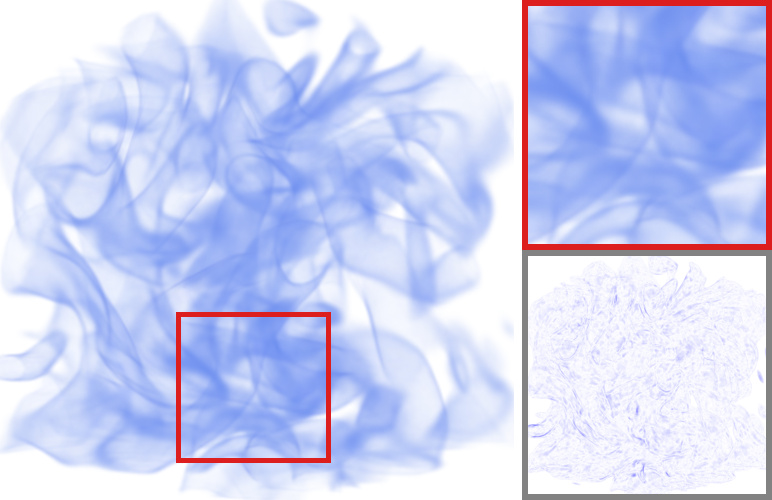} &
\includegraphics[height=0.7in]{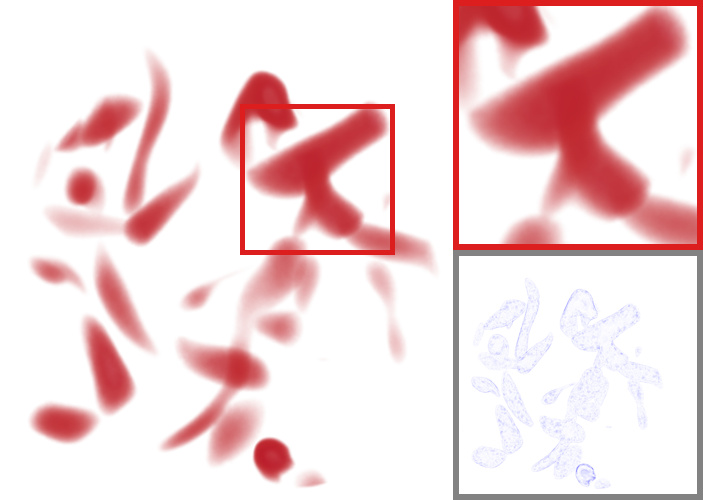} &
\includegraphics[height=0.7in]{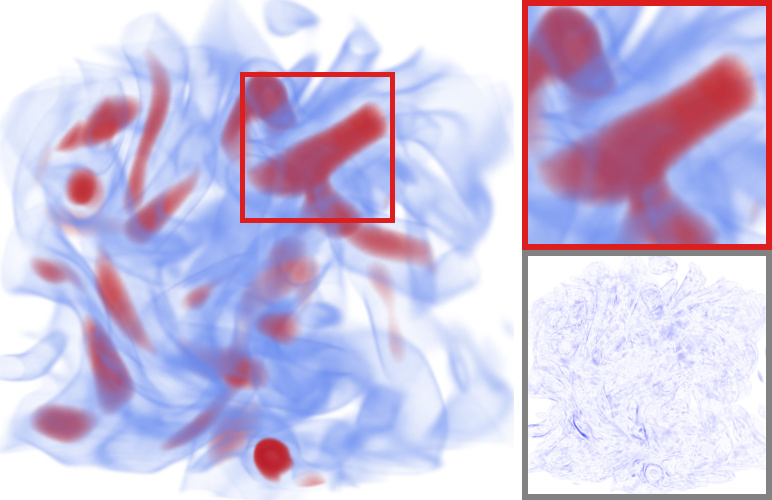} \\

 &
\mbox{\hot{\small (a) TF-blue}} &
\mbox{\hot{\small (b) TF-red}} &
\mbox{\hot{\small (c) composed}}
\end{array}$
%}
\end{center}
\vspace{-.25in}
\caption{\hot{Novel view synthesis results of ECoNGS on fuzzy VolVis scene of the vortex dataset. The difference image in the bottom-right corner shows the pixel-wise perceptible difference (blue to red indicates low to high) in the CIELUV color space.}}
\label{fig:fuzzy-tf}
\vspace{-.1in}
\end{figure}
%--------------------------------------

\vspace{-0.05in}
\section{\hot{Evaluation on Isosurface Scenes}}
\label{sec:iso-mesh}

\hot{Beyond volume rendering scenes included in the main paper, ECoNGS can also represent isosurface scenes.
To understand how ECoNGS differs from, and where it improves upon, the conventional approach of directly extracting a triangular mesh, we evaluate ECoNGS against the triangular mesh solution on the combustion (MF) dataset, extracting three isosurfaces at different isovalues over the value range $[-1, 1]$ to form the basic scenes for comparison.
Figure~\ref{fig:iso-mesh} and Table~\ref{tab:iso-mesh} report the qualitative and quantitative results, respectively.
As shown, ECoNGS reconstructs the isosurface scenes with high fidelity while keeping the model size small. It therefore serves as a scene representation that is far more compact than the triangular meshes extracted directly from the large-scale volume data.
One minor downside is that ECoNGS has a slightly higher GPU/host memory footprint during rendering than the mesh. However, the memory cost stays roughly constant regardless of scene complexity and is negligible on modern GPUs.}

\vspace{-0.05in}
\section{\hot{Evaluation on Fuzzy VolVis Scenes}}
\label{sec:fuzzy-tf}

\hot{Unless otherwise specified, all VolVis scenes are rendered using ParaView's NVIDIA IndeX plugin with depth enhancement~\cite{Zheng-TVCG13}, which provides lighting and clear boundaries for high-quality volume visualization.
However, ECoNGS can also work on fuzzy VolVis scenes.
To this end, we apply two TFs to the vortex dataset and turn off depth enhancement to create fuzzy VolVis scenes.
Figure~\ref{fig:fuzzy-tf} and Table~\ref{tab:fuzzy-tf} present the qualitative and quantitative results for the corresponding basic scenes and the composed scene.
We observe that ECoNGS achieves high quality for both basic scenes and the composed scene. 
Comparing Table~\ref{tab:fuzzy-tf} with the reconstruction accuracy of scenes rendered with depth enhancement, the metric values for the fuzzy VolVis scene are significantly higher. 
This is mainly because turning off depth enhancement reduces strong surface-lighting effects and sharp boundary cues, thereby alleviating the need to model complex normal-dependent shading variations.
Consequently, the fuzzy scenes are comparatively easier to fit during optimization.}

%--------------------------------------
\begin{table}[!h]
\centering
\caption{Comparison of ECoNGS and iVR-GS under limited training views on the argon bubble dataset. Average PSNR (dB), LPIPS, total training time (TT, in minutes), and total model size (MS, in MB) are reported under varying training views.}
\vspace{-0.1in}
\resizebox{0.75\linewidth}{!}{%
\begin{tabular}{cccccc}
\# views & method & PSNR $\uparrow$ & LPIPS $\downarrow$ & TT $\downarrow$ & MS $\downarrow$ \\
\hline
\multirow{2}{*}{6} & iVR-GS & 17.31 & 0.192 & 15.0 & 3.63 \\
 & ECoNGS & 24.07 & 0.051 & 4.5 & 1.42 \\
\hline
\multirow{2}{*}{12} & iVR-GS & 25.39 & 0.034 & 15.4 & 3.86 \\
 & ECoNGS & 25.34 & 0.035 & 4.6 & 1.37 \\
\hline
\multirow{2}{*}{42} & iVR-GS & 30.12 & 0.016 & 16.2 & 4.38 \\
 & ECoNGS & 30.35 & 0.015 & 9.3 & 1.42 \\
\hline
\multirow{2}{*}{92} & iVR-GS & 31.30 & 0.015 & 15.1 & 4.25 \\
 & ECoNGS & 31.76 & 0.015 & 9.3 & 1.27 \\
\end{tabular}
}
\label{tab:limited-views}
\vspace{-.1in}
\end{table}
%--------------------------------------

%--------------------------------------
\begin{figure}[!h]
 \begin{center}
$\begin{array}{c@{\hspace{0.05in}}c}

\includegraphics[width=0.465\linewidth]{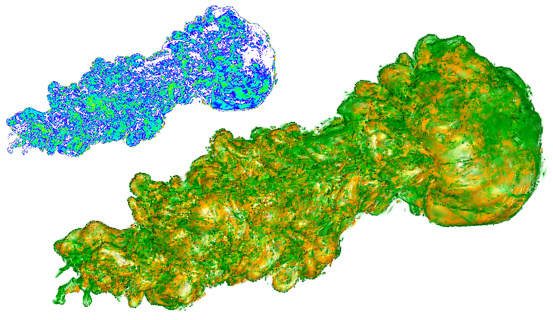} &
\includegraphics[width=0.465\linewidth]{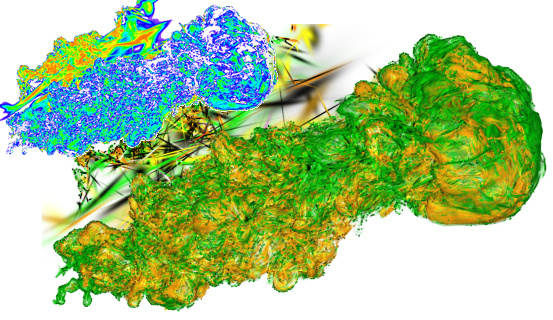} \\

\includegraphics[width=0.465\linewidth]{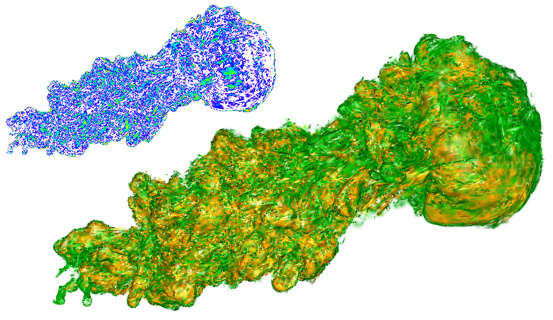} &
\includegraphics[width=0.465\linewidth]{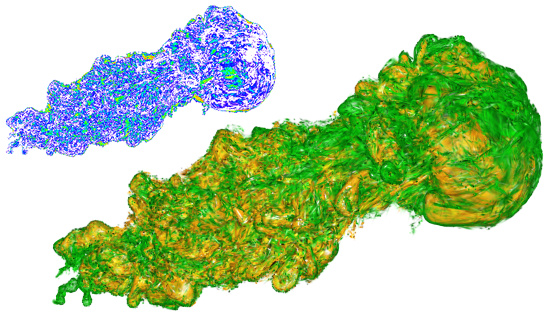} \\

\includegraphics[width=0.465\linewidth]{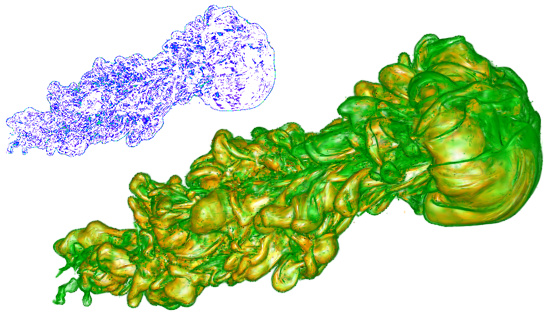} &
\includegraphics[width=0.465\linewidth]{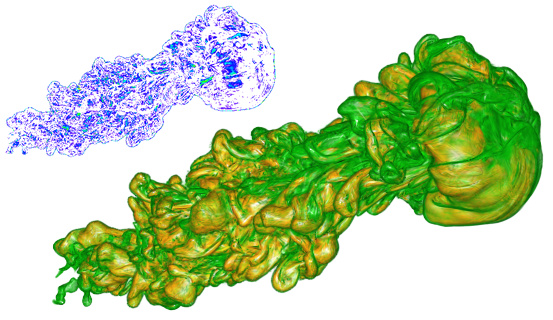} \\

\includegraphics[width=0.465\linewidth]{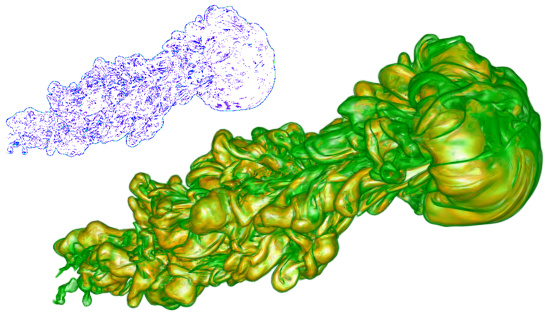} &
\includegraphics[width=0.465\linewidth]{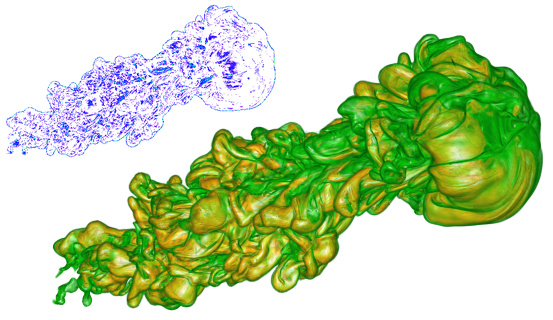} \\

\mbox{\small (a) ECoNGS} &
\mbox{\small (b) iVR-GS}

\end{array}$

\end{center}
\vspace{-.25in}
\caption{Comparing ECoNGS and iVR-GS under varying training views on the argon bubble dataset. Top to bottom: model trained under 6, 12, 42, and 92 views.}
\label{fig:limited-views}
\vspace{-.1in}
\end{figure}
%--------------------------------------

%--------------------------------------
\begin{figure}[!t]
\begin{center}
%\resizebox{\linewidth}{!}{%
$\begin{array}{c@{\hspace{0.025in}}c@{\hspace{0.025in}}c}
\includegraphics[width=0.315\linewidth]{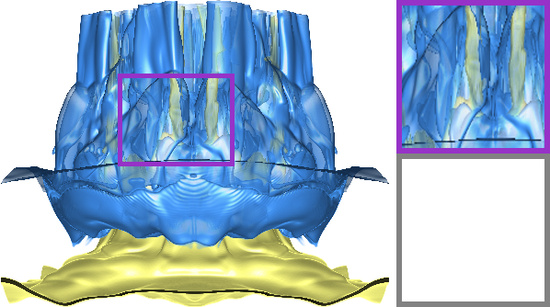} &
\includegraphics[width=0.315\linewidth]{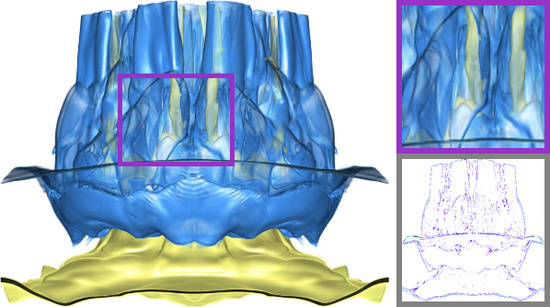} &
\includegraphics[width=0.315\linewidth]{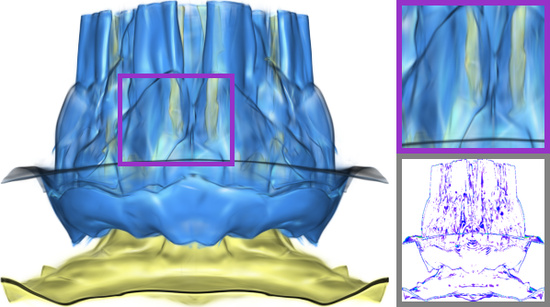} \\

\includegraphics[width=0.315\linewidth]{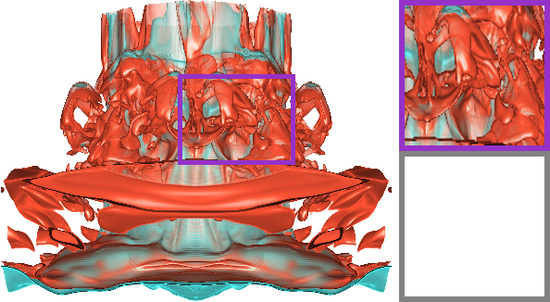} &
\includegraphics[width=0.315\linewidth]{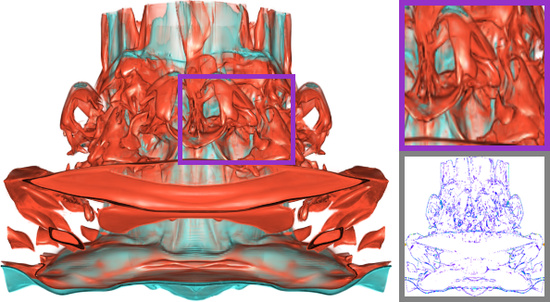} &
\includegraphics[width=0.315\linewidth]{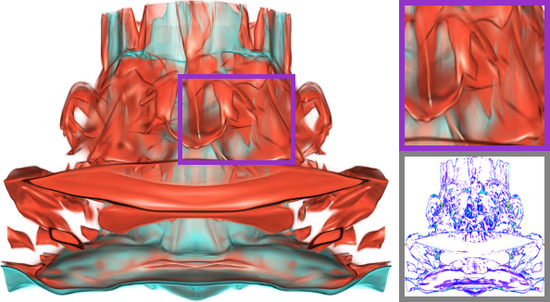} \\

\mbox{\small \hot{(a) GT}} &
\mbox{\small \hot{(b) ECoNGS}} &
\mbox{\small \hot{(c) VEG}}
\end{array}$
%}
\end{center}
\vspace{-.25in}
\caption{\hot{Novel view synthesis results of ECoNGS and VEG on the ionization (T) dataset.}}
\label{fig:veg}
\vspace{-.1in}
\end{figure}
%--------------------------------------

%--------------------------------------
\begin{table}[htb]
\centering
\caption{\hot{Average PSNR (dB), LPIPS, rendering framerate (FPS), and total model size (MS, in MB) of VEG and ECoNGS on the basic scenes and the composed scene of the ionization (T) dataset.}}
\vspace{-0.1in}\hot{
\resizebox{\linewidth}{!}{%
\begin{tabular}{l|ccc|ccc|c}
\multicolumn{1}{c}{}  & \multicolumn{3}{c}{basic scene} & \multicolumn{3}{c}{composed scene} & \\
method & PSNR $\uparrow$ & LPIPS $\downarrow$ & FPS $\uparrow$ & PSNR $\uparrow$ & LPIPS $\downarrow$ & FPS $\uparrow$ & MS $\downarrow$ \\
\hline
VEG~\cite{Dyken-VEG25} & 30.38 & 0.037 & 496.8 & 28.94 & 0.048 & \textbf{407.9} & \textbf{3.01} \\
ECoNGS & \textbf{34.34} & \textbf{0.019} & \textbf{548.0} & \textbf{33.45} & \textbf{0.024} & 374.9 & 3.32 \\
\end{tabular}
}}
\label{tab:veg}
\vspace{-.1in}
\end{table}
%--------------------------------------

\vspace{-0.05in}
\section{Robustness to Limited Training Views}
\label{sec:failure-cases}

In practical scenarios, the number of available training views may be limited due to rendering budget constraints.
To investigate the robustness of ECoNGS under such conditions, we evaluate both ECoNGS and iVR-GS with progressively fewer training views on the argon bubble dataset. 
Figure~\ref{fig:limited-views} and Table~\ref{tab:limited-views} present the qualitative and quantitative results.
As the number of training views decreases, both methods degrade in quality.
However, ECoNGS demonstrates greater robustness than iVR-GS, maintaining higher PSNR and lower LPIPS across all view settings.
This robustness stems from the joint learning mechanism in ECoNGS, where shared anchor primitives and MLP parameters effectively regularize the optimization and prevent overfitting to the limited training data.
In contrast, iVR-GS, which optimizes each scene independently, is more prone to artifacts and missing structures when training views are sparse.

\vspace{-0.05in}
\section{\hot{Comparison with Volume Encoding Gaussians}}
\label{sec:veg}

\hot{A concurrent work, volume encoding Gaussians (VEG)~\cite{Dyken-VEG25}, can also represent the VolVis scene given training images from multiple basic scenes. 
Instead of optimizing a separate model for each basic scene, VEG trains a single set of Gaussians, with each Gaussian encoding the underlying scalar value, thereby decoupling visual appearance from the data representation. The user-defined TF can then be applied at render time to map these scalar values to color and opacity. By training on multiple basic scenes, a single VEG model can therefore also render the composed scene.
To compare VEG with ECoNGS, we evaluate both on the ionization (T) dataset, using its four basic scenes and two composed scenes (each obtained by combining two basic TFs). 
For a fair comparison, we train VEG on the four basic-TF views with Blinn-Phong shading to match ECoNGS.
Figure~\ref{fig:veg} and Table~\ref{tab:veg} report the qualitative and quantitative results, respectively. ECoNGS achieves substantially higher reconstruction accuracy than VEG on both basic and composed scenes.
VEG, on the other hand, is slightly more compact and renders the composed scene faster, as it maintains only one set of Gaussians instead of compositing several.
However, the single set of Gaussians in VEG is less effective at capturing view-dependent lighting and fine structural detail. As shown in Figure~\ref{fig:veg}, VEG reconstructs the detailed structures less accurately than ECoNGS, leading to its slightly lower visual quality.}

%--------------------------------------
\begin{table}[!t]
\centering
\caption{Ablation on implicit vs.\ explicit attribute storage on the ionization (T) dataset.
Average PSNR (dB), LPIPS, rendering framerate (FPS), total training time (TT, in minutes), and total model size (MS, in MB) are reported.}
\label{tab:ablation-architecture}
\vspace{-0.1in}
\resizebox{0.85\linewidth}{!}{%
\begin{tabular}{lccccc}
variant & PSNR $\uparrow$ & LPIPS $\downarrow$ & FPS $\uparrow$ & TT $\downarrow$ & MS $\downarrow$ \\
\hline
ECoNGS & \textbf{34.34} & \textbf{0.019} & 548.81 & 10.36 & 3.32 \\
all explicit & 34.01 & 0.020 & \textbf{572.82} & 12.28 & 5.67 \\
all implicit & 33.79 & 0.023 & 519.05 & \textbf{9.21} & \textbf{2.48} \\
partial explicit & 34.21 & 0.020 & 502.73 & 12.90 & 3.17 \\
\end{tabular}
}
\vspace{-.1in}
\end{table}
%--------------------------------------

\vspace{-0.05in}
\section{Ablation on Implicit vs.\ Explicit Attribute Storage}
\label{sec:ablation-architecture}

With our ECoNGS method, offset attributes $\Delta\mathbf{x}^g$ are stored explicitly and entropy-coded into the bitstream, while shading-related attributes $\{\Delta\mathbf{c}, \bm{n}, k^a, k^d, k^s, \beta\}$ are predicted implicitly by lightweight MLPs.
To investigate how this design choice affects performance, we compare three variants on the ionization (T) dataset:
(1) {\em all explicit} stores both offsets and shading-related attributes in the bitstream;
(2) {\em all implicit} implicitly encodes all attributes into lightweight MLPs; and
(3) {\em partial explicit} stores both offsets and normals explicitly in the bitstream while predicting the remaining attributes via MLPs.

Table~\ref{tab:ablation-architecture} reports the results.
ECoNGS achieves the best PSNR and LPIPS while maintaining a competitive model size.
``All explicit'' stores all attributes in the bitstream, which incurs the largest model size due to the explicitly stored shading-related attributes and yields the lowest PSNR, showing that implicit prediction via lightweight MLPs is more parameter-efficient for these attributes.
``All implicit'' achieves the smallest model size by encoding all attributes implicitly, but suffers the worst reconstruction accuracy, suggesting that MLPs cannot fully recover the fine-grained spatial detail captured by explicit offsets.
``Partial explicit'' stores normals explicitly, achieving comparable PSNR to ECoNGS but at the cost of slower rendering and longer training time.
These results confirm that the hybrid ECoNGS design, combining explicit offsets with implicit shading attributes, achieves the best trade-off between compactness and reconstruction quality.

%--------------------------------------
\begin{figure}[!t]
\begin{center}
%\resizebox{\linewidth}{!}{%
$\begin{array}{c@{\hspace{0.02in}}c@{\hspace{0.02in}}c@{\hspace{0.02in}}c@{\hspace{0.02in}}c}

\rotatebox{90}{\makebox[0.215\linewidth][c]{\small \hot{CUPID}}} &
\includegraphics[height=0.8125in]{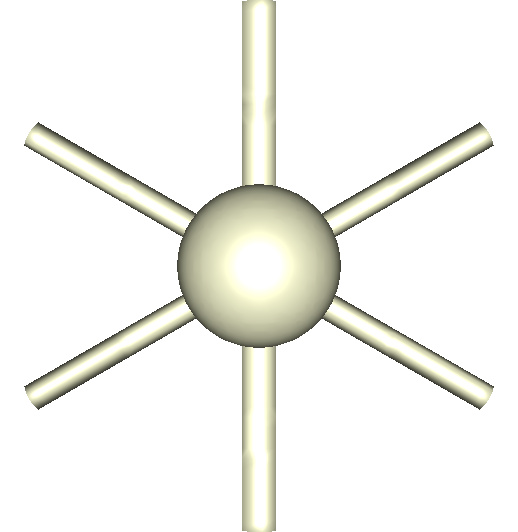} &
\includegraphics[height=0.8125in]{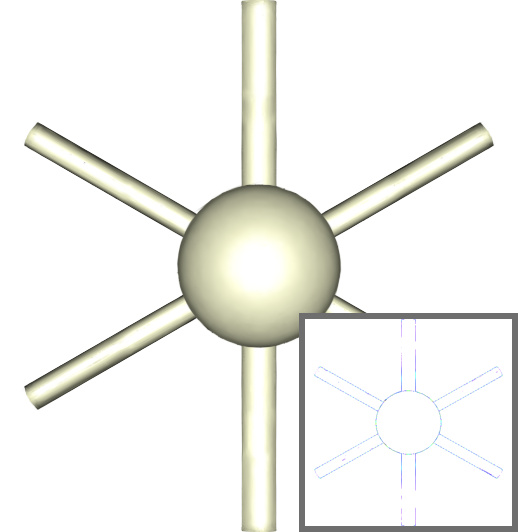} &
\includegraphics[height=0.8125in]{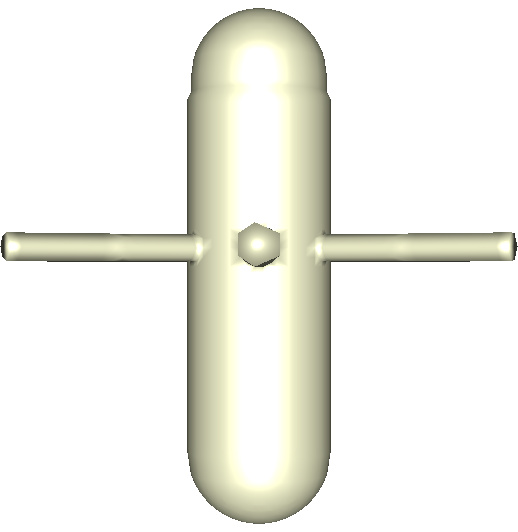} &
\includegraphics[height=0.8125in]{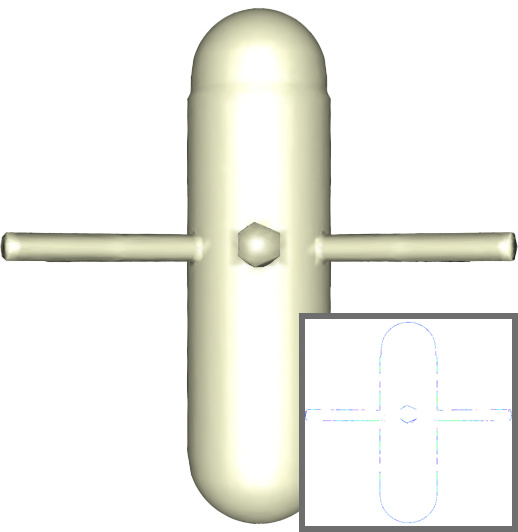} \\
\rotatebox{90}{\makebox[0.215\linewidth][c]{\small \hot{RPCOM}}} &
\includegraphics[height=0.8125in]{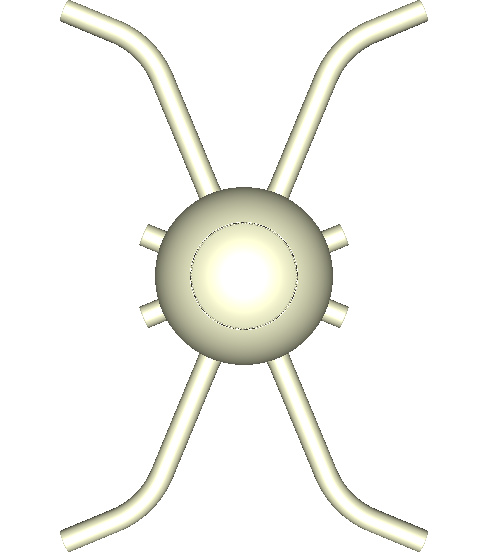} &
\includegraphics[height=0.8125in]{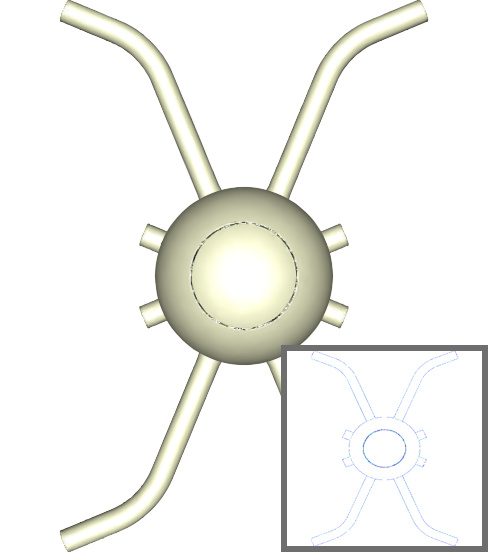} &
\includegraphics[height=0.8125in]{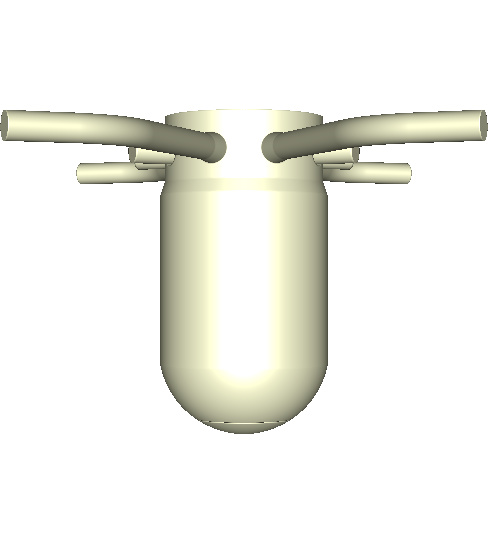} &
\includegraphics[height=0.8125in]{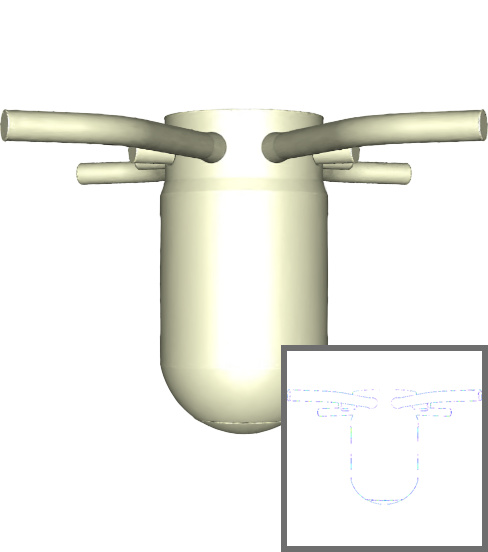} \\
 &
\mbox{\small \hot{(a) GT}} &
\mbox{\small \hot{(b) ECoNGS}} &
\mbox{\small \hot{(c) GT}} &
\mbox{\small \hot{(d) ECoNGS}} \\
\end{array}$
%}
\end{center}
\vspace{-.25in}
\caption{\hot{Novel view synthesis results of ECoNGS on two unstructured-mesh datasets, CUPID (top) and RPCOM (bottom).}}
\label{fig:unstruct}
\vspace{-.1in}
\end{figure}
%--------------------------------------

%--------------------------------------
\begin{table}[!t]
\centering
\caption{\hot{Evaluation of ECoNGS on two unstructured-mesh datasets. Average PSNR (dB), LPIPS, total model size (MS, in MB), and rendering framerate (FPS) are reported.}}
\vspace{-0.1in}\hot{
\resizebox{0.7\linewidth}{!}{%
\begin{tabular}{lcccc}
dataset & PSNR $\uparrow$ & LPIPS $\downarrow$ & MS $\downarrow$ & FPS $\uparrow$ \\
\hline
CUPID & 36.39 & 0.018 & 0.64 & 424.7 \\
RPCOM & 36.30 & 0.018 & 0.57 & 455.6 \\
\end{tabular}
}}
\label{tab:unstruct}
\vspace{-.1in}
\end{table}
%--------------------------------------

\vspace{-0.05in}
\section{\hot{Evaluation on Unstructured-Mesh Data}}
\label{sec:unstruct}

\hot{Since ECoNGS operates only on rendered multi-view images and a point cloud sampled from the data for initialization, it is agnostic to the underlying grid type and can represent scenes built from unstructured data equally well. 
We verify this on two unstructured-mesh datasets~\cite{Cho-NED19}, CUPID and RPCOM, each containing 1,279,375 and 12,440,366 spatial coordinates, respectively.
As shown in Figure~\ref{fig:unstruct} and Table~\ref{tab:unstruct}, ECoNGS reconstructs both scenes with high fidelity, confirming that ECoNGS is compatible and can generalize to scenes built from unstructured data without any modification to its pipeline.}

\vspace{-0.05in}
\section{\hot{Rendering Time Breakdown: 3DGS vs.\ MLP}}
\label{sec:timing-breakdown}

\hot{To study how the MLP in ECoNGS affects rendering performance, we measure the per-view rendering time and decompose it into two parts: the MLP part, which generates neural Gaussians from the anchor features, and the Gaussian rasterization part, which includes Blinn-Phong shading and rasterization. We include vanilla 3DGS~\cite{Kerbl-TOG23} as the baseline method for comparison and evaluate over all basic scenes in the univariate dataset from the main paper.

%--------------------------------------
\begin{table}[htb]
\centering
\caption{\hot{Per-view rendering time of ECoNGS and 3DGS on basic scenes from different datasets. We report the time for the MLP feedforward part (MLP), the Gaussian rasterization part (rastr), and the total rendering time (total), all measured in milliseconds.}}
\label{tab:timing-breakdown}
\vspace{-0.1in}\hot{
\resizebox{0.8\linewidth}{!}{%
\begin{tabular}{llcccc}
dataset & method & MLP & rastr & total & \# G \\
\hline
\multirow{2}{*}{ionization (T)}  & 3DGS   & -- & 1.43 & 1.43 & 105,730 \\
                                 & ECoNGS & 1.06 & 0.76 & 1.82 & 97,306 \\
\hline
\multirow{2}{*}{combustion (MF)} & 3DGS   & -- & 1.35 & 1.35 & 154,319 \\
                                 & ECoNGS & 1.04 & 0.65 & 1.69 & 62,572 \\
\hline
\multirow{2}{*}{supernova}       & 3DGS   & -- & 1.47 & 1.47 & 185,046 \\
                                 & ECoNGS & 1.16 & 0.78 & 1.94 & 145,359 \\
\hline
\multirow{2}{*}{vortex}          & 3DGS   & -- & 1.42 & 1.42 & 134,156 \\
                                 & ECoNGS & 1.00 & 0.69 & 1.69 & 70,905 \\
\end{tabular}
}}
\vspace{-.1in}
\end{table}
%--------------------------------------

Table~\ref{tab:timing-breakdown} reports the timing results. The MLP feedforward, which is absent in vanilla 3DGS, adds computational cost and is the primary source of ECoNGS's rendering overhead. However, this overhead is largely offset by the Gaussian rasterization step. Specifically, ECoNGS predicts view-dependent opacity via the MLP, allowing Gaussians that make negligible contributions to the current view to be culled before rasterization. As a result, fewer Gaussians are processed during rasterization, making the Gaussian rasterization part faster than that of 3DGS. Therefore, the additional cost of the MLP is largely offset by the reduced rasterization workload, and the overall rendering speed of ECoNGS remains only slightly slower than 3DGS while still achieving real-time performance. Given the substantially more compact and editable representation provided by ECoNGS, this minor overhead is negligible for interactive exploration.}

\section{\hot{Evaluation on Different Hyperparameter Settings}}
\label{sec:abl-feat-k}

\hot{ECoNGS has two key hyperparameters: the dimension of the per-anchor feature $\bm{f}^a$ and the number of neural Gaussians $K$ decoded from each anchor. We study their effects on the supernova dataset by varying one hyperparameter at a time while keeping the other at its default value used in the paper. Table~\ref{tab:abl-feat-k} reports the results.
As shown in the table, increasing the feature dimension from $50$ to $100$ yields similar reconstruction quality but noticeably increases the model size. Increasing $K$ from $10$ to $20$ slightly improves reconstruction quality, but the gain does not offset the larger model size and slower rendering speed due to the increased number of decoded Gaussians. On the other hand, using a smaller feature dimension or a smaller $K$ leads to lower reconstruction quality. Therefore, we choose a feature dimension of $50$ and $K=10$ as a balanced tradeoff among reconstruction quality, compactness, and rendering efficiency.}

%--------------------------------------
\begin{table}[htb]
\centering
\caption{\hot{Average PSNR (dB), LPIPS, model size (MS, in MB), and rendering framerate (FPS) for ECoNGS with different hyperparameter settings on the basic scenes of the supernova dataset. The best results are highlighted in bold.}}
\vspace{-0.1in}
\hot{
\resizebox{0.8\linewidth}{!}{%
\begin{tabular}{llcccc}
hyperparameters & value & PSNR $\uparrow$ & LPIPS $\downarrow$ & MS $\downarrow$ & FPS $\uparrow$ \\
\hline
\multirow{3}{*}{feature dimension}   & 32  & 31.25 & 0.045 & 3.37 & 505 \\
                                        & 50 & \textbf{31.55} & \textbf{0.043} & \textbf{4.40} & \textbf{514} \\
                                        & 100 & 31.51 & \textbf{0.043} & 7.68 & 488 \\
\hline
\multirow{3}{*}{$K$} & 5  & 31.36 & 0.046 & 4.33 & \textbf{538} \\
                                        & 10 & 31.55 & 0.043 & \textbf{4.40} & 514 \\
                                        & 20 & \textbf{31.63} & \textbf{0.040} & 5.24 & 469 \\
\end{tabular}
}}
\label{tab:abl-feat-k}
\vspace{-0.1in}
\end{table}
%--------------------------------------

% \clearpage % flush remaining appendix floats so references come last

% if specified like this the section will be committed in review mode
\vspace{-0.1in}
\acknowledgments{This research was supported in part by the U.S.\ National Science Foundation through grants IIS-2101696, OAC-2104158, IIS-2401144, and CCF-2550610, and the U.S.\ Department of Energy through grant DE-SC0023145. The authors thank the anonymous reviewers for their insightful comments.}

\vspace{-0.05in}
\bibliographystyle{abbrv-doi-hyperref-narrow}

\bibliography{template}

\end{document}